\documentclass{article}

\usepackage[main,final]{neurips_2026}
\usepackage{enumitem}
\usepackage[utf8]{inputenc} 
\usepackage[T1]{fontenc}    
\usepackage{url}            
\usepackage{booktabs}       
\usepackage{amsfonts}       
\usepackage{nicefrac}       
\usepackage{microtype}      
\usepackage{xcolor}         

\usepackage{microtype}
\usepackage{graphicx}
\usepackage{subfigure}
\usepackage{subcaption}
\usepackage{comment}
\usepackage{appendix}
\usepackage{verbatim}
\usepackage{comment}
\usepackage{algorithm}
\usepackage{algorithmic}
\usepackage{multicol}

\usepackage{caption}

\usepackage{amsmath}
\usepackage{amssymb}
\usepackage{mathtools}
\usepackage{amsthm}
\usepackage[normalem]{ulem}
\newcommand{\redstrike}{%
  \bgroup
  \markoverwith{\textcolor{red}{\rule[0.5ex]{2pt}{0.4pt}}}%
  \ULon
}

\usepackage{hyperref}
\usepackage[capitalize,noabbrev]{cleveref}

\usepackage{multirow}

\theoremstyle{plain}
\newtheorem{theorem}{Theorem}[section]
\newtheorem{proposition}[theorem]{Proposition}
\newtheorem{lemma}[theorem]{Lemma}
\newtheorem{corollary}[theorem]{Corollary}
\theoremstyle{definition}
\newtheorem{definition}[theorem]{Definition}
\newtheorem{assumption}[theorem]{Assumption}
\theoremstyle{remark}
\newtheorem{remark}[theorem]{Remark}

\title{Robust Satisficing Ensemble: Scalable Model Aggregation Under Distribution Shifts}

\author{
  Ahmet Faruk Cetinkaya\thanks{Equal contribution.} \hspace{0.001em} \thanks{Work completed while at Bilkent University. Currently at the School of Electrical and Computer Engineering, Georgia Institute of Technology.} \\
  Dept. of Electrical and Electronics Eng.\\
  Bilkent University\\
  Ankara, Türkiye \\
  \texttt{a.cetinkaya3@gatech.edu} \\
  \And
  Enes Agirman\footnotemark[1] \\
  Dept. of Electrical and Electronics Eng.\\
  Bilkent University \\
  Ankara, Türkiye \\
  \texttt{enes.agirman@bilkent.edu.tr} \\
  \And
  Cem Tekin \\
  Dept. of Electrical and Electronics Eng.\\
  Bilkent University \\
  Ankara, Türkiye \\
  \texttt{cemtekin@ee.bilkent.edu.tr} \\
}

\begin{document}

\maketitle

\begin{abstract}

Standard
ensemble methods can improve robustness by
mitigating individual model stochasticity, but they
frequently fail to generalize beyond the training
distribution, leaving systems vulnerable to
out-of-distribution (OOD) data. We introduce the Robust Satisficing Ensemble (RSE), a scalable probabilistic aggregation framework designed to secure model combinations by minimizing their fragility against distributional changes. By bridging a robust satisficing objective with a quadratic risk surrogate, RSE derives an analytical inner adversary. This fundamental reformulation enables highly efficient weight updates without the computational bottleneck of standard minimax optimization, crucially decoupling optimization complexity from sample size to scale effectively to large datasets. In particular, we prove a geometrically decaying truncation error for the fragility approximation via Lagrange-Bürmann expansion and establish stationarity guarantees for RSE. We evaluate RSE across a comprehensive spectrum of distribution shifts, including adversarial label shift (SST-5, TREC), natural subpopulation and temporal shifts (CivilComments, HuffPost). Our results demonstrate that RSE consistently outperforms strong robust baselines, including GroupDRO, IRM, and SRM. These findings establish RSE as a practical, theoretically grounded safeguard for reliable model aggregation in dynamic environments.

\end{abstract}

\section{Introduction}

Individual models that perform well on training data may remain fragile in deployment due to sensitivity to inherent stochasticity \citep{zhang2025stochasticity}, or spurious correlations in the training data \citep{li2025the}. Model aggregation is a common response to this fragility. Deep ensembles \citep{lakshminarayanan2017deep}, ensemble search \citep{zaidi2021neural}, and combinatorial or averaging-based aggregation schemes \citep{shen2022deep,arpit2022ensemble} exploit diversity across predictors to reduce variance and improve uncertainty estimation. These gains, however, are distribution-dependent. Because aggregation weights and selection rules are typically tuned on finite reference data, these configurations frequently fail when deployment conditions change. Recent studies for predictive uncertainty and real-world shift consistently report substantial in-distribution / out-of-distribution gaps, including under natural, temporal, and subgroup shift \citep{ovadia2019can,koh2021wilds,gardner2023benchmarking}. Under such shifts, constituent predictors can fail coherently rather than independently, so naive aggregation may inherit rather than cancel their errors.

Recent work detects or adapts to shift at test time, through label-shift correction and test-time adaptation \citep{lipton2018detecting,sun2020test,wang2021tent}; recent benchmarks also document substantial practical pitfalls in these adaptation pipelines \citep{zhao2023pitfalls}.  Yet standard aggregation objectives still optimize weights for a fixed reference distribution, leaving the learned combination vulnerable when deployment conditions vary. Moreover, this vulnerability is not confined to distribution shift: even under the standard i.i.d. assumption, optimizing aggregation weights on finite calibration data can exploit sampling noise in the empirical metric, inducing a gap between calibration performance and true out-of-sample reliability, an instance of the optimizer's curse \citep{smith2006}.

\begin{figure}[t]
    \centering
    
    \includegraphics[width=\textwidth]{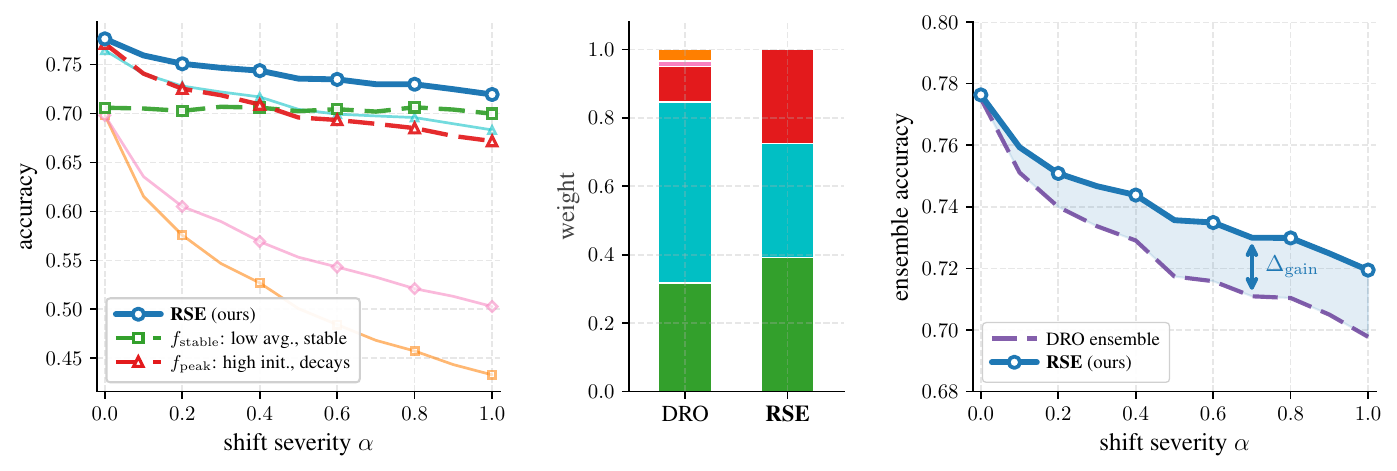}

    \vspace{0.1cm}
    \makebox[\textwidth][c]{
        \makebox[0.33\textwidth][c]{\small (\textbf{a}) Individual model profiles}
        \makebox[0.33\textwidth][c]{\small (\textbf{b}) Weight assignment}
        \makebox[0.33\textwidth][c]{\small (\textbf{c}) Ensemble performance under shift}
    }
    \vspace{0.1cm}

    \caption{\textbf{RSE combines stable and high-peak models to prevent catastrophic failure.} 
    \textbf{(a)} Constituent models face a trade-off between high initial accuracy that decays rapidly ($f_{\text{peak}}$, dashed red triangles) and lower, stable performance ($f_{\text{stable}}$, dashed green squares). The faded solid lines represent the remaining active constituent models participating in the ensemble. The resulting RSE ensemble (thick blue with circles) consistently outperforms
individual models across the entire shift trajectory.
    \textbf{(b)} Colors match (a). While DRO yields a suboptimal weight allocation, RSE explicitly minimizes fragility to strategically balance highly stable and high-peak predictors.
    \textbf{(c)} This integration allows RSE to maintain superior, stable accuracy across all shift severities $\alpha$, achieving a significant gain ($\Delta_{\text{gain}}$) over DRO, which collapses alongside its fragile components. 
    \emph{(Note: Visualizations reflect actual empirical results on the TREC dataset. See Section~\ref{sec:experiments} for detailed setups and benchmarks.)}}
    \label{fig:concept_teaser}
\end{figure}

We introduce the RSE, a probabilistic aggregation framework for fixed pre-trained predictors under distribution shift grounded in a novel robust satisficing (RS) formulation. Figure~\ref{fig:concept_teaser} previews the central effect: under increasing label shift, RSE remains substantially more stable than its constituent models and the Distributionally Robust Optimization (DRO) method. Our contributions are:
\begin{itemize}
    \item \textbf{First tractable robust satisficing for aggregation.} We propose the first RS framework designed for the aggregation of fixed pre-trained models through a fundamental reformulation of its optimization mechanics. Specifically, under KL label shift and Brier-score risk, the inner adversary takes an exact exponential-tilt form, allowing the scalar fragility root to be computed via a Lagrange--Bürmann expansion rather than bisection.
    \item \textbf{Scalable and efficient optimization with guarantees.} By operating purely on precomputed sufficient statistics, RSE reduces weight updates to simplex-constrained convex Quadratic Programs (QP). This decouples the optimization complexity from the dataset size $N$. With proximal stabilization, we prove sufficient decrease, stationarity of limit points, and a stopping certificate.
    \item \textbf{Robustness gains across benchmarks.} Across diverse distribution-shift tasks, RSE outperforms standard aggregation and robust optimization baselines, especially in severe-shift regimes where competing methods degrade sharply.
\end{itemize}

\subsection{Related Works}

\begin{table}[htbp]
\centering
\caption{Conceptual comparison of RSE against existing aggregation frameworks. \textit{Analytical Inner Solution} refers to an analytical determination of the inner solution, including any parameters needed
to specify the adversary, for given model weights.}
\label{tab:conceptual_comparison}
\resizebox{\textwidth}{!}{%
\begin{tabular}{l l c c c c}
\toprule
\textbf{Method} & \textbf{Objective} & \textbf{Ambiguity-Set Free} & \textbf{Analytical Inner Solution} & \textbf{$N$-Indep. Update} \\
\midrule
ERM & Average Risk & $\checkmark$ & N/A & $\times$ \\
GroupDRO & Worst-Case & $\times$ & $\times$ \textit{(Gradient)} & $\times$ \\
SRM & Worst-Case & $\times$ & $\times$ \textit{(Iterative)} & $\times$ \\
KL-RS & Fragility & $\checkmark$ & $\times$ \textit{(Bisection)} & $\times$ \\
\midrule
\textbf{RSE (Ours)} & \textbf{Fragility} & $\checkmark$ & $\checkmark$ & $\checkmark$ \\
\bottomrule
\end{tabular}%
}
\end{table}

Classical ensemble methods improve predictive reliability by aggregating diverse
models \citep{breiman96,lakshminarayanan2017deep}, with deep learning extensions
leveraging optimization-trajectory diversity or direct weight averaging
\citep{huang2017snapshot,wortsman2022model}. With large-scale foundation models, inference-time aggregation has emerged as a primary mechanism for reliability \citep{li2024more, ai2025beyond}. 
Let $\Delta_K$ denote probability simplex of dimension K, 
$\mathbf w\in\Delta_M$ aggregation weights, $\mathbf q_{\rm ref}\in\Delta_K$
the reference data distribution, and $R(\mathbf w,\mathbf q)$ the ensemble risk.
Most calibration-based aggregation schemes choose weights by solving, explicitly
or implicitly, $\min_{\mathbf w\in\Delta_M} R(\mathbf w,\mathbf q_{\rm ref})$.
This average-risk objective can improve performance on the calibration
distribution, but it gives no direct control of $R(\mathbf w,\mathbf q)$ when the
deployment distribution $\mathbf q$ moves away from $\mathbf q_{\rm ref}$.

A parallel line of work addresses distribution shift through DRO, replacing the
fixed-reference objective with
$\min_{\mathbf w\in\Delta_M}\sup_{\mathbf q\in\mathcal U_\rho(\mathbf q_{\rm ref})}
R(\mathbf w,\mathbf q)$, where
$\mathcal U_\rho(\mathbf q_{\rm ref})
:=\{\mathbf q\in\Delta_K:\Delta(\mathbf q,\mathbf q_{\rm ref})\le\rho\}$ is an
ambiguity set around $q_{\rm ref}$
\citep{namkoong2016,Sagawa*2020Distributionally}. This formulation protects
against shifts contained in $\mathcal U_\rho$, but its behavior depends critically
on the radius $\rho$: small sets may miss the deployment distribution, while large
sets can yield overly conservative weights. Moreover, the resulting min--max objectives introduce significant computational overhead. Structure-informed Risk Minimization
(SRM) applies DRO to ensemble learning by dynamically adjusting ensemble weights using a distributional graph \citep{qiao2025structureinformed}.
Yet, it still relies on an iterative alternating gradient procedure, inheriting computational bottlenecks (see Table~\ref{tab:conceptual_comparison}).

RS replaces the ambiguity-set radius $\rho$ with an aspiration level $\tau$ and
optimizes fragility, the degradation rate needed to maintain a risk guarantee. With ensemble weights $\mathbf w\in\Delta_M$, fragility is defined as
$\kappa_\tau(\mathbf w)
:=\inf\{\kappa\ge0:
R(\mathbf w,\mathbf q)\le
\tau+\kappa\Delta(\mathbf q,\mathbf q_{\rm ref}),\ \forall \mathbf q\in\Delta_K\}$,
and the RS objective is $\min_{\mathbf w\in\Delta_M}\kappa_\tau(\mathbf w)$
\citep{long2023,sim2025analytics}. Thus, unlike DRO, which optimizes worst-case risk over a prescribed ambiguity ball, RS specifies an interpretable aspiration level $\tau$ and controls risk deterioration away from the reference distribution, reducing sensitivity to conservative or
misspecified ambiguity sets. Early studies demonstrate its effectiveness in decision making and Bayesian satisficing \citep{saday2023robust, saday2025robust}. Current RS implementations, however, are bottlenecked by intractable minimax dual problems and rely on slow iterative bisection searches \citep{yan2024kl}. Ours is the first work to adapt RS for pre-trained model aggregation. By coupling the RS objective with a quadratic risk surrogate, RSE derives an analytical series representation of the
fragility parameter through Lagrange--B\"urmann inversion. Together with precomputed quadratic risk statistics, this
yields an optimization procedure whose weight updates are simplex-constrained convex QPs independent of the training set size.

\section{Problem Formulation}
\label{sec:problem_formulation}

We consider supervised prediction with an ensemble of $M$ pre-trained models on a $K$-class task with the label set \(
\mathcal{Y} := \{y_1,\dots,y_K\}.
\)
Let $\mathcal{F} = \{f_1,\dots,f_M\}$ denote the set of base models and let
\(
\Delta_d := \big\{\mathbf{z} \in \mathbb{R}^d_{\ge 0} : \mathbf{1}^\top \mathbf{z} = 1 \big\}
\)
denote the probability simplex in $d$ dimensions.
Given an input $x$, model $f_m$ outputs a probability vector
\(
\mathbf{p}_m(x) \in \Delta_K .
\)
The ensemble is parameterized by weights $\mathbf{w} = (w_1,\dots,w_M)^\top \in \Delta_M$ and outputs the aggregated probability distribution
\(
\widehat{\mathbf{p}}_{\mathbf{w}}(x)
=
\sum_{m=1}^M w_m\,\mathbf{p}_m(x).
\)
Let $L \colon \mathcal{Y} \times \Delta_K \to \mathbb{R}_{\ge 0}$ be a loss function.
For each class $y_j \in \mathcal{Y}$ we define the class-conditional risk
\begin{equation}
\label{eq:class_conditional_risk_main}
r(\mathbf{w}, y_j)
:=
\mathbb{E}_{X \sim P(X \mid Y = y_j)}
\Big[
L\big(y_j, \widehat{\mathbf{p}}_{\mathbf{w}}(X)\big)
\Big].
\end{equation}
Collecting these into a vector gives
\begin{equation} \label{eq:r(w)}
\mathbf{r}(\mathbf{w})
:=
\big(r(\mathbf{w},y_1),\dots,r(\mathbf{w},y_K)\big)^\top
\in [0,\infty)^K .
\end{equation} 

A label distribution is a vector $\mathbf{q} \coloneqq (q_1,\dots,q_K)^\top \in \Delta_K$ with $q_j := P(Y = y_j)$.
We fix a \emph{reference} label distribution
\(
\mathbf{q}_{\mathrm{ref}} \in \Delta_K,
\)
representing the nominal operating regime (e.g., a held-out calibration set, the training set). Throughout our KL-based theoretical
analysis, we assume that the reference distribution has
full support, i.e., $(q_{\mathrm{ref}})_j>0$ for every $j$.
For any $\mathbf{q} \in \Delta_K$, the corresponding expected risk of the ensemble is
\begin{equation}
\label{eq:mixture_risk_main}
R(\mathbf{w}, \mathbf{q})
:=
\sum_{j=1}^K q_j\,r(\mathbf{w},y_j)
=
\mathbf{q}^\top \mathbf{r}(\mathbf{w}) .
\end{equation}
To quantify how different a prospective deployment distribution $\mathbf q$ is from the reference $\mathbf q_{\mathrm{ref}}$, we use a nonnegative discrepancy $\Delta(\mathbf q,\mathbf q_{\mathrm{ref}})\ge 0$, with $\Delta(\mathbf q,\mathbf q_{\mathrm{ref}})=0$ if and only if $\mathbf q=\mathbf q_{\mathrm{ref}}$. The framework below is stated for a general $\Delta$; in our algorithm, we instantiate $\Delta$ with KL divergence.

Instead of minimizing $R(\mathbf{w},\mathbf{q}_{\mathrm{ref}})$ directly, we adopt a robust satisficing view.
We fix an \emph{aspiration level} $\tau > 0$ and require the ensemble to be satisfactory at the reference distribution:
\(
R(\mathbf{w},\mathbf{q}_{\mathrm{ref}})
\le
\tau .
\)
The parameter $\tau$ sets the target in-distribution performance: smaller $\tau$ demands a stronger baseline at $\mathbf{q}_{\mathrm{ref}}$, while larger $\tau$ allows more slack in exchange for stability under changes in data distribution. Our goal is to find robustly satisfactory weights $\mathbf{w}$ by minimizing the ensemble's ``fragility'' against distribution shifts.
\begin{definition}[Fragility]
\label{def:fragility_main}
Fix a reference distribution $\mathbf{q}_{\mathrm{ref}} \in \Delta_K$ and an aspiration level $\tau>0$. For a weight vector $\mathbf{w} \in \Delta_M$, the fragility $\kappa_{\tau}(\mathbf{w})$ is defined as the minimum $\kappa\ge 0$ such that for all $\mathbf{q} \in \Delta_K$:
\(
R(\mathbf{w}, \mathbf{q}) \le \tau + \kappa\, \Delta(\mathbf{q}, \mathbf{q}_{\mathrm{ref}}).
\)
\end{definition}

\begin{remark}[Shift-dependent certificate]
\label{rem:shift_certificate_main}
Definition~\ref{def:fragility_main} implies that for any $\mathbf w$ with finite $\kappa_\tau(\mathbf w)$ and any shift budget
$\varepsilon\ge0$,
\(
\sup_{\mathbf q\in\Delta_K:\,\Delta(\mathbf q,\mathbf q_{\mathrm{ref}})\le\varepsilon}
R(\mathbf w,\mathbf q)
\le
\tau+\kappa_\tau(\mathbf w)\varepsilon .
\)
Hence $\kappa_\tau(\mathbf w)$ is the certified slope of worst-case risk
deterioration away from $\mathbf q_{\mathrm{ref}}$.
\end{remark}

Fix $\tau>0$, $\mathbf q_{\mathrm{ref}}\in\Delta_K$.
Let
\(
\mathcal W_\tau
:=
\{
\mathbf w\in\Delta_M:\;
R(\mathbf w,\mathbf q_{\mathrm{ref}})<\tau<\max_{j\in\{1,\dots,K\}} r(\mathbf w,y_j)
\}.
\) Outside $\mathcal W_\tau$, we follow the extended-value convention:
$\kappa_\tau(\mathbf w)=0$ if
$\tau\ge\max_j r(\mathbf w,y_j)$; otherwise,
$\kappa_\tau(\mathbf w)=+\infty$ if
$R(\mathbf w,\mathbf q_{\mathrm{ref}})\ge\tau$.

\begin{remark}
    While the theoretical formulation relies on the restricted domain $\mathcal{ W_\tau}$, in our experiments we found it sufficient to constrain $\mathbf{w} \in \Delta_M$. Across all datasets evaluated, the entire optimization trajectory naturally remained within $\mathcal{ W_\tau}$ without explicit enforcement.
\end{remark}

\noindent
Equivalently, $\kappa_\tau(\mathbf w)$ can be viewed as the largest excess risk
per unit discrepancy from $\mathbf{q}_{\mathrm{ref}}$. As shown in \citep{saday2023robust}, this definition implies that fragility can be expressed as:
\begin{equation}
\label{eq:fragility_calculation}
\kappa_{\tau}(\mathbf{w})
=
\sup_{\mathbf{q} \in \Delta_K,\; \mathbf{q}\neq \mathbf{q}_{\mathrm{ref}}}
\frac{
R(\mathbf{w},\mathbf{q}) - \tau
}{
\Delta(\mathbf{q},\mathbf{q}_{\mathrm{ref}})
} .
\end{equation}
We therefore choose $\mathbf w\in\mathcal W_\tau$ minimizing $\kappa_\tau(\mathbf w)$. Equivalently, we may view the optimization over $\Delta_M$ under the extended definition, which assigns $\kappa_\tau(\mathbf w)=+\infty$ when $R(\mathbf w,\mathbf q_{\mathrm{ref}})\ge\tau$.

\section{The RSE Algorithm}
\label{sec:method}

\subsection{Quadratic Risk Surrogate for Model Aggregation}
\label{subsec:quad-risk}

We take $L$ to be the \textit{Brier Score} between the one-hot label and the ensemble predicted label distribution:
\(
L\big(y_j,\widehat{\mathbf{p}}_{\mathbf{w}}(x)\big)
=
\big\|\mathbf{e}_j - \widehat{\mathbf{p}}_{\mathbf{w}}(x)\big\|_2^2
\)
using \(\widehat{\mathbf{p}}_{\mathbf{w}}(x)
=
\sum_{m=1}^M w_m\,\mathbf{p}_m(x)\) and expanding the square in~\eqref{eq:class_conditional_risk_main} gives exactly,
\(
r(\mathbf{w},y_j)
:= \mathbb{E}_{X\mid y_j} \Bigg[ \bigg\| \mathbf{e}_j - \sum_{m=1}^M w_m \mathbf{p}_m(X) \bigg\|_2^2 \Bigg] \nonumber 
= \mathbb{E}_{X\mid y_j} \Bigg[ \mathbf{e}_j^\top \mathbf{e}_j - 2 \sum_{a} w_a \mathbf{e}_j^\top \mathbf{p}_a(X) \nonumber 
+ \sum_{a,b} w_a w_b \mathbf{p}_a(X)^\top \mathbf{p}_b(X) \Bigg].\)
By linearity of expectation, and noting $\mathbf{e}_j^\top \mathbf{e}_j = 1$, we can bring the expectation inside the summations:
\(r(\mathbf{w},y_j)
= 1 - 2 \sum_{a} w_a \mathbb{E}_{X\mid y_j} [p_{a,j}(X)]
+ \sum_{a,b} w_a w_b \mathbb{E}_{X\mid y_j} \big[\mathbf{p}_a(X)^\top \mathbf{p}_b(X)\big].
\label{eq:quad-risk-expanded}
\)
Define the per-class first and second moments
\(
\bar{\mathbf{p}}_m^{(j)}
:=
\mathbb{E}_{X\mid Y=y_j}[\mathbf{p}_m(X)],
\)
\(
P_{ab}^{(j)}
:=
\mathbb{E}_{X\mid Y=y_j}\!\big[\mathbf{p}_a(X)^\top \mathbf{p}_b(X)\big]
\)
and collect them into
\(
\mathbf{A}_j \in \mathbb{R}^{M\times M},
(\mathbf{A}_j)_{ab} = P_{ab}^{(j)},
\)
\(
\mathbf{b}_j \in \mathbb{R}^M,
(\mathbf{b}_j)_m = -2 (\bar{\mathbf{p}}_m^{(j)})_j.
\)
Then $r(\mathbf{w},y_j)$ can be written as the convex quadratic form
\begin{equation}
\label{eq:quad-risk-matrix}
r(\mathbf{w},y_j)
=
\mathbf{w}^\top \mathbf{A}_j \mathbf{w}
+
\mathbf{b}_j^\top \mathbf{w}
+
1 .
\end{equation}

Each $\mathbf{A}_j$ is positive semidefinite. In practice, the expectations in $\bar{\mathbf{p}}_m^{(j)}$ and $P_{ab}^{(j)}$ are estimated from the training data. The derivation and further details are in Appendix~\ref{app:psd_derivation}.

\subsection{KL-Based Fragility and Inner Adversary}
\label{subsec:fragility-kl}

We instantiate the discrepancy in Definition~\ref{def:fragility_main} with the Kullback--Leibler divergence between label distributions:
\(
\Delta(\mathbf{q},\mathbf{q}_{\mathrm{ref}})
=
D_{\mathrm{KL}}(\mathbf{q}\,\|\,\mathbf{q}_{\mathrm{ref}})
=
\sum_{j=1}^K
q_j \log\frac{q_j}{(q_{\mathrm{ref}})_j}.
\)
For fixed $\mathbf{w}$ and $\kappa>0$, it is convenient to work with the KL-penalized inner problem where we want to find the label shift that maximizes the penalized expected risk.
\begin{equation}
\label{eq:penalized-kl}
\Phi(\kappa;\mathbf{w})
=
\sup_{\mathbf{q}\in\Delta_K}
\Big\{
\mathbf{q}^\top \mathbf{r}(\mathbf{w})
-
\kappa\, D_{\mathrm{KL}}(\mathbf{q}\,\|\,\mathbf{q}_{\mathrm{ref}})
\Big\},
\end{equation}
where $\mathbf{r}(\mathbf{w})$ is built from~\eqref{eq:r(w)}. See Appendix \ref{app:phi_derivation} for the derivation.
The form of the optimal adversary and the value of~\eqref{eq:penalized-kl} can be made explicit.

\begin{lemma}[KL inner problem]
\label{lem:kl-inner}
For any fixed $\mathbf{w}$ and $\kappa>0$, the problem~\eqref{eq:penalized-kl} has a unique maximizer $\mathbf{q}^\star\in\Delta_K$ given by
\begin{equation}
\label{eq:qstar-softmax-method}
(q^\star)_j
=
\frac{(q_{\mathrm{ref}})_j \exp\big(r(\mathbf{w},y_j)/\kappa\big)}
{\sum_{k=1}^K (q_{\mathrm{ref}})_k \exp\big(r(\mathbf{w},y_k)/\kappa\big)},
\end{equation}
and the corresponding optimal value is
\begin{equation}
\label{eq:Phi-def}
\Phi(\kappa;\mathbf{w})
=
\kappa \log
\Big(
\sum_{j=1}^K
(q_{\mathrm{ref}})_j
\exp\big(r(\mathbf{w},y_j)/\kappa\big)
\Big).
\end{equation}
\end{lemma}
The proof is given in Appendix~\ref{app:kl_derivation}. Thus, under a KL penalty, the adversary corresponds to an exponential tilt of the reference label distribution and $\Phi(\kappa;\mathbf{w})$ plays the role of a log-moment generating function of the class-wise risks. Combining this explicit inner solution with the fragility ratio definition  in~\eqref{eq:fragility_calculation} yields a one-dimensional characterization of the fragility index in the KL case.

\begin{lemma}[Fragility for KL discrepancy]
\label{lem:kl-fragility-scalar}
Let
$\mathbf w\in\mathcal W_\tau$, i.e.,
$R(\mathbf w,\mathbf q_{\mathrm{ref}})<\tau
<\max_j r(\mathbf w,y_j)$.
Then there exists a \emph{unique} $\kappa_\tau(\mathbf w)\in(0,\infty)$ such that
\begin{align}
& \tau
=
 \Phi\big(\kappa_{\tau}(\mathbf{w});\mathbf{w}\big) 
=
\kappa_{\tau}(\mathbf{w})
\log
\Big(
\sum_{j=1}^K
(q_{\mathrm{ref}})_j
\exp\big(r(\mathbf{w},y_j)/\kappa_{\tau}(\mathbf{w})\big)
\Big), \label{eq:tau-kappa-main}
\end{align}
and the associated worst-case label distribution is obtained by substituting $\kappa=\kappa_{\tau}(\mathbf{w})$ into~\eqref{eq:qstar-softmax-method}.
\end{lemma}

Equation~\eqref{eq:tau-kappa-main} characterizes KL fragility as a scalar root in
$\kappa$. Its derivation follows from Appendix~\ref{app:phi_derivation}
together with the closed form of $\Phi$ in Appendix~\ref{app:kl_derivation};
existence and uniqueness are established in Appendix~\ref{app:solve_kappa}.
To evaluate this root without inner bisection, we apply Lagrange--B\"urmann
inversion after the change of variables $S=\kappa^{-1}$.

\begin{proposition}[Lagrange--B\"urmann expansion of KL fragility]
\label{prop:lb-kappa-main}
Fix $\mathbf w\in\mathcal W_\tau$ and set
$\alpha_j:=r(\mathbf w,y_j)$,
$\tau_0:=R(\mathbf w,\mathbf q_{\mathrm{ref}})$, and
$z:=\tau-\tau_0$. Define $f:\mathbb R\to\mathbb R$ by $f(0):=\tau_0$ and
$f(S):=S^{-1}\log\!\big(\sum_{j=1}^K
(q_{\mathrm{ref}})_j e^{\alpha_j S}\big)$
for $S\neq0$,
and write its Taylor expansion at $S=0$ as
\(
f(S)=\sum_{m=0}^{\infty}a_{m+1}S^m
=
\tau_0+a_2S+a_3S^2+\cdots.
\)
Thus $a_1=\tau_0$ and
\(
a_2=\frac12\sum_{j=1}^K(q_{\mathrm{ref}})_j(\alpha_j-\tau_0)^2
\),
the half-variance of the class-wise risks under $\mathbf q_{\mathrm{ref}}$. Lagrange--B\"urmann inversion yields a local
inverse series whose convergence radius $R_z(\mathbf w)>0$
is defined in~\eqref{eq:Rz_def}.
For
$0<z=\tau-\tau_0<R_z(\mathbf w)$,
\(
S_\tau(\mathbf w)
:=
\kappa_\tau(\mathbf w)^{-1}
=
\sum_{n=1}^{\infty}c_n z^n,
\)
where the coefficients $\{c_n\}_{n\ge1}$ are given
by~\eqref{eq:LB-general}. Taking the reciprocal gives, for sufficiently small
$z>0$, the induced Laurent expansion
$\kappa_\tau(\mathbf w)=\sum_{m=-1}^{\infty}d_m z^m$,
whose first two terms, for fixed $\mathbf w$, are
\(
\kappa_\tau(\mathbf w)
=
\frac{a_2}{z}+\frac{a_3}{a_2}+O(z),
z=\tau-\tau_0\downarrow0.
\)
Here $a_3=\frac16\sum_{j=1}^K(q_{\mathrm{ref}})_j
(\alpha_j-\tau_0)^3$, and the remainder vanishes at least
linearly as $\tau$ approaches $\tau_0$ from above.

\end{proposition}

\begin{corollary}[Finite-truncation error]
\label{cor:lb-trunc-main}
Let
$S_N(z):=\sum_{n=1}^N c_nz^n$ and
$\kappa_N(z):=\sum_{m=-1}^N d_mz^m$
denote the order-$N$ truncations of the Lagrange--B\"urmann inverse expansion
and the induced Laurent expansion in Proposition~\ref{prop:lb-kappa-main}.
For the fixed weight vector $\mathbf w$, choose
$\varrho$ and $B_H$ satisfying Assumption~\ref{ass:H_disk},
and set \(
U:=\varrho/B_H
\)
and
\(
\eta(z):=|z|/U.
\) Here $U\le R_z(\mathbf w)$ is a sufficient convergence
radius for the inverse series, as established after
Theorem~\ref{thm:S_trunc}.
The fragility bounds below are evaluated at
$z=\tau-\tau_0>0$ for $\mathbf w\in\mathcal W_\tau$. Then, for $|z|<U$,
\(
|S_\tau(\mathbf w)-S_N(z)|
\le
\frac{\varrho\,\eta(z)^{N+1}}{1-\eta(z)}.
\)
Moreover, fix any $r\in(0,U)$ satisfying
\(
\frac{B_H\,\eta_r}{1-\eta_r}\le \frac{c_1}{2},
\eta_r:=r/U.
\)
Then, for $0<|z|<r$,
\(
|\kappa_\tau(\mathbf w)-\kappa_N(z)|
\le
\frac{2}{c_1}
\frac{(|z|/r)^{N+2}}{1-|z|/r}
\frac{1}{|z|}.
\)
Thus the inverse-series error decays geometrically in $N$, and the induced
Laurent approximation error for $\kappa_\tau$ decays geometrically up to the
expected pole factor $1/|z|$.
\end{corollary}

RSE computes $\kappa_t$ by truncating the
Lagrange--B\"urmann inverse expansion and taking the
reciprocal of the resulting finite sum, avoiding
a root solve inside each iteration. The coefficient formula and Laurent
recursion are given in~\eqref{eq:LB-general} and
\eqref{eq:kappa-laurent}--\eqref{eq:d-recursion}; the proof of the truncation
bounds is given in Appendix~\ref{app:solve_kappa_trunc}.

\subsection{Robust Satisficing Ensemble Algorithm}
\label{subsec:rse-main}

We combine the quadratic risk model and the KL-based fragility into an
alternating procedure that approximately solves
$\min_{\mathbf w\in\mathcal W_\tau}\kappa_\tau(\mathbf w)$ by reweighting the
ensemble on the simplex. Given $\mathbf w_t$, we first compute
the per-class risks $r_j := r(\mathbf w_t,y_j)$ using~\eqref{eq:quad-risk-matrix}. We then compute $\kappa_t$, an approximation to
$\kappa_\tau(\mathbf w_t)$ obtained by truncating
the inverse series in
Proposition~\ref{prop:lb-kappa-main} and taking its reciprocal.

\[
(q_t^\star)_j
=
\frac{(q_{\mathrm{ref}})_j \exp\big(r_j^{(t)}/\kappa_t\big)}
{\sum_{k=1}^K (q_{\mathrm{ref}})_k \exp\big(r_k^{(t)}/\kappa_t\big)}.
\]
For a candidate $\mathbf{w}$ we define the adversarially tilted quadratic risk
\(
\tilde R_t(\mathbf{w})
=
\sum_{j=1}^K (q_t^\star)_j\,r(\mathbf{w},y_j)
\nonumber
=
\mathbf{w}^\top \mathbf{A}_{\mathrm{adv},t} \mathbf{w}
+
\mathbf{b}_{\mathrm{adv},t}^\top \mathbf{w}
+
1 ,
\)
where
\(
\mathbf{A}_{\mathrm{adv},t}
=
\sum_{j=1}^K (q_t^\star)_j \mathbf{A}_j
\), \(
\mathbf{b}_{\mathrm{adv},t}
=
\sum_{j=1}^K (q_t^\star)_j \mathbf{b}_j
\).
Since each $\mathbf{A}_j$ is PSD and $(q_t^\star)_j\ge 0$, $\mathbf{A}_{\mathrm{adv},t}$ is PSD as well.
The quadratic subproblem induced by $\mathbf{A}_{\mathrm{adv},t}\succeq 0$ can be ill-conditioned
(or admit multiple minimizers on the simplex) when the base models are highly correlated.
We therefore perform a \emph{proximal} update by adding a stabilizer
$\frac{\rho}{2}\|\mathbf{w}-\mathbf{w}_t\|_2^2$ with $\rho>0$.
This keeps the step a simplex-constrained convex QP while making it $\rho$-strongly convex,
hence ensuring a unique update $\mathbf{w}_{t+1}$.

Beyond well-posedness, the proximal term also provides a principled optimization guarantee:
Appendix~\ref{app:rse_convergence} shows that, under mild regularity conditions on the KL fragility root,
the update yields a \emph{uniform sufficient decrease} in the fragility objective and converges to a stationary point
of $\kappa_\tau(\cdot)$ over the simplex. Also, the commonly used stopping criterion
$\|\mathbf{w}_{t+1}-\mathbf{w}_t\|_2\le \varepsilon_{\mathrm{stop}}$ admits an explicit
certificate in terms of the KKT residual of the QP subproblem. The weight update solves the QP:
\begin{equation}
\label{eq:adv-qp-method}
\begin{aligned}
\mathbf{w}_{t+1}
=
\arg\min_{\mathbf{w}}
\quad &
\mathbf{w}^\top \mathbf{A}_{\mathrm{adv},t} \mathbf{w}
+
\mathbf{b}_{\mathrm{adv},t}^\top \mathbf{w}
+
\frac{\rho}{2}\|\mathbf{w}-\mathbf{w}_t\|_2^2
\quad \text{s.t.}\quad &
\mathbf{w} \in \Delta_M.
\end{aligned}
\end{equation} 
Any off-the-shelf convex QP solver (e.g., via \texttt{cvxpy}) can be used to compute $\mathbf{w}_{t+1}$. See Algorithm ~\ref{alg:rse} for the overall procedure for implementing RSE.

\begin{algorithm}[t]
\caption{Robust Satisficing Ensemble (RSE)}
\label{alg:rse}
\begin{multicols}{2}
\begin{algorithmic}[1]
\REQUIRE $\mathbf{q}_{\mathrm{ref}}$, $\tau$, $\{\mathbf{A}_j,\mathbf{b}_j\}_{j=1}^K$, truncation order $N$, $\rho>0$, $\varepsilon_{\mathrm{stop}}$
\STATE Init $\mathbf{w}_0 \in \Delta_M$; $\Delta_w \gets +\infty$, $t \gets 0$
\WHILE{$\Delta_w > \varepsilon_{\mathrm{stop}}$}
    \FOR{$j = 1,\dots,K$}
        \STATE $r_j \gets r(\mathbf{w}_t, y_j)$ \quad \textit{// Eq.~\eqref{eq:quad-risk-matrix}}
    \ENDFOR
    \STATE $\kappa_t \gets \kappa_{\tau}(\mathbf{w}_t)$ \quad \textit{// Prop.~\ref{prop:lb-kappa-main}}
    \FOR{$j = 1,\dots,K$}
        \STATE $(q_t^\star)_j \propto (q_{\mathrm{ref}})_j\,e^{r_j/\kappa_t}$ \quad \textit{// Eq.~\eqref{eq:qstar-softmax-method}}
    \ENDFOR
    \STATE $\mathbf{A}_{\mathrm{adv},t} \gets \textstyle\sum_{j}(q_t^\star)_j\,\mathbf{A}_j$
    \STATE $\mathbf{b}_{\mathrm{adv},t} \gets \textstyle\sum_{j}(q_t^\star)_j\,\mathbf{b}_j$
    \STATE $\mathbf{w}_{t+1} \gets$ solve~\eqref{eq:adv-qp-method}
    \STATE $\Delta_w \gets \|\mathbf{w}_{t+1}-\mathbf{w}_t\|_2$; \quad $t \gets t+1$
\ENDWHILE
\STATE \textbf{return} $\mathbf{w}_t$, $\kappa_{\tau}(\mathbf{w}_t)$
\end{algorithmic}
\end{multicols}
\end{algorithm}

We analyze the idealized exact-root iteration in
Proposition~\ref{prop:rse_descent_main} and separately
provide a stationarity error certificate for the
implemented truncated-root update in
Proposition~\ref{prop:approx_root_stationarity_main}.

\begin{proposition}[Descent, stationarity, and a stopping certificate for proximal RSE]
\label{prop:rse_descent_main}
Assume that the iterates $\{\mathbf w_t\}$ produced by
Algorithm~\ref{alg:rse} remain in a compact subset
$\mathcal K\subset\mathcal W_\tau$.
Let $c_{\min}>0$ be the uniform lower bound from Lemma~\ref{lem:cmin}.  Let $\epsilon>0$ be such that the closed $\epsilon$-neighborhood
$\mathcal K_\epsilon\subset\mathcal W_\tau$, and $M_{\mathrm{grad}}<\infty$ be as in
Lemma~\ref{lem:prox_suff_decrease_appconv}. Let $L$ be a Lipschitz constant of $\nabla\kappa_\tau$
on $\mathcal K_\epsilon$. If
\(
\rho >
\max\left\{
\frac{L}{2c_{\min}},
\frac{M_{\mathrm{grad}}}{\epsilon}
\right\},
\)
then the sequence $\{\kappa_\tau(\mathbf w_t)\}_{t\ge 0}$ is nonincreasing and convergent,
$\|\mathbf w_{t+1}-\mathbf w_t\|_2\to 0$, and every limit point of $\{\mathbf w_t\}$ is stationary for
$\min_{\mathbf w\in\Delta_M}\kappa_\tau(\mathbf w)$ in the sense of Definition~\ref{def:stationary_kappa_appconv}. Furthermore, the stopping rule
$\|\mathbf w_{t+1}-\mathbf w_t\|_2\le \varepsilon_{\mathrm{stop}}$
guarantees that the stationarity residual of the tilted-risk problem at
$\mathbf w_{t+1}$ is at most
$\rho\,\varepsilon_{\mathrm{stop}}$
(Corollary~\ref{cor:stopping_certificate_appconv}).

\end{proposition}

\noindent\textbf{Proof sketch.}
Differentiate the KL root $\Phi(\kappa;\mathbf w)=\tau$ to express $\nabla \kappa_\tau(\mathbf w)$ via the tilted risk gradient, then use $\rho$-strong convexity of the proximal QP to obtain sufficient decrease and the KKT-based stopping certificate. Full proofs are given in Appendix~\ref{app:rse_convergence}.

\begin{proposition}[Stationarity certificate with truncated fragility roots]
\label{prop:approx_root_stationarity_main}
Proposition~\ref{prop:rse_descent_main} treats the KL fragility root as exact,
whereas Algorithm~\ref{alg:rse} uses the finite Lagrange--B\"urmann truncation
$\widehat\kappa_t$ in place of $\kappa_t=\kappa_\tau(\mathbf w_t)$. The root
residual $\varepsilon_t:=\Phi(\widehat\kappa_t;\mathbf w_t)-\tau$ controls the
stationarity gap of the exact fragility objective. Assume $\widehat\kappa_t>0$ for all $t$ and the iterates remain in
a compact subset of $\mathcal W_\tau$ and the segment between $\widehat\kappa_t$ and $\kappa_t$
lies in a region where
$|\partial_\kappa\Phi(\kappa;\mathbf w_t)|
\ge m_\kappa>0$ uniformly in $t$.
Let $\widehat{\mathbf q}_t=\mathbf q^\star(\widehat\kappa_t,\mathbf w_t)$, and
let $\mathbf w_{t+1}$ be the proximal QP update formed with
$\widehat{\mathbf q}_t$. Then there exist constants
$C_\varepsilon,C_d<\infty$, independent of $t$, such that
\(
\operatorname{dist}\!\left(
\mathbf 0,\,
\nabla_{\mathbf w}\kappa_\tau(\mathbf w_{t+1})
+
N_{\Delta_M}(\mathbf w_{t+1})
\right)
\le
C_\varepsilon|\varepsilon_t|
+
C_d\|\mathbf w_{t+1}-\mathbf w_t\|_2
\) where $N_{\Delta_M}(\mathbf w)$ denotes the normal cone of the simplex $\Delta_M$ at $\mathbf w$.
Consequently, if $|\varepsilon_t|\le\varepsilon_{\mathrm{root}}$ and
$\|\mathbf w_{t+1}-\mathbf w_t\|_2\le\varepsilon_{\mathrm{stop}}$, then
$\mathbf w_{t+1}$ is
$(C_\varepsilon\varepsilon_{\mathrm{root}}
+
C_d\varepsilon_{\mathrm{stop}})$-stationary for
$\min_{\mathbf w\in\Delta_M}\kappa_\tau(\mathbf w)$.
\end{proposition}

\noindent\textbf{Proof sketch.}
Let $\kappa_t=\kappa_\tau(\mathbf w_t)$ be the exact root. Since
$\Phi(\kappa_t;\mathbf w_t)=\tau$, the mean-value theorem and the uniform slope
condition imply
\(
|\widehat\kappa_t-\kappa_t|\le |\varepsilon_t|/m_\kappa
\).
On the compact well-posed region, the adversarially tilted gradient is
Lipschitz, so using $\widehat\kappa_t$ instead of $\kappa_t$ perturbs the exact
gradient by
\(
O(|\varepsilon_t|+\|\mathbf w_{t+1}-\mathbf w_t\|_2)
\).
Combining this perturbation bound with the proximal-QP KKT condition and the
implicit-gradient identity for $\nabla\kappa_\tau$ yields the stated
stationarity residual. The stopping certificate follows by substituting the two
stopping tolerances. Full proof is given in
Appendix ~\ref{app:rse_convergence:approx_root}.

\section{Experiments}
\label{sec:experiments}
We evaluate RSE across four diverse tasks targeting core vulnerabilities of pre-trained model aggregation. For adversarial label shifts, we apply a severe-shift protocol to SST-5 \citep{socher-etal-2013-recursive} and TREC \citep{li-roth-2002-learning}. For natural distribution shifts, we use CivilComments \citep{koh2021wilds} (subpopulation) and HuffPost \citep{yao2022wild} (temporal drift). Implementation details, including baseline hyperparameters, aspiration level $\tau$ selection, and input formatting, are provided in Appendix~\ref{app:implementation}; individual-model results, additional experiments, and extended results are in Appendix~\ref{app:additional_results}. Our code is available at \url{https://github.com/Bilkent-CYBORG/RSE}.

\noindent \textbf{Baselines.} We compare RSE against Non-optimization baselines (Uniform, Accuracy-Weighted, Majority Vote) and Optimization-based baselines: Empirical Risk Minimization (ERM) \citep{vapnik1998statistical}, Group DRO \citep{Sagawa*2020Distributionally}, Invariant Risk Minimization (IRM) \citep{arjovsky2019invariant}, SRM \citep{qiao2025structureinformed}, and KL-Divergence-based Robust Satisficing (KL-RS) \citep{yan2024kl}.

\noindent \textbf{Models and Setup.} To construct the ensemble, we employ a set of $M=8$ pre-trained Large Language Models (LLMs) with varying architectures and sizes:  Mistral-Small-24B-Instruct~\citep{mistral2025small24b}, Mistral-7B-Instruct-v0.3~\citep{jiang2023mistral7b}, Phi-4~\citep{abdin2024phi}, Phi-3.5 mini~\citep{abdin2024phi3technicalreporthighly}, Qwen3-4B~\citep{yang2025qwen3}, Gemma-3-12B-Instruct~\citep{team2025gemma}, GPT-oss-20B~\citep{agarwal2025gpt}, and Falcon-3-7B-Instruct~\citep{Falcon3}. We extract token-level confidence scores to yield normalized probability vectors rather than relying on verbalized confidence. 

\subsection{Robustness to Adversarial Shifts}
To validate RSE in adversarial environments, we evaluate it on SST-5 and TREC. We partition each original training set into training and test pools by randomly splitting within each class, preserving the original label proportions. We then synthesize label shifts on the test pools using a minority target shift protocol. For $\alpha\in[0,1]$, the target label distribution is
$P_{\mathrm{test}}=(1-\alpha)P_{\mathrm{pool}}+\alpha\delta_{\mathrm{hard}}$,
where $P_{\mathrm{pool}}$ is the empirical label distribution of the unshifted test pool and $\delta_{\mathrm{hard}}$ concentrates mass on the hardest class, identified by the lowest mean accuracy across the $M$ base models on a held-out validation set. Thus, $\alpha=0$ corresponds to the unshifted test-pool label distribution, while $\alpha=1$ yields a test set containing only the hardest class.

\textbf{SST-5 Dataset.}
We first evaluate RSE on SST-5, a fine-grained sentiment classification benchmark consisting of movie reviews labeled on a scale of 1 (very negative) to 5 (very positive). This domain offers a challenging test for robustness due to the nuanced semantic differences between adjacent sentiment classes. We use the official training split (8,544 instances).

\textbf{TREC Dataset.} 
Then, we evaluate RSE on the TREC dataset, a widely used benchmark for question classification that comprises questions categorized into six semantic types: abbreviation, description, entity, human, location, and numeric. The dataset's distinct semantic categories make it an effective testbed for robust classification. We use the official train split of TREC (5452 instances).

\noindent \textbf{Severe Shift Results.} Figure~\ref{fig:figure2} shows how accuracy changes with shift severity on SST-5 and TREC. While RSE achieves the highest average accuracy across all shift spectrums (detailed in Appendix~\ref{app:additional_results}), its primary mathematical objective is preventing catastrophic failure in highly unpredictable environments. Table~\ref{tab:severe_shift_results} isolates the Severe Shift ($\alpha=1.00$) regime. Standard baselines collapse under these conditions. On the nuanced sentiment analysis of SST-5, RSE achieves 36.3\%, outperforming the strongest robust baseline (DRO) by \textbf{+3.8\%} and ERM by \textbf{+8.0\%}. On the TREC dataset, RSE achieves 72.0\%, surpassing DRO by \textbf{+2.1\%}. Two-sided paired $t$-tests on 20 split-level mean accuracies
support RSE's advantage over all five optimization baselines
at the evaluated shift levels $\alpha\ge0.45$ on SST-5
($p<0.05$) and $\alpha\ge0.32$ on TREC ($p<0.01$).
Detailed results are reported in
Tables~\ref{tab:sst5_significance} and~\ref{tab:trec_significance}.

\begin{table}[t!] 
    \centering
    \caption{\textbf{Under Severe Distribution Shifts.} We report accuracy under extreme out-of-distribution conditions: complete adversarial shift to the hardest class for SST-5 and TREC ($\alpha=1.00$). Across both tasks, RSE significantly outperforms baselines in high-risk regimes. Best results in \textbf{bold}.} 
    \vspace{1em}
    \small 
    \setlength{\tabcolsep}{4pt} 
    \begin{tabular}{l ccc cccccc}
        \toprule
        \multirow{2}{*}{\textbf{Dataset}} & \multicolumn{3}{c}{\textbf{Non-optimization-based}} & \multicolumn{6}{c}{\textbf{Optimization-based}} \\
        \cmidrule(lr){2-4} \cmidrule(lr){5-10}
         & \textbf{Uniform} & \textbf{Acc. Wgh.} & \textbf{Maj. Vote} & \textbf{ERM} & \textbf{KL-RS} & \textbf{SRM} & \textbf{IRM} & \textbf{DRO} & \textbf{RSE}  \\
        \midrule
        SST-5 & 25.1\% & 24.5\% & 25.3\% & 28.3\% & 28.8\% & 29.3\% & 28.0\% & 32.5\% & \textbf{36.3\%} \\
        TREC & 61.7\% & 62.1\% & 60.8\% & 67.1\% & 66.9\% & 67.7\% & 66.7\% & 69.9\% & \textbf{72.0\%} \\
        \bottomrule
    \end{tabular}
    \label{tab:severe_shift_results} 
    
    \vspace{0.2em} 

    \begin{minipage}{0.40\textwidth}
        \centering
        \includegraphics[width=0.82\linewidth]{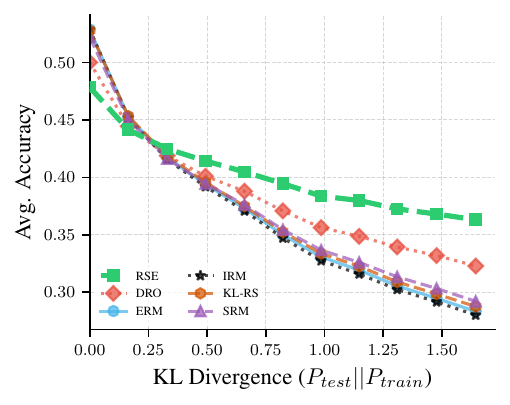}
        \centerline{\small (a) SST-5 Accuracy}
    \end{minipage}
    \hfill
    \begin{minipage}{0.40\textwidth}
        \centering
        \includegraphics[width=0.82\linewidth]{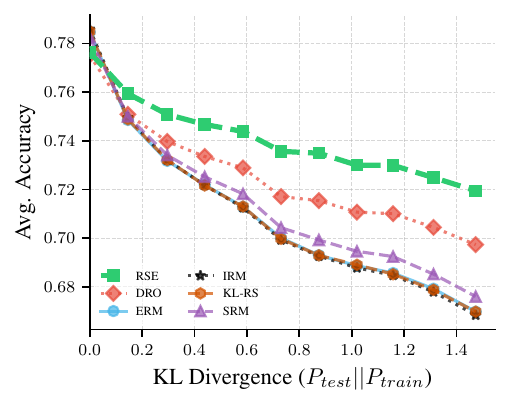}
        \centerline{\small (b) TREC Accuracy}
    \end{minipage}
    \vspace{1em}
    \captionof{figure}{\textbf{Performance under adversarial shift.} We report overall accuracy across 11 shift magnitudes with $\alpha$\ values selected to produce equally-spaced KL divergences $D_{\mathrm{KL}}(P_{\mathrm{test}} \| P_{\mathrm{train}})$, ranging from  $\alpha$\ = 0 (in-distribution) to  $\alpha$\ = 1 (most severe shift: test set composed entirely of the hardest class).  RSE consistently maintains higher accuracy compared to baselines as shift severity increases.}
    \label{fig:figure2}
\end{table}

\begin{table}[t]
\centering
\begin{minipage}[t]{0.51\textwidth} 
    \vspace{1pt} 
    \caption{Performance under natural distribution shifts. Worst Group (Year) and overall accuracies across all groups (years) are given for CivilComments and HuffPost. Best results in \textbf{bold}.}
    \vspace{1em}
    \label{tab:real_world_shifts}
    \centering
    \small
    \begin{tabular}{l cc cc}
    \toprule
     & \multicolumn{2}{c}{CivilComments} & \multicolumn{2}{c}{HuffPost} \\
    \cmidrule(lr){2-3} \cmidrule(lr){4-5}
    Method & Acc & WGA & Acc & WYA \\
    \midrule
    ERM & 70.1 & 56.8 & 78.1 & 76.8 \\
    DRO & 70.6 & 57.7 & 76.3 & 75.6 \\
    IRM & 69.8 & 56.5 & 78.4 & 77.0 \\
    SRM & 68.8 & 55.1 & 78.4 & 77.0 \\
    KL-RS & 69.6 & 56.3 & 78.1 & 76.7 \\
    \midrule
    RSE (Ours) & \textbf{72.1} & \textbf{59.7} & \textbf{78.8} & \textbf{77.2} \\
    \bottomrule
    \end{tabular}
\end{minipage}
\hfill
\begin{minipage}[t]{0.46\textwidth} 
    \vspace{1pt} 
    \caption{Per-iteration complexity and optimization times on HuffPost. $M$: models, $K$: classes, $N$: samples.}
    \vspace{1em}
    \label{tab:training_times}
    
    \centering
    \small
    \begin{sc}
    \begin{tabular}{lcc}
    \toprule
    Model & Per-Iter. Complexity & Time (s) \\
    \midrule
    RSE   & $O(M^2K + M^3)$  & 1.41 \\
    ERM   & $O(NM)$           & 1.70 \\
    KL-RS & $O(MN)$          & 2.59 \\
    IRM   & $O(MNK)$         & 3.03 \\
    DRO   & $O(MNK)$         & 4.67 \\
    SRM   & $O(M^2NK)$        & 8.30 \\
    \bottomrule
    \end{tabular}
    \end{sc}
\end{minipage}
\end{table}

\subsection{Robustness to Natural Distribution Shifts}

\textbf{Subpopulation Shift: CivilComments.}
To evaluate robustness under subpopulation shift, we use the CivilComments dataset from the WILDS benchmark, which contains online comments annotated for toxicity along with demographic group labels (e.g., Christian, Muslim, LGBTQ, male, female, White, Black). Following standard protocol with official train and test splits, we report both overall accuracy and Worst-Group Accuracy (WGA), which is a critical metric for fairness-aware applications. Table~\ref{tab:real_world_shifts} presents the results. Detailed per-group accuracy breakdowns are available in Table~\ref{tab:civil_comments_subpop} of Appendix~\ref{app:additional_results}. RSE achieves highest overall accuracy (72.1\%) and substantially outperforms all baselines on WGA (59.7\%), representing 2.0 points of absolute improvement over the next-best method (DRO at 57.7\%).

\textbf{Temporal Shift: HuffPost.}
Temporal distribution shifts pose a significant challenge for language models, as changes in writing style, topic prevalence, and current events lead to performance degradation over time. We evaluate temporal robustness on the HuffPost dataset, where the task is news category prediction from article headlines. We retain categories that appear consistently across all years, resulting in 9 categories. Following the same temporal split protocol from \citep{yao2022wild}, we train algorithms on articles from 2012--2015 and evaluate on 2016--2018. Table~\ref{tab:real_world_shifts} reports overall accuracy and Worst-Year Accuracy (WYA). Year-wise performance details are provided in Table~\ref{tab:huffpost_temporal} of Appendix~\ref{app:additional_results}. RSE achieves the best WYA (77.2\%) and overall accuracy (78.8\%), outperforming the second-best methods (IRM and SRM) by 0.4\%.

{
Bootstrap comparisons on the fixed test sets ($B=1000$)
provide additional evidence for these gains.
Results and statistical procedures are reported in
Tables~\ref{tab:civilcomments_significance} and
\ref{tab:huffpost_significance} in Appendix~\ref{app:additional_results}.
}

\subsection{Ablation Study}
\textbf{Computational Complexity.}
RSE has a one-time preprocessing stage that constructs the moment matrices,
followed by weight optimization using only these cached statistics.
Table~\ref{tab:training_times} compares the optimization runtimes on HuffPost;
RSE is the fastest (1.41s), achieving a 1.2--6$\times$ speedup.
The moment construction took only 0.03s and the
subsequent RSE optimization took 1.41s (1.44s total), excluding the shared
base-LLM inference cost. Crucially, RSE's per-iteration cost
$O(M^2K + M^3)$ is independent of dataset size $N$. By precomputing moment
matrices $\{\mathbf{A}_j, \mathbf{b}_j\}_{j=1}^K$ once in $O(NM^2K)$ time,
optimization iterations operate purely on $O(M^2K)$ sufficient statistics. This decoupling is invaluable for large-scale deployments where $N$ is massive but $M$ and $K$ remain modest. This architecture makes RSE exceptionally scalable, effectively managing the massive sample sizes of CivilComments (over 269,000 instances) and HuffPost (over 84,000 instances) without computational burden. Memory overhead is similarly independent of $N$, requiring only $O(M^2 K)$ space, unlike DRO and IRM which must retain per-sample predictions ($O(NMK)$). While the $O(M^3)$ QP solve dominates as $M$ grows, it remains computationally negligible for typical ensemble sizes.

\noindent \textbf{Comparison with Best Single Model.} As shown in Table~\ref{tab:shift_results}, RSE consistently outperforms the best model across all datasets. Notably, on TREC, RSE improves accuracy by over 2.8 points (74.1\% vs. 71.3\%). This confirms that RSE effectively leverages the diversity of the model pool to achieve robustness superior to any individual model.

\noindent \textbf{Aspiration level sensitivity analysis for RSE.} To assess the role of $\tau$ as a reference-risk target, we vary this hyperparameter within the solvable range $\tau \in [0.65, 0.94]$ on SST-5. The results reveal a clear robustness-accuracy trade-off governed by $\tau$. Under severe shift (Figure~\ref{fig:tau_analysis}a), increasing $\tau$ substantially improves worst-case accuracy, with RSE surpassing DRO and IRM beyond $\tau \approx 0.75$ and peaking near $\tau \approx 0.9$. Conversely, under no shift (Figure~\ref{fig:tau_analysis}b), higher $\tau$ values degrade in-distribution accuracy, allowing practitioners to mathematically bound their risk tolerance.

\noindent\textbf{Sensitivity to Model-Pool Composition.}
We further test whether RSE's gains depend on the particular eight-model
pool used in the main experiments. We consider leave-one-out ensembles,
random subsets containing 2--7 models, and pools augmented with highly
correlated, uncalibrated, or weak predictors. As shown for TREC in Table~\ref{tab:model_pool_sensitivity_trec},
RSE achieves the highest mean accuracy in every evaluated configuration.
The same pattern holds on SST-5, with the corresponding results and full
construction details reported in
Appendix~\ref{app:model_pool_sensitivity}.

\begin{table}[t]
\centering
\caption{\textbf{Sensitivity to model-pool composition on TREC.}
Mean accuracy (\%) across the full shift-severity spectrum. We modify the
original eight-model pool through leave-one-out removal, random subsets,
or the addition of correlated, uncalibrated, or weak predictors.
The highest mean accuracy before rounding in each row is in \textbf{bold}.}
\label{tab:model_pool_sensitivity_trec}
\small
\setlength{\tabcolsep}{4.2pt}
\begin{tabular}{lcccccc}
\toprule
\textbf{Scenario} &
\textbf{ERM} &
\textbf{IRM} &
\textbf{DRO} &
\textbf{KL-RS} &
\textbf{SRM} &
\textbf{RSE} \\
\midrule
Original (8 Models) & 71.3 & 71.1 & 72.8 & 71.1 & 71.7 & \textbf{74.1} \\
Leave-one-out       & 70.7 & 70.4 & 72.4 & 70.5 & 71.4 & \textbf{73.7} \\
Random Subset       & 68.4 & 68.3 & 69.4 & 68.1 & 69.1 & \textbf{70.5} \\
Original + Correlated        & 71.2 & 71.2 & 72.7 & 71.0 & 71.6 & \textbf{74.1} \\
Original + Uncalibrated      & 71.8 & 71.5 & 74.0 & 71.9 & 72.5 & \textbf{74.0} \\
Original + Weak Models       & 71.1 & 70.6 & 72.8 & 70.9 & 71.5 & \textbf{74.1} \\
\bottomrule
\end{tabular}
\end{table}

\begin{figure}[t]
    \centering
    \begin{minipage}[t]{0.35\textwidth}
        \vspace{-0.4em} 
        \captionof{table}{Average test accuracy (\%) of best single model versus RSE. (For SST-5 and TREC numbers are avg. accuracy across all shift magnitudes.)}
        \vspace{1em}
        \label{tab:shift_results}
        \centering
        \small
        \setlength{\tabcolsep}{3pt}
        \begin{tabular}{lcc}
        \toprule
        Dataset & Best Single & RSE \\
        \midrule
        SST-5        & 39.9 & \textbf{40.3} \\
        TREC         & 71.3 & \textbf{74.1} \\
        HuffPost     & 78.2 & \textbf{78.8} \\
        CivilComm.   & 71.6 & \textbf{72.1} \\
        \bottomrule
        \end{tabular}
    \end{minipage}
    \hfill
    \begin{minipage}[t]{0.60\textwidth}
        \vspace{0pt} 
        \centering
        \begin{minipage}[t]{0.49\linewidth}
            \vspace{0pt}
            \centering
            \includegraphics[width=\linewidth]{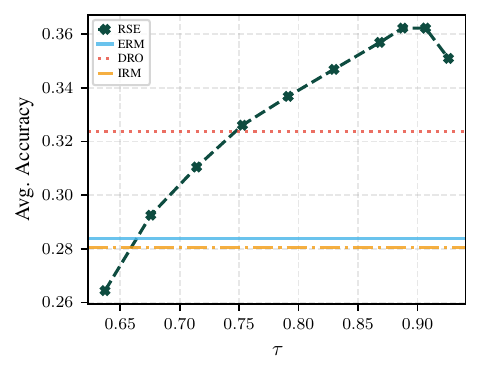}
            \vspace{1em}
            \centerline{\scriptsize (a) Severe shift ($\alpha = 1$)}
        \end{minipage}
        \hfill
        \begin{minipage}[t]{0.49\linewidth}
            \vspace{0pt}
            \centering
            \includegraphics[width=\linewidth]{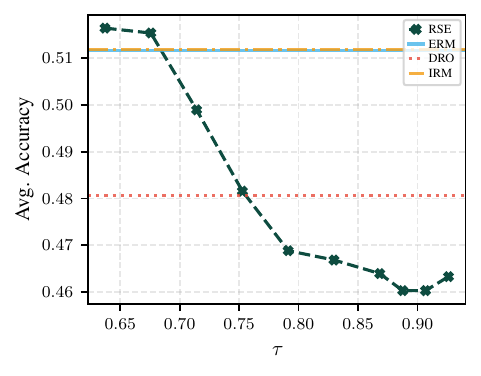}
            \vspace{0.1cm}
            \centerline{\scriptsize (b) No shift ($\alpha = 0$)}
        \end{minipage}
        \vspace{0.1cm}
        \caption{Sensitivity of RSE to aspiration level $\tau$ on SST-5.}
        \label{fig:tau_analysis}
    \end{minipage}
\end{figure}

\section{Conclusion, Limitations, and Future Work}
\label{sec:conclusion}

We introduced RSE, a probabilistic aggregation framework that secures pre-trained model ensembles against unpredictable distribution shifts by optimizing for \emph{fragility}. Evaluated across a diverse spectrum of shifts, RSE consistently prevented catastrophic failures and outperformed strong robust baselines while remaining the computationally lightest optimization-based method. Currently, RSE requires access to full predictive probability distributions. Extending it to black-box systems, where predictions can be represented as
one-hot vectors but confidence information is unavailable, remains an
important empirical direction. Additionally, adapting RSE for online sequential monitoring, dynamically updating moment matrices as $q_{\mathrm{ref}}$ drifts, is a critical direction for non-stationary environments. Finally, extending the quadratic risk surrogate beyond the Brier score to handle complex structured prediction tasks will further broaden RSE's applicability.

\section*{Acknowledgements}

\textbf{Funding Transparency Statement.} This work was supported by the Scientific and Technological Research Council of T\"{u}rkiye (TUBITAK) under Grant 124E065; TUBITAK 2024 Incentive Award; by the Turkish Academy of Sciences Distinguished Young Scientist Award Program (TUBA-GEBIP-2023).

\textbf{Competing Interests Statement.} The authors declare no competing interests.

\bibliographystyle{unsrtnat}
\bibliography{references}

@inproceedings{lipton2018detecting,
  title={Detecting and correcting for label shift with black box predictors},
  author={Lipton, Zachary and Wang, Yu-Xiang and Smola, Alexander},
  booktitle={Proc. 35th International Conference on Machine Learning},
  year={2018},
  pages= {3122--3130}
}

@article{smith2006,
author = {Smith, James E. and Winkler, Robert L.},
title = {The Optimizer’s Curse: Skepticism and Postdecision Surprise in Decision Analysis},
journal = {Management Science},
volume = {52},
number = {3},
pages = {311-322},
year = {2006},


}

@article{namkoong2016,
 author = {Namkoong, Hongseok and Duchi, John C},
 journal = {Advances in Neural Information Processing Systems},
 title = {Stochastic Gradient Methods for Distributionally Robust Optimization with f-divergences},
 volume = {29},
 year = {2016},
 pages={2216 - 2224}
}

@Article{long2023,
journal={Operations Research},
author={Daniel Zhuoyu Long and Melvyn Sim and Minglong Zhou},
title={Robust Satisficing},
year={2023},
pages={61-82},
volume={71},
number={1},

}

@article{
saday2023robust,
title={Robust {Bayesian} Satisficing},
author={Artun Saday and Y. Cahit Y{\i}ld{\i}r{\i}m and Cem Tekin},
journal={Advances in Neural Information Processing Systems},
volume = {36},
year={2023},
pages={69253 - 69269}
}

@inproceedings{
Sagawa*2020Distributionally,
title={Distributionally Robust Neural Networks for Group Shifts: On the Importance of Regularization for Worst-Case Generalization},
author={Shiori Sagawa and Pang Wei Koh and Tatsunori B. Hashimoto and Percy Liang},
booktitle={Proc. 8th International Conference on Learning Representations},
year={2020},
}

@misc{mistral2025small24b,
  author       = {{Mistral AI}},
  title        = {Mistral Small 24B Instruct 2501},
  year         = {2025},
  month        = jan,
  url          = {https://mistral.ai/news/mistral-small-3/},
  publisher    = {Mistral AI},
}

@inproceedings{socher-etal-2013-recursive,
    title = "Recursive Deep Models for Semantic Compositionality Over a Sentiment Treebank",
    author = "Socher, Richard  and
      Perelygin, Alex  and
      Wu, Jean  and
      Chuang, Jason  and
      Manning, Christopher D.  and
      Ng, Andrew  and
      Potts, Christopher",
    booktitle = "Proc. 2013 Conference on Empirical Methods in Natural Language Processing",
    year = "2013",
    pages = "1631--1642"
}

@book{vapnik1998statistical,
  title={Statistical Learning Theory},
  author={Vapnik, Vladimir N.},
  year={1998},
  publisher={Wiley},
  address={New York}
}

@inproceedings{
li2025the,
title={The Devil Is in the Details: Tackling Unimodal Spurious Correlations for Generalizable Multimodal Reward Models},
author={Zichao Li and Xueru Wen and Jie Lou and Yuqiu Ji and Yaojie Lu and Xianpei Han and Debing Zhang and Le Sun},
booktitle={Proc. 42nd International Conference on Machine Learning},
year={2025},
pages={36281--36297},

}

@inproceedings{koh2021wilds,
  title={{WILDS}: A benchmark of in-the-wild distribution shifts},
  author={Koh, Pang Wei and Sagawa, Shiori and Marklund, Henrik and Xie, Sang Michael and Zhang, Marvin and Balsubramani, Akshay and Hu, Weihua and Yasunaga, Michihiro and Phillips, Richard Lanas and Gao, Irena and others},
  booktitle={Proc. 38th International Conference on Machine Learning},
  pages={5637--5664},
  year={2021}
}

@article{yan2024kl,
  title={On the {KL}-Divergence-based Robust Satisficing Model},
  author={Yan, Haojie and Zhou, Minglong and Guo, Jiayi},
  journal={arXiv preprint arXiv:2408.09157},
  year={2024}
}

@article{yao2022wild,
  title={Wild-time: A benchmark of in-the-wild distribution shift over time},
  author={Yao, Huaxiu and Choi, Caroline and Cao, Bochuan and Lee, Yoonho and Koh, Pang Wei W and Finn, Chelsea},
  journal={Advances in Neural Information Processing Systems},
  volume={35},
  year={2022},
  pages={10309 - 10324}
}

@article{arjovsky2019invariant,
  title={Invariant risk minimization},
  author={Arjovsky, Martin and Bottou, L{\'e}on and Gulrajani, Ishaan and Lopez-Paz, David},
  journal={arXiv preprint arXiv:1907.02893},
  year={2019}
}

@article{breiman96,
  author = {Breiman, Leo},
  journal = {Machine Learning},
  pages = {123-140},
  title = {Bagging Predictors},
  volume = {24},
  year = {1996}
}

@article{lakshminarayanan2017deep,
  title={Simple and scalable predictive uncertainty estimation using deep ensembles},
  author={Lakshminarayanan, Balaji and Pritzel, Alexander and Blundell, Charles},
  journal={Advances in Neural Information Processing Systems},
  volume={30},
  year={2017},
  pages={6405 - 6416}
}

@inproceedings{huang2017snapshot,
  title={Snapshot ensembles: Train 1, get {$\mathit{M}$} for free},
  author={Huang, Gao and Li, Yixuan and Pleiss, Geoff and Liu, Zhuang and Hopcroft, John E and Weinberger, Kilian Q},
  booktitle = {Proc. 5th International Conference on Learning Representations},
  year      = {2017}
}

@inproceedings{wortsman2022model,
  title={Model soups: averaging weights of multiple fine-tuned models improves accuracy without increasing inference time},
  author={Wortsman, Mitchell and Ilharco, Gabriel and Gadre, Samir Ya and Roelofs, Rebecca and Gontijo-Lopes, Raphael and Morcos, Ari S and Namkoong, Hongseok and Farhadi, Ali and Carmon, Yair and Kornblith, Simon and others},
  booktitle={Proc. 39th International Conference on Machine Learning},
  pages={23965--23998},
  year={2022}
}

@article{liu2024lost,
  title={Lost in the middle: How language models use long contexts},
  author={Liu, Nelson F and Lin, Kevin and Hewitt, John and Paranjape, Ashwin and Bevilacqua, Michele and Petroni, Fabio and Liang, Percy},
  journal={Transactions of the Association for Computational Linguistics},
  volume={12},
  pages={157--173},
  year={2024}
}

@inproceedings{ai2025beyond,
  title={Beyond majority voting: {LLM} aggregation by leveraging higher-order information},
  author={Ai, Rui and Pan, Yuqi and Simchi-Levi, David and Tambe, Milind and Xu, Haifeng},
  booktitle={Proc. 43rd International Conference on Machine Learning},
  year={2026},
  pages={1311-1340}
}

@article{zhang2025stochasticity,
  title={From Stochasticity to Signal: A {Bayesian} Latent State Model for Reliable Measurement with {LLMs}},
  author={Zhang, Yichi and Martinez, Ignacio},
  journal={arXiv preprint arXiv:2510.23874},
  year={2025}
}

@article{ovadia2019can,
  title={Can you trust your model's uncertainty? evaluating predictive uncertainty under dataset shift},
  author={Ovadia, Yaniv and Fertig, Emily and Ren, Jie and Nado, Zachary and Sculley, David and Nowozin, Sebastian and Dillon, Joshua and Lakshminarayanan, Balaji and Snoek, Jasper},
  journal={Advances in Neural Information Processing Systems},
  volume={32},
  year={2019},
  pages={14003 - 14014}
}

@article{
sun2023beyond,
title={Beyond Invariance: Test-Time Label-Shift Adaptation for Addressing "Spurious" Correlations},
author={Qingyao Sun and Kevin Patrick Murphy and Sayna Ebrahimi and Alexander D'Amour},
journal={Advances in Neural Information Processing Systems},
volume={36},
year={2023},
pages={23789 - 23812}
}

@article{
      li2024more,
      title={More agents is all you need},
      author={Li, Junyou and Zhang, Qin and Yu, Yangbin and Fu, Qiang and Ye, Deheng},
      journal={Transactions on Machine Learning Research},
      year={2024},
      url={https://openreview.net/forum?id=bgzUSZ8aeg},
      note={}
}

@article{sim2025analytics,
author = {Sim, Melvyn and Tang, Qinshen and Zhou, Minglong and Zhu, Taozeng},
title = {The Analytics of Robust Satisficing: Predict, Optimize, Satisfice, Then Fortify},
journal = {Operations Research},
volume = {73},
number = {5},
pages = {2708-2728},
year = {2025},
}

@misc{jiang2023mistral7b,
      title={Mistral 7B}, 
      author={Albert Q. Jiang and Alexandre Sablayrolles and Arthur Mensch and Chris Bamford and others},
      year={2023},
      eprint={2310.06825},
      archivePrefix={arXiv},
      primaryClass={cs.CL}, 
      url={https://mistral.ai/news/announcing-mistral-7b/}
}

@article{team2025gemma,
  title={Gemma 3 technical report},
  author={{Gemma Team} and Kamath, Aishwarya and Ferret, Johan and Pathak, Shreya and Vieillard, Nino and Merhej, Ramona and Perrin, Sarah and Matejovicova, Tatiana and Ram{\'e}, Alexandre and Rivi{\`e}re, Morgane and others},
  journal={arXiv preprint arXiv:2503.19786},
  year={2025}
}

@article{yang2025qwen3,
  title={Qwen3 technical report},
  author={Yang, An and Li, Anfeng and Yang, Baosong and Zhang, Beichen and Hui, Binyuan and Zheng, Bo and Yu, Bowen and Gao, Chang and Huang, Chengen and Lv, Chenxu and others},
  journal={arXiv preprint arXiv:2505.09388},
  year={2025}
}

@article{abdin2024phi,
  title={Phi-4 technical report},
  author={Abdin, Marah and Aneja, Jyoti and Behl, Harkirat and Bubeck, S{\'e}bastien and Eldan, Ronen and Gunasekar, Suriya and Harrison, Michael and Hewett, Russell J and Javaheripi, Mojan and Kauffmann, Piero and others},
  journal={arXiv preprint arXiv:2412.08905},
  year={2024}
}

@inproceedings{
qiao2025structureinformed,
title={Structure-informed Risk Minimization for Robust Ensemble Learning},
author={Fengchun Qiao and Yanlin Chen and Xi Peng},
booktitle={Proc. 42nd International Conference on Machine Learning},
year={2025},
pages= {50130-50141}
}

@article{sohoni2020no,
  title={No subclass left behind: Fine-grained robustness in coarse-grained classification problems},
  author={Sohoni, Nimit and Dunnmon, Jared and Angus, Geoffrey and Gu, Albert and R{\'e}, Christopher},
  journal={Advances in Neural Information Processing Systems},
  volume={33},
  year={2020},
  pages={19339 - 19352}
}

@article{roberts2022unsupervised,
  title={Unsupervised learning under latent label shift},
  author={Roberts, Manley and Mani, Pranav and Garg, Saurabh and Lipton, Zachary},
  journal={Advances in Neural Information Processing Systems},
  volume={35},
  year={2022},
  pages={18763 - 18778}
}

@article{
saday2025robust,
title={Robust Satisficing {Gaussian} Process Bandits Under Adversarial Attacks},
author={Artun Saday and Y. Cahit Y{\i}ld{\i}r{\i}m and Cem Tekin},
journal={Advances in Neural Information Processing Systems},
volume = {38},
year={2025},
pages={50381 - 50408}
}

@article{agarwal2025gpt,
  title={gpt-oss-120b \& gpt-oss-20b model card},
  author={Agarwal, Sandhini and Ahmad, Lama and Ai, Jason and Altman, Sam and Applebaum, Andy and Arbus, Edwin and Arora, Rahul K and Bai, Yu and Baker, Bowen and Bao, Haiming and others},
  journal={arXiv preprint arXiv:2508.10925},
  year={2025}
}

@misc{Falcon3,
    title = {The Falcon 3 family of Open Models},
    author = {{TII Team}},
    month = {December},
    year = {2024}
}

@misc{abdin2024phi3technicalreporthighly,
      title={Phi-3 Technical Report: A Highly Capable Language Model Locally on Your Phone}, 
      author={Marah Abdin and Jyoti Aneja and Hany Awadalla and Ahmed Awadallah and Ammar Ahmad Awan and others},
      year={2024},
      eprint={2404.14219},
      archivePrefix={arXiv},
      primaryClass={cs.CL},
      url={https://arxiv.org/abs/2404.14219},
}

@inproceedings{li-roth-2002-learning,
    title = "Learning Question Classifiers",
    author = "Li, Xin  and
      Roth, Dan",
    booktitle = "Proc. 19th International Conference on Computational Linguistics",
    year = "2002",
}

@article{zaidi2021neural,
  title={Neural ensemble search for uncertainty estimation and dataset shift},
  author={Zaidi, Sheheryar and Zela, Arber and Elsken, Thomas and Holmes, Chris C and Hutter, Frank and Teh, Yee W.},
  journal={Advances in Neural Information Processing Systems},
  volume={34},
  year={2021},
  pages={7898 - 7911}
}

@article{shen2022deep,
  title={Deep combinatorial aggregation},
  author={Shen, Yuesong and Cremers, Daniel},
  journal={Advances in Neural Information Processing Systems},
  volume={35},
  year={2022},
  pages={32299 - 32310}
}

@article{arpit2022ensemble,
  title={Ensemble of averages: Improving model selection and boosting performance in domain generalization},
  author={Arpit, Devansh and Wang, Huan and Zhou, Yingbo and Xiong, Caiming},
  journal={Advances in Neural Information Processing Systems},
  volume={35},
  year={2022},
  pages={8265 - 8277}
}

@article{gardner2023benchmarking,
  title={Benchmarking distribution shift in tabular data with {TableShift}},
  author={Gardner, Josh and Popovic, Zoran and Schmidt, Ludwig},
  journal={Advances in Neural Information Processing Systems},
  volume={36},
  year={2023},
  pages={53385 - 53432}
}

@inproceedings{sun2020test,
  title={Test-time training with self-supervision for generalization under distribution shifts},
  author={Sun, Yu and Wang, Xiaolong and Liu, Zhuang and Miller, John and Efros, Alexei and Hardt, Moritz},
  booktitle={Proc. 37th International Conference on Machine Learning},
  pages={9229--9248},
  year={2020}
}

@inproceedings{
wang2021tent,
title={Tent: Fully Test-Time Adaptation by Entropy Minimization},
author={Dequan Wang and Evan Shelhamer and Shaoteng Liu and Bruno Olshausen and Trevor Darrell},
booktitle={Proc. 9th International Conference on Learning Representations},
year={2021},
}

@inproceedings{zhao2023pitfalls,
  title={On pitfalls of test-time adaptation},
  author={Zhao, Hao and Liu, Yuejiang and Alahi, Alexandre and Lin, Tao},
  booktitle={Proc. 40th International Conference on Machine Learning},
  pages = {42058--42080},
  year={2023}
}

\newpage

\appendix
\onecolumn
\section{Implementation Details}
\label{app:implementation}
For our computational setup, the LLMs were executed using an NVIDIA A100 GPU server (80 GB VRAM), whereas the experimental evaluations were performed locally on a personal computer.
\subsection{RSE Parameters and Selection of the Aspiration Level (\texorpdfstring{$\tau$}{tau})}
\label{app:tau_selection}

In all reported experiments, the reference label distribution $\mathbf{q}_{\mathrm{ref}}$ was set exactly to the empirical marginal class distribution of the calibration training set. Because all evaluated datasets contain strictly positive sample counts for every class within their respective training splits, no artificial smoothing was required or applied.
The proximal weight $\rho$ and $\varepsilon_{\text{stop}}$ are chosen as 5 and $10^{-5}$, respectively.  The infinite series obtained from the Lagrange--B\"urmann expansion is approximated by retaining terms up to order \(N=8\). All approximate fragility estimates $\widehat{\kappa}_t$
remained strictly positive throughout the reported experiments. To determine the aspiration level $\tau$, we employ a data-driven heuristic that balances performance on the source distribution with robustness against worst-case class shifts. We first compute the standard Empirical Risk Minimization (ERM) solution, denoted as $\mathbf{w}_{\text{ERM}}$, on the training source distribution $P_{\text{train}}$. Using this baseline, we evaluate
the corresponding class-conditional risks $\{r(\mathbf{w}_{\mathrm{ERM}},y_j)\}_{j=1}^K$.
We then define the source risk and worst-class risk as
\[
R_{\mathrm{source}} := R(\mathbf{w}_{\mathrm{ERM}},\mathbf{q}_{\mathrm{ref}})
= \sum_{j=1}^K (q_{\mathrm{ref}})_j\, r(\mathbf{w}_{\mathrm{ERM}},y_j),
\qquad
R_{\mathrm{worst}} := \max_{j\in\{1,\dots,K\}} r(\mathbf{w}_{\mathrm{ERM}},y_j).
\]

The threshold $\tau$ is then defined as a linear interpolation between the source risk and the worst-case risk, controlled by a hyperparameter $\lambda \in [0, 1]$:
\begin{equation}
    \tau = R_{\text{source}} + \lambda \left( R_{\text{worst}} - R_{\text{source}} \right)
\end{equation}

The hyperparameter $\lambda$ determines the tolerance for relaxing the source risk constraint to gain robustness. In our experiments, we intentionally selected $\tau$ values that correspond to the maximum feasible values within the solvable range. Specifically, we set $\lambda = 0.70$ for SST-5, $\lambda = 0.62$ for TREC, $\lambda = 0.95$ for selective copying, $\lambda = 0.80$ for CivilComments, and $\lambda = 0.19$ for HuffPost. This strategy ensures the lowest possible fragility $\kappa$, thereby maximizing the ensemble's robustness against severe distribution shifts.

\textbf{How to Pick $\tau$ in Practice}: Higher values of $\tau$ typically reduce the fragility coefficient $\kappa$, 
but they also relax the nominal risk requirement. Therefore, picking $\tau$ is not a blind hyperparameter search, but a deterministic reflection of the deployment environment. Our mathematical bounds provide the exact minimum and maximum feasible values for $\tau$. Practitioners simply use an interpolation parameter $\lambda \in [0, 1]$ to select their desired risk profile:

\begin{itemize}
    \item \textbf{Low-Risk, Predictable Environments:} If the practitioner expects minimal distribution shift, they set $\tau$ close to the nominal ERM risk ($\lambda \approx 0$). This explicitly guarantees on-par performance at low shift regimes.
    \item \textbf{High-Risk, Volatile Environments:} If the practitioner expects severe shifts, the optimal strategy is to select a high value of $\tau$ ($\lambda \to 1$). In practice, the upper limit of $\lambda$ is dictated by the numerical stability of the inner fragility root. As $\tau$ approaches the worst-case risk, the fragility $\kappa$ approaches $0$. The practitioner can therefore increase $\lambda$ to the highest value before the exponential tilt adversary ($e^{r_j/\kappa}$) encounters numerical overflow limits, thereby guaranteeing the maximum computationally stable protection.
\end{itemize}

This approach is completely deterministic and automatically ensures the ensemble gains the optimal level of protection without breaking the numerical optimization. Crucially, it entirely eliminates the need for expensive hyperparameter tuning on a validation set.

\subsection{Hyperparameter Settings for the Baseline Algorithms}
\label{app:hyperparameters}

For the competing methods, we used the hyperparameter settings reported in their original papers or official implementations, and then kept these settings fixed across the corresponding experiments. In particular, we did not perform per-dataset or per-shift hyperparameter sweeps for other methods. We provide below complete hyperparameter settings for all methods to ensure reproducibility. In label shift experiments, we repeated the experiments $20$ times and reported the averaged results.

\paragraph{ERM.}
It optimizes ensemble weights by minimizing cross-entropy loss on the training set. We use a learning rate of $0.01$ with $500$ optimization iterations. The weights are parameterized via softmax to ensure simplex constraints.

\paragraph{Group DRO.} We implement Group DRO using exponentiated gradient ascent for the adversarial distribution $q$ and gradient descent for the ensemble weights $w$. The method runs for $200$ iterations, utilizing a step size of $\eta_q = 0.1$ for the $q$ update and a learning rate of $\eta_w = 0.1$ for the $w$ update. We perform $5$ inner optimization steps per iteration and optimize using the cross-entropy loss function.\paragraph{IRM.} For Invariant Risk Minimization, we treat each class as a distinct environment and apply the IRMv1 penalty. The model is trained for $200$ iterations using a learning rate of $0.01$ and a penalty weight set to $\lambda_{\text{IRM}} = 0.4$.

\paragraph{SRM.} For Structure-informed Risk Minimization, we construct the distributional graph utilizing 2-Wasserstein distances between class-conditional prediction distributions, employing an extended feature representation of concatenated per-model predictions. We determine node importance via closeness centrality and train for $200$ iterations with a KL regularization weight of $\lambda_{\text{reg}} = 1.0$, using a learning rate of $0.1$ for both the weights $w$ and the distribution $q$.

\paragraph{KL-RS.} For KL-Divergence-based Robust Satisficing, we employ a bisection search over the range $\lambda \in [10^{-5}, 10^{3}]$ to identify the optimal robustness parameter $\lambda^*$. We set the target loss factor to $\tau_{\text{factor}} = 1.1$ (yielding $\tau = 1.1 \times L_{\text{ERM}}$) and utilize a convergence threshold of $\epsilon = 10^{-3}$. The inner optimization process involves $300$ iterations with a learning rate of $0.05$.\paragraph{Non-optimization Baselines.} For the static methods, the \textbf{Uniform Ensemble} assigns equal weights $w_m = 1/M$ for all models $m \in [M]$. The \textbf{Accuracy-Weighted} approach assigns weights proportional to the validation accuracy of each model, such that $w_m \propto \text{Acc}_m$. Finally, the \textbf{Majority Vote} baseline utilizes hard voting derived from the $\arg\max$ predictions of each individual model.

\subsection{Dataset Splits and Partitioning}
\label{app:dataset_splits}

To ensure rigorous evaluation and reproducibility across our diverse experimental settings, we apply specific partitioning strategies depending on the nature of the distribution shift.

\paragraph{Adversarial Label Shift Datasets (SST-5 and TREC):} 
For the SST-5 and TREC datasets, we utilize their respective official training splits as our primary foundational data. Because our adversarial shift protocol requires synthesizing specific target shifts from a held-out pool, we partition this main official train split into three subsets:
\begin{itemize}
    \item \textbf{Test Pool:} 10\% of the primary split is reserved as the test pool (\texttt{TEST\_POOL\_FRACTION = 0.1}). We draw from this pool to synthesize the varying degrees of test-time label shift ($\alpha$).
    \item \textbf{Validation Set:} 10\% of the primary split is allocated for validation purposes (\texttt{VALIDATION\_FRACTION = 0.1}), which is utilized to identify the hardest class for the severe shift protocol.
    \item \textbf{Training Set:} The remaining 80\% of the primary split is used as the calibration training set to estimate the base class-conditional risks and compute the moment matrices for the ensemble.
\end{itemize}
Our SST-5 experiments utilize the official training split containing 8,544 instances. For TREC, we use the official train split consisting of 5,452 instances.

\paragraph{Natural Distribution Shift Datasets (CivilComments and HuffPost):}
For our evaluations targeting natural, real-world distribution shifts, we strictly adhere to the established benchmark splits from the WILDS framework to maintain comparability with prior literature. 
\begin{itemize}
    \item \textbf{CivilComments:} We use the official train and test splits provided by the WILDS benchmark to evaluate subpopulation shift. The dataset comprises 269,038 training samples and 133,782 test samples.
    \item \textbf{HuffPost:} We utilize the official temporal train and test splits provided by the WILDS benchmark, where the ensemble is calibrated on the earlier years (2012–2015, comprising 35,558 training samples) and evaluated on the subsequent years (2016–2018, comprising 23,103 test samples) to measure robustness against temporal drift.
\end{itemize}
\subsection{Probability Vector Extraction and Prompt Templates}
\label{app:prob_extraction}

To obtain a probability distribution $\mathbf{p}_m(x) \in \Delta_K$ for a given input $x$ and model $m$, we utilize a constrained log-probability scoring method. This approach bypasses the stochasticity of standard text generation and ensures valid, dense probability vectors even for smaller models that might otherwise hallucinate formats.

For each task, we map the $K$ class labels to a fixed set of identifiers $\mathcal{A} = \{A, B, \dots\}$. For every candidate identifier $y_k \in \mathcal{A}$, we construct a target continuation string $s_k$ that represents the strictly formatted answer expected from the model (e.g., \texttt{\{"answer": "A"\}}).

We compute the scalar score $z_k$ for class $k$ as the sum of the log-probabilities of the tokens in $s_k$, conditioned on the input prompt $x_{\text{prompt}}$:
\begin{equation}
    z_k = \sum_{t=1}^{|s_k|} \log P_{\theta}\left(s_k^{(t)} \mid x_{\text{prompt}}, s_k^{(1:t-1)}\right)
\end{equation}
where $s_k^{(t)}$ denotes the $t$-th token of the answer string. Once the raw scores $\mathbf{z} = [z_1, \dots, z_K]$ are computed for all $K$ classes, we apply a  softmax function to normalize them into a probability distribution:
\begin{equation}
    p_m(x)_k = \frac{\exp(z_k)}{\sum_{j=1}^{K} \exp(z_j)}
\end{equation}
This vector $\mathbf{p}_m(x)$ is then used directly as the input to the ensemble aggregation algorithms.

We employ a standard zero-shot formatting across all models. In Figure~\ref{fig:prompt_templates}, the specific prompt templates used for each dataset can be seen.

\textbf{A Note on Probability Calibration and the Brier Score.} 
Constrained log-probability scoring inherently sharpens output distributions, yielding overconfident vectors that distort absolute risk. Crucially, this empirical misalignment does not compromise our theoretical guarantees. The framework's mathematical foundation, specifically the positive semi-definiteness of the moment matrices $\mathbf{A}_j$ and the convergence of proximal updates, relies solely on the simplex constraint ($\mathbf{p}_m \in \Delta_K$) and the algebraic structure of the Brier score, making it entirely agnostic to perfect calibration. Miscalibration can change the numerical values of the Brier risks, but the
positive-semidefiniteness and optimization results require only that
$\mathbf p_m(x)\in\Delta_K$; they do not require perfectly calibrated
probabilities.

\begin{figure*}[htbp!] 
    \centering
    \footnotesize 
    \definecolor{boxgray}{gray}{0.95} 
    
    \noindent
    \fcolorbox{black}{boxgray}{%
    \begin{minipage}[t]{0.47\textwidth}
        \textbf{Dataset: SST-5 (Sentiment Analysis)} \\
        \hrulefill \\
        \textbf{[System Instruction]} \\
        You are an expert at sentiment analysis. Classify the user's text into one of five categories: very negative, negative, neutral, positive, or very positive. Select the correct answer. Respond with ONLY a JSON object in this exact format: \texttt{\{"answer": "A"\}} where the value is A, B, C, D, or E.
        
        \vspace{0.3em}
        \textbf{[User Input]} \\
        Review: \texttt{[Input Text]} \\
        Choices: \\
        A. very negative \\
        B. negative \\
        C. neutral \\
        D. positive \\
        E. very positive
        
        \vspace{0.3em}
        \textbf{Target:} \texttt{\{"answer": "E"\}}
    \end{minipage}%
    }
    \hfill
    \fcolorbox{black}{boxgray}{%
    \begin{minipage}[t]{0.47\textwidth}
        \textbf{Dataset: TREC (Question Classification)} \\
        \hrulefill \\
        \textbf{[System Instruction]} \\
        You are an expert at question classification. Classify the user's question into one of six categories: description and abstract concepts, entities, abbreviation, human beings, numeric values, or locations. Select the correct answer. Respond with ONLY a JSON object in this exact format: \texttt{\{"answer": "A"\}} where the value is A, B, C, D, E, or F.
        
        \vspace{0.3em}
        \textbf{[User Input]} \\
        Question: \texttt{[Input Question]} \\
        Choices: \\
        A. abbreviation \\
        B. description and abstract concepts \\
        C. entities \\
        D. human beings \\
        E. locations \\
        F. numeric values
        
        \vspace{0.3em}
        \textbf{Target:} \texttt{\{"answer": "F"\}}
    \end{minipage}%
    }

    \vspace{1em} 

    \noindent
    \fcolorbox{black}{boxgray}{%
    \begin{minipage}[t]{0.47\textwidth}
        \textbf{Dataset: CivilComments (Toxicity)} \\
        \hrulefill \\
        \textbf{[System Instruction]} \\
        You are an expert at detecting toxic comments online. Classify whether the following comment is toxic or non-toxic. A toxic comment is rude, disrespectful, or likely to make someone leave a discussion. Respond with ONLY: \texttt{\{"answer": "X"\}} where X is A or B.
        
        \vspace{0.3em}
        \textbf{[User Input]} \\
        Comment: \texttt{[Input Text]} \\
        Choices: \\
        A. non-toxic \\
        B. toxic
        
        \vspace{2.5em} 
        \textbf{Target:} \texttt{\{"answer": "B"\}}
    \end{minipage}%
    }
    \hfill
    \fcolorbox{black}{boxgray}{%
    \begin{minipage}[t]{0.47\textwidth}
        \textbf{Dataset: Selective Copying (Context Position Bias)} \\
        \hrulefill \\
        \textbf{[System Instruction]} \\
        Extract the single uppercase letter hidden in the sequence of numbers below.
        
        \vspace{0.3em}
        \textbf{[User Input --- Easy split: target at END]} \\
        Sequence: \texttt{[Noise Tokens] [Target Letter]} \\
        The hidden letter is: \texttt{[Target Letter]}
        
        \vspace{0.3em}
        \textbf{[User Input --- Hard split: target in MIDDLE]} \\
        Sequence: \texttt{[Noise Tokens] [Target Letter] [Noise Tokens]} \\
        The hidden letter is: \texttt{[Target Letter]}
        
        \vspace{0.3em}
        \textbf{Target:} \texttt{"X"}
    \end{minipage}%
    }

    \vspace{1em} 

    \noindent
    \fcolorbox{black}{boxgray}{%
    \begin{minipage}[t]{0.965\textwidth}
        \textbf{Dataset: HuffPost (News Categorization)} \\
        \hrulefill \\
        \textbf{[System Instruction]} \\
        You are an expert at news classification. Classify the user's headline into one of 9 categories. Select the correct answer. Respond with ONLY a JSON object in this exact format: \texttt{\{"answer": "A"\}} where the value is the corresponding letter.
        
        \vspace{0.3em}
        \textbf{[User Input]} \\
        Headline: \texttt{[Input Headline]} \\
        Choices: \\
        A. Black Voices \quad B. Business \quad C. Comedy \quad D. Entertainment \quad E. Science \quad F. Travel \quad G. Sports \quad H. Crime \quad I. Queer Voices 
        
        \vspace{0.3em}
        \textbf{Target:} \texttt{\{"answer": "A"\}}
    \end{minipage}%
    }

    \caption{Prompt templates used for the five evaluated datasets. The system instruction defines the task and response format, while the user input contains the instance and, where applicable, the answer choices. Target entries illustrate example outputs in the expected format.}
    \label{fig:prompt_templates}
\end{figure*}

\newpage
\raggedbottom
\section{Additional Experiments}
\label{app:additional_results}

In this section, we present comprehensive numerical results supplementing the main experimental evaluation. We provide granular performance breakdowns for both the adversarial distribution shift experiments (SST-5, TREC) and the natural distribution shift experiments (CivilComments, HuffPost).

\subsection{Label Shift Datasets}
\label{app:label_shift}

In this subsection, we present the extended results under minority target label shift for the SST-5 and TREC datasets. We evaluate the performance stability of RSE compared to baselines across the full spectrum of shift severities, denoted by the shift magnitude parameter $\alpha \in [0, 1]$. 
\begin{itemize}
    \item $\alpha = 0.00$: Represents the \textbf{In-Distribution} setting (test distribution matches training).
    \item $\alpha = 1.00$: Represents the \textbf{Worst-Case} shift, where the test set is composed entirely of the class that is hardest for the ensemble to predict.
\end{itemize}

\noindent\textbf{Statistical Significance.} For SST-5 and TREC, we average the five test-draw accuracies
within each randomized split and report the mean and sample
standard deviation across 20 splits. All methods use the same
test examples within each draw. For each baseline and shift
level, we apply a two-sided paired $t$-test to the 20
split-level accuracy differences from RSE.
The statistic is $t=\bar d/(s_d/\sqrt{20})$, where $\bar d$
and $s_d$ are the mean and sample standard deviation of these
differences, with 19 degrees of freedom.
No multiple-testing correction is applied.
Because the test uses paired differences, overlapping
standard deviations of individual methods do not preclude
a significant difference. These comparisons assess variability
across randomized splits of the fixed dataset.

\textbf{Extended Accuracy Breakdowns.} For complete transparency, Tables~\ref{tab:sst5_detailed} and \ref{tab:trec_detailed} supply the granular accuracy breakdowns for both datasets. These tables include not only the primary optimization algorithms but also simple non-optimization heuristics and the raw performance of all individual constituent models in the ensemble.

\begin{table*}[t]
\centering
\caption{SST-5 accuracy (\%) under increasing label shift.
Values are means $\pm$ sample standard deviations across 20 split
averages, each averaging five test draws.
Two-sided paired $t$-tests compare each baseline with RSE using these split averages. $^{*}$ and $^{**}$ denote $p<0.05$ and $p<0.01$, respectively, for each individual comparison. Bold indicates the highest mean accuracy.}
\label{tab:sst5_significance}

\resizebox{\textwidth}{!}{%
\begin{tabular}{c c c c c c c}
\toprule
$\alpha$ & RSE (Ours) & DRO & IRM & SRM & KL-RS & ERM \\
\midrule
0.00 & $48.0 \pm 1.2$
& $50.0 \pm 1.9^{**}$
& $\mathbf{52.9 \pm 1.7}^{**}$
& $52.2 \pm 1.5^{**}$
& $52.8 \pm 1.7^{**}$
& $\mathbf{52.9 \pm 1.7}^{**}$ \\

0.31 & $44.2 \pm 1.7$
& $44.5 \pm 2.1$
& $45.2 \pm 1.8^{**}$
& $45.1 \pm 1.8^{**}$
& $\mathbf{45.4 \pm 1.9}^{**}$
& $45.3 \pm 1.9^{**}$ \\

0.45 & $\mathbf{42.5 \pm 2.2}$
& $42.0 \pm 2.5^{*}$
& $41.6 \pm 2.1^{**}$
& $41.7 \pm 2.1^{**}$
& $41.9 \pm 2.1^{*}$
& $41.7 \pm 2.2^{**}$ \\

0.56 & $\mathbf{41.5 \pm 2.3}$
& $40.2 \pm 2.3^{**}$
& $39.2 \pm 2.0^{**}$
& $39.5 \pm 2.0^{**}$
& $39.6 \pm 2.1^{**}$
& $39.4 \pm 2.1^{**}$ \\

0.65 & $\mathbf{40.5 \pm 2.9}$
& $38.9 \pm 2.7^{**}$
& $37.1 \pm 2.5^{**}$
& $37.7 \pm 2.5^{**}$
& $37.5 \pm 2.5^{**}$
& $37.3 \pm 2.5^{**}$ \\

0.73 & $\mathbf{39.5 \pm 2.8}$
& $37.2 \pm 2.7^{**}$
& $34.7 \pm 2.6^{**}$
& $35.5 \pm 2.5^{**}$
& $35.3 \pm 2.6^{**}$
& $35.0 \pm 2.6^{**}$ \\

0.80 & $\mathbf{38.4 \pm 3.3}$
& $35.8 \pm 3.2^{**}$
& $32.7 \pm 2.8^{**}$
& $33.7 \pm 3.0^{**}$
& $33.4 \pm 3.0^{**}$
& $33.1 \pm 2.9^{**}$ \\

0.86 & $\mathbf{38.0 \pm 3.4}$
& $35.0 \pm 3.2^{**}$
& $31.6 \pm 3.1^{**}$
& $32.7 \pm 3.0^{**}$
& $32.3 \pm 3.1^{**}$
& $31.9 \pm 3.1^{**}$ \\

0.92 & $\mathbf{37.2 \pm 3.5}$
& $34.1 \pm 3.3^{**}$
& $30.2 \pm 3.2^{**}$
& $31.5 \pm 3.2^{**}$
& $30.9 \pm 3.2^{**}$
& $30.6 \pm 3.2^{**}$ \\

0.96 & $\mathbf{36.8 \pm 3.7}$
& $33.4 \pm 3.5^{**}$
& $29.1 \pm 3.3^{**}$
& $30.4 \pm 3.4^{**}$
& $29.9 \pm 3.4^{**}$
& $29.5 \pm 3.3^{**}$ \\

1.00 & $\mathbf{36.3 \pm 3.6}$
& $32.5 \pm 3.4^{**}$
& $28.0 \pm 3.3^{**}$
& $29.3 \pm 3.2^{**}$
& $28.8 \pm 3.3^{**}$
& $28.3 \pm 3.3^{**}$ \\
\bottomrule
\end{tabular}%
}
\end{table*}

\begin{table}[htbp]
    \centering
    \caption{\textbf{Significance testing on TREC
    under increasing minority target shift.}
    Accuracy (\%) is reported as mean $\pm$ sample standard
    deviation across 20 randomized splits, averaging five test
    draws within each split. Two-sided paired $t$-tests compare
    each baseline with RSE using these split averages.
    $^{*}$ and $^{**}$ denote $p<0.05$ and $p<0.01$,
    respectively, for each individual comparison.
    Bold indicates the highest mean accuracy.}
    \label{tab:trec_significance}
    \small
    \setlength{\tabcolsep}{4pt}
    \begin{tabular}{l c ccccc}
        \toprule
        \textbf{Shift ($\alpha$)} & \textbf{RSE (Ours)}
        & \textbf{DRO} & \textbf{IRM} & \textbf{SRM}
        & \textbf{KL-RS} & \textbf{ERM} \\
        \midrule
        0.00 & 77.6 $\pm$ 1.8 & 77.5 $\pm$ 1.6 & \textbf{78.6 $\pm$ 1.6}$^{**}$ & 78.3 $\pm$ 1.8$^{**}$ & 78.5 $\pm$ 1.7$^{**}$ & 78.5 $\pm$ 1.8$^{**}$ \\
        0.32 & \textbf{75.9 $\pm$ 2.7} & 75.2 $\pm$ 2.4$^{**}$ & 74.9 $\pm$ 2.5$^{**}$ & 75.1 $\pm$ 2.5$^{**}$ & 74.8 $\pm$ 2.5$^{**}$ & 74.9 $\pm$ 2.6$^{**}$ \\
        0.46 & \textbf{75.1 $\pm$ 3.1} & 74.0 $\pm$ 2.5$^{**}$ & 73.2 $\pm$ 2.8$^{**}$ & 73.5 $\pm$ 2.7$^{**}$ & 73.2 $\pm$ 2.8$^{**}$ & 73.2 $\pm$ 2.9$^{**}$ \\
        0.57 & \textbf{74.6 $\pm$ 3.5} & 73.4 $\pm$ 3.3$^{**}$ & 72.1 $\pm$ 3.6$^{**}$ & 72.6 $\pm$ 3.3$^{**}$ & 72.1 $\pm$ 3.6$^{**}$ & 72.2 $\pm$ 3.6$^{**}$ \\
        0.66 & \textbf{74.4 $\pm$ 3.7} & 72.9 $\pm$ 3.3$^{**}$ & 71.2 $\pm$ 3.8$^{**}$ & 71.9 $\pm$ 3.4$^{**}$ & 71.2 $\pm$ 3.8$^{**}$ & 71.4 $\pm$ 3.8$^{**}$ \\
        0.74 & \textbf{73.6 $\pm$ 4.1} & 71.9 $\pm$ 3.7$^{**}$ & 69.8 $\pm$ 4.0$^{**}$ & 70.5 $\pm$ 3.6$^{**}$ & 69.9 $\pm$ 4.0$^{**}$ & 70.1 $\pm$ 4.0$^{**}$ \\
        0.81 & \textbf{73.5 $\pm$ 4.2} & 71.7 $\pm$ 3.8$^{**}$ & 69.2 $\pm$ 4.3$^{**}$ & 70.0 $\pm$ 3.8$^{**}$ & 69.2 $\pm$ 4.3$^{**}$ & 69.4 $\pm$ 4.3$^{**}$ \\
        0.87 & \textbf{73.0 $\pm$ 4.4} & 71.2 $\pm$ 4.0$^{**}$ & 68.7 $\pm$ 4.5$^{**}$ & 69.6 $\pm$ 3.9$^{**}$ & 68.8 $\pm$ 4.4$^{**}$ & 68.9 $\pm$ 4.5$^{**}$ \\
        0.92 & \textbf{73.0 $\pm$ 4.5} & 71.1 $\pm$ 4.2$^{**}$ & 68.3 $\pm$ 4.9$^{**}$ & 69.3 $\pm$ 4.3$^{**}$ & 68.5 $\pm$ 4.7$^{**}$ & 68.6 $\pm$ 4.8$^{**}$ \\
        0.96 & \textbf{72.5 $\pm$ 4.9} & 70.6 $\pm$ 4.5$^{**}$ & 67.7 $\pm$ 4.9$^{**}$ & 68.6 $\pm$ 4.3$^{**}$ & 67.8 $\pm$ 4.7$^{**}$ & 68.0 $\pm$ 4.8$^{**}$ \\
        1.00 & \textbf{72.0 $\pm$ 5.0} & 69.9 $\pm$ 4.9$^{**}$ & 66.7 $\pm$ 5.2$^{**}$ & 67.7 $\pm$ 4.6$^{**}$ & 66.9 $\pm$ 5.1$^{**}$ & 67.1 $\pm$ 5.1$^{**}$ \\
        \bottomrule
    \end{tabular}
\end{table}

\begin{table}[htbp]
\centering
\caption{\textbf{SST-5}: Detailed accuracy (\%) for all methods and individual models across shift magnitudes. Bold indicates the best ensemble results at the displayed precision.}
\label{tab:sst5_detailed}
\vspace{0.1cm}
\resizebox{\textwidth}{!}{%
\begin{tabular}{l *{11}{c}}
\toprule
\multirow{2}{*}{\textbf{Method}} & \multicolumn{11}{c}{\textbf{Shift Magnitude ($\alpha$)}} \\
\cmidrule(lr){2-12}
 & \textbf{0.00} & \textbf{0.31} & \textbf{0.45} & \textbf{0.56} & \textbf{0.65} & \textbf{0.73} & \textbf{0.80} & \textbf{0.86} & \textbf{0.92} & \textbf{0.96} & \textbf{1.00} \\
\midrule
\multicolumn{12}{l}{\textbf{Optimization Baselines}} \\
RSE                 & 48.0 & 44.2 & \textbf{42.5} & \textbf{41.5} & \textbf{40.5} & \textbf{39.5} & \textbf{38.4} & \textbf{38.0} & \textbf{37.2} & \textbf{36.8} & \textbf{36.3} \\
DRO                 & 50.0 & 44.5 & 42.0 & 40.2 & 38.9 & 37.2 & 35.8 & 35.0 & 34.1 & 33.4 & 32.5 \\
IRM                 & \textbf{52.9} & 45.2 & 41.6 & 39.2 & 37.1 & 34.7 & 32.7 & 31.6 & 30.2 & 29.1 & 28.0 \\
SRM                 & 52.2 & 45.1 & 41.7 & 39.5 & 37.7 & 35.5 & 33.7 & 32.7 & 31.5 & 30.4 & 29.3 \\
KL-RS               & 52.8 & \textbf{45.4} & 41.9 & 39.6 & 37.5 & 35.3 & 33.4 & 32.3 & 30.9 & 29.9 & 28.8 \\
ERM                 & \textbf{52.9} & 45.3 & 41.7 & 39.4 & 37.3 & 35.0 & 33.1 & 31.9 & 30.6 & 29.5 & 28.3 \\
\midrule
\multicolumn{12}{l}{\textbf{Non-Optimization Baselines}} \\
Uniform             & 52.1 & 43.8 & 39.8 & 37.2 & 34.8 & 32.4 & 30.2 & 28.9 & 27.4 & 26.2 & 25.1 \\
Majority Vote       & 51.1 & 43.2 & 39.4 & 36.8 & 34.6 & 32.2 & 30.2 & 29.0 & 27.5 & 26.3 & 25.3 \\
Acc. Weighted Model & 52.7 & 43.9 & 39.8 & 37.1 & 34.6 & 32.1 & 29.9 & 28.5 & 26.9 & 25.7 & 24.5 \\
\midrule
\multicolumn{12}{l}{\textbf{Individual Models}} \\
GPT-OSS-20b         & 48.8 & 44.6 & 42.6 & 41.3 & 40.2 & 38.9 & 37.8 & 37.2 & 36.4 & 35.9 & 35.2 \\
Qwen3-4b            & 44.7 & 37.0 & 33.2 & 30.6 & 28.4 & 26.1 & 24.3 & 22.9 & 21.4 & 20.6 & 19.3 \\
Falcon3-7b-It       & 48.5 & 41.3 & 37.9 & 35.7 & 33.6 & 31.7 & 30.2 & 29.0 & 27.4 & 26.5 & 25.6 \\
Phi-4-15b          & 51.3 & 40.4 & 35.4 & 31.7 & 28.5 & 25.5 & 23.1 & 21.2 & 19.3 & 17.8 & 16.3 \\
Mistral-7b-It      & 39.4 & 37.9 & 37.5 & 37.2 & 36.8 & 36.4 & 35.9 & 36.1 & 35.7 & 35.3 & 35.3 \\
Gemma-3-12b        & 51.7 & 44.9 & 41.5 & 39.2 & 37.4 & 35.4 & 33.6 & 32.9 & 31.2 & 30.6 & 29.4 \\
Mistral-Small-24b  & 52.9 & 44.8 & 41.3 & 38.5 & 36.3 & 34.0 & 31.9 & 30.7 & 29.2 & 28.0 & 26.9 \\
Phi-3.5-mini-It    & 51.8 & 42.3 & 38.1 & 34.9 & 32.2 & 29.5 & 27.5 & 25.9 & 24.0 & 22.7 & 21.6 \\
\bottomrule
\end{tabular}%
}
\end{table}

\begin{table}[htbp]
\centering
\caption{\textbf{TREC}: Detailed accuracy breakdown for all methods and individual models across shift magnitudes.}
\label{tab:trec_detailed}
\vspace{0.1cm}
\resizebox{\textwidth}{!}{%
\begin{tabular}{l *{11}{c}}
\toprule
\multirow{2}{*}{\textbf{Method}} & \multicolumn{11}{c}{\textbf{Shift Magnitude ($\alpha$)}} \\
\cmidrule(lr){2-12}
 & \textbf{0.00} & \textbf{0.32} & \textbf{0.46} & \textbf{0.57} & \textbf{0.66} & \textbf{0.74} & \textbf{0.81} & \textbf{0.87} & \textbf{0.92} & \textbf{0.96} & \textbf{1.00} \\
\midrule
\multicolumn{12}{l}{\textbf{Optimization Baselines}} \\
RSE                 & 77.6 & \textbf{75.9} & \textbf{75.1} & \textbf{74.6} & \textbf{74.4} & \textbf{73.6} & \textbf{73.5} & \textbf{73.0} & \textbf{73.0} & \textbf{72.5} & \textbf{72.0} \\
DRO & 77.5 & 75.2 & 74.0 & 73.4 & 72.9
& 71.9 & 71.7
& 71.2 & 71.1
& 70.6 & 69.9 \\
IRM                 & \textbf{78.6} & 74.9 & 73.2 & 72.1 & 71.2 & 69.8 & 69.2 & 68.7 & 68.3 & 67.7 & 66.7 \\
SRM & 78.3 & 75.1 & 73.5 & 72.6 & 71.9 & 70.5
& 70.0 & 69.6 & 69.3 & 68.6 & 67.7 \\
KL-RS               & 78.5 & 74.8 & 73.2 & 72.1 & 71.2 & 69.9 & 69.2 & 68.8 & 68.5 & 67.8 & 66.9 \\
ERM                 & 78.5 & 74.9 & 73.2 & 72.2 & 71.4 & 70.1 & 69.4 & 68.9 & 68.6 & 68.0 & 67.1 \\
\midrule
\multicolumn{12}{l}{\textbf{Non-Optimization Baselines}} \\
Uniform             & 77.6 & 72.8 & 70.4 & 69.0 & 67.6 & 66.0 & 65.2 & 64.4 & 63.7 & 62.7 & 61.7 \\
Majority Vote       & 76.6 & 71.8 & 69.3 & 68.1 & 66.6 & 65.0 & 64.2 & 63.4 & 62.7 & 61.7 & 60.8 \\
Acc. Weighted Model & 77.7 & 72.9 & 70.6 & 69.2 & 67.9 & 66.3 & 65.5 & 64.8 & 64.1 & 63.1 & 62.1 \\
\midrule
\multicolumn{12}{l}{\textbf{Individual Models}} \\
GPT-OSS-20b         & 70.6 & 70.5 & 70.3 & 70.7 & 70.6 & 70.2 & 70.4 & 70.2 & 70.6 & 70.4 & 70.0 \\
Qwen3-4b            & 69.8 & 63.6 & 60.5 & 59.0 & 56.9 & 55.3 & 54.3 & 53.3 & 52.1 & 51.3 & 50.3 \\
Falcon3-7b-It       & 69.8 & 61.5 & 57.6 & 54.7 & 52.7 & 50.1 & 48.4 & 46.8 & 45.7 & 44.3 & 43.3 \\
Phi-4-15b           & 77.2 & 74.1 & 72.5 & 71.9 & 70.9 & 69.6 & 69.3 & 69.0 & 68.5 & 67.7 & 67.2 \\
Mistral-7b-It       & 62.0 & 54.6 & 51.1 & 48.9 & 46.6 & 45.0 & 43.5 & 42.1 & 40.7 & 39.8 & 38.8 \\
Gemma-3-12b         & 67.4 & 63.2 & 61.1 & 60.0 & 58.9 & 57.0 & 56.5 & 55.9 & 55.2 & 54.3 & 53.8 \\
Mistral-Small-24b   & 76.5 & 74.1 & 72.8 & 72.2 & 71.7 & 70.4 & 69.9 & 69.8 & 69.6 & 69.0 & 68.3 \\
Phi-3.5-mini-It     & 71.2 & 59.6 & 54.1 & 51.1 & 47.2 & 44.4 & 42.5 & 39.8 & 38.3 & 36.2 & 34.9 \\
\bottomrule
\end{tabular}%
}
\end{table}

\subsection{Subpopulation Shift: CivilComments}
\label{app:civil_comments}

We use $B=1000$ paired bootstrap resamples of the fixed
CivilComments test set. Each resample contains the original
number of observations sampled with replacement, with identical
sampled indices used for all methods and ensemble weights held
fixed. Overall accuracy and WGA are recomputed in each resample;
WGA is the minimum accuracy over groups present in that resample.
We report bootstrap means and standard deviations.
For each baseline, the one-sided bootstrap tail fraction is
$\widehat p_{\mathrm{tail}}
=B^{-1}\sum_{b=1}^{B}
\mathbf{1}\{T_b^{\mathrm{baseline}}\ge T_b^{\mathrm{RSE}}\}$,
where $T$ is the evaluated metric.
This provides an approximate bootstrap comparison conditional
on the fitted ensembles; no multiple-testing correction is applied.

\noindent\textbf{Statistical Significance.}{
Table~\ref{tab:civilcomments_significance} reports bootstrap
mean overall accuracy and WGA. RSE outperforms all five
optimization baselines in both metrics, with one-sided
bootstrap tail fractions below $0.01$. Its mean WGA exceeds
DRO's by 2.0 percentage points, while also achieving higher
overall accuracy.
}

\textbf{Detailed Subpopulation Breakdown.} For granular transparency, Table~\ref{tab:civil_comments_subpop} details the accuracy across the eight specific demographic groups. RSE consistently dominates or matches the best performance across almost all individual groups, demonstrating broad robustness across diverse subpopulations rather than over-optimizing for a single demographic.

\begin{table}[htbp]
    \centering
    \caption{{
CivilComments performance under subpopulation shift.
Values are bootstrap mean accuracies (\%) $\pm$ bootstrap
standard deviations from $B=1000$ paired resamples.
$^{*}$ and $^{**}$ indicate one-sided bootstrap tail fractions
$\widehat p_{\mathrm{tail}}<0.05$ and $<0.01$, respectively.
Bold indicates the highest mean.
}}
    \label{tab:civilcomments_significance}
    \small
    \begin{tabular}{l cc}
        \toprule
        \textbf{Method} & \textbf{Overall Accuracy} & \textbf{Worst-Group Accuracy (WGA)} \\
        \midrule
        \textbf{RSE (Ours)} & \textbf{72.1 $\pm$ 0.1} & \textbf{59.7 $\pm$ 0.7} \\
        DRO                 & 70.6 $\pm$ 0.1$^{**}$   & 57.7 $\pm$ 0.8$^{**}$ \\
        IRM                 & 69.8 $\pm$ 0.1$^{**}$   & 56.5 $\pm$ 0.8$^{**}$ \\
        SRM                 & 68.8 $\pm$ 0.1$^{**}$   & 55.1 $\pm$ 0.8$^{**}$ \\
        KL-RS               & 69.6 $\pm$ 0.1$^{**}$   & 56.3 $\pm$ 0.8$^{**}$ \\
        ERM                 & 70.1 $\pm$ 0.1$^{**}$   & 56.8 $\pm$ 0.8$^{**}$ \\
        \bottomrule
    \end{tabular}
\end{table}

\begin{table*}[htbp]
\centering
\caption{\textbf{CivilComments:} Detailed accuracy breakdown across subpopulation shifts for specific demographic groups.}
\label{tab:civil_comments_subpop}
\setlength{\tabcolsep}{5pt}
\begin{tabular}{lcccccccc}
\toprule
\textbf{Method} & \textbf{Black} & \textbf{Christian} & \textbf{LGBTQ} & \textbf{Muslim} & \textbf{White} & \textbf{Female} & \textbf{Male} & \textbf{Other Rel.} \\
\midrule
\multicolumn{9}{l}{\textbf{Optimization Baselines}} \\
RSE & \textbf{65.4} & \textbf{77.2} & \textbf{68.0} & 60.8 & \textbf{59.7} & \textbf{65.6} & \textbf{73.7} & \textbf{67.1} \\
DRO & 64.3 & 76.2 & 66.3 & 59.1 & 57.8 & 64.2 & 72.2 & 64.8 \\
IRM & 63.5 & 75.4 & 65.5 & 57.8 & 56.5 & 63.1 & 71.5 & 63.8 \\
SRM & 62.8 & 74.3 & 64.2 & 56.9 & 55.1 & 62.1 & 70.5 & 61.8 \\
KL-RS & 63.3 & 75.1 & 65.0 & 57.8 & 56.3 & 62.9 & 71.2 & 63.5 \\
ERM & 63.8 & 75.7 & 65.7 & 58.3 & 56.8 & 63.5 & 71.7 & 64.1 \\
\midrule
\multicolumn{9}{l}{\textbf{Individual Models}} \\
Falcon3-7b-It & 63.9 & 75.1 & 64.9 & 57.7 & 56.0 & 64.5 & 71.5 & 61.1 \\
Gemma3-12b-It & 59.0 & 71.2 & 61.7 & 52.6 & 50.9 & 59.0 & 66.0 & 55.3 \\
GPT-OSS-20b & 47.6 & 55.2 & 47.6 & 38.7 & 39.5 & 45.7 & 57.0 & 37.9 \\
Mistral-24b & 64.4 & 76.2 & 65.7 & 61.2 & 57.8 & 64.2 & 71.0 & 67.0 \\
Mistral-7b & 50.3 & 59.5 & 51.2 & 42.9 & 42.0 & 47.1 & 55.9 & 46.8 \\
Phi-3.5-Mini & 65.2 & 75.5 & 66.8 & \textbf{62.0} & 58.6 & 65.2 & 73.3 & 67.0 \\
Phi-4 & \textbf{65.4} & 75.1 & 66.0 & 57.7 & \textbf{59.7} & 64.8 & 72.6 & 64.0 \\
Qwen3-4b & 61.7 & 69.4 & 63.1 & 58.1 & 54.4 & 60.6 & 64.8 & 62.7 \\
\bottomrule
\end{tabular}
\end{table*}

\subsection{Temporal Shift: HuffPost}
\label{app:huffpost}

In this subsection, we evaluate robustness against temporal distribution shifts using the HuffPost dataset. Models are trained on articles published between 2012 and 2015, and evaluated on subsequent years (2016--2018). As with the subpopulation shift analysis, we employ non-parametric bootstrap resampling ($B=1000$ iterations) on the test predictions to rigorously compute statistical significance over the fixed temporal split.

\textbf{Statistical Significance.} Table~\ref{tab:huffpost_significance} presents the bootstrapped Overall Accuracy and Worst-Year Accuracy (WYA). RSE yields statistically significant improvements across both metrics. Notably, RSE maintains a definitive edge ($p < 0.05$) in worst-year performance even against strong invariant-learning baselines like IRM and SRM, while vastly outperforming DRO ($p < 0.01$). This confirms that explicitly optimizing for ensemble stability translates directly into resilience against chronological data drift.

\textbf{Detailed Temporal Breakdown.} For granular transparency, Table~\ref{tab:huffpost_temporal} details the accuracy for each specific year in the test phase. RSE shows consistent temporal resilience, achieving the best absolute performance in every single evaluation year (2016, 2017, and 2018) compared to all optimization baselines and individual constituent models.

\begin{table}[htbp]
    \centering
    \caption{\textbf{Significance testing on HuffPost under temporal shift.} We report mean accuracy (\%) $\pm$ standard deviation obtained via $B=1000$ bootstrap resamples of the test set (2016--2018). The highest accuracy for each metric is marked in \textbf{bold}. Statistical significance against our proposed method (RSE) is determined via a one-sided bootstrap hypothesis test: $^{*}$ denotes $p < 0.05$ and $^{**}$ denotes $p < 0.01$. RSE significantly outperforms all baselines across both Average Yearly Accuracy (AYA) and Worst-Year Accuracy (WYA) metrics.}
    \label{tab:huffpost_significance}
    \small
    \begin{tabular}{l cc}
        \toprule
        \textbf{Method} & \textbf{Average Yearly Accuracy (AYA)} & \textbf{Worst-Year Accuracy (WYA)} \\
        \midrule
        \textbf{RSE (Ours)} & \textbf{78.8 $\pm$ 0.3} & \textbf{77.2 $\pm$ 0.4} \\
        DRO                 & 76.3 $\pm$ 0.3$^{**}$   & 75.5 $\pm$ 0.4$^{**}$ \\
        IRM                 & 78.4 $\pm$ 0.3$^{**}$   & 76.9 $\pm$ 0.4$^{*}$ \\
        SRM                 & 78.4 $\pm$ 0.3$^{**}$   & 76.9 $\pm$ 0.4$^{*}$ \\
        KL-RS               & 78.1 $\pm$ 0.3$^{**}$   & 76.6 $\pm$ 0.4$^{**}$ \\
        ERM                 & 78.1 $\pm$ 0.3$^{**}$   & 76.7 $\pm$ 0.4$^{**}$ \\
        \bottomrule
    \end{tabular}
\end{table}

\begin{table}[htbp]
\centering
\caption{\textbf{HuffPost:} Detailed accuracy breakdown across temporal shifts (2016--2018) for models trained on 2012--2015 data.}
\label{tab:huffpost_temporal}
\vspace{0.1cm}
\setlength{\tabcolsep}{12pt}
\begin{tabular}{lccc}
\toprule
\textbf{Method} & \textbf{2016} & \textbf{2017} & \textbf{2018} \\
\midrule
\multicolumn{4}{l}{\textbf{Optimization Baselines}} \\
RSE & \textbf{77.2} & \textbf{78.1} & \textbf{81.0} \\
DRO & 76.1 & 75.6 & 77.2 \\
IRM & 77.0 & 77.5 & 80.6 \\
SRM & 77.0 & 77.7 & 80.5 \\
KL-RS & 76.7 & 77.3 & 80.4 \\
ERM & 76.8 & 77.3 & 80.3 \\
\midrule
\multicolumn{4}{l}{\textbf{Individual Models}} \\
Falcon3-7b-It & 69.2 & 66.0 & 70.5 \\
Gemma3-12b-It & 75.0 & 74.4 & 76.8 \\
GPT-OSS-20b & 71.6 & 72.1 & 75.4 \\
Mistral-24b & 76.9 & 77.7 & 80.0 \\
Mistral-7b-It & 70.2 & 67.0 & 69.1 \\
Phi-3.5-Mini-It & 71.5 & 70.9 & 74.7 \\
Phi-4-15b & 71.6 & 69.7 & 71.9 \\
Qwen3-4b & 72.4 & 74.1 & 76.0 \\
\bottomrule
\end{tabular}
\end{table}

\subsection{Additional Aspiration Level Sensitivity Analysis}
\label{app:tau_sensitivity}

In this section, we extend the sensitivity analysis of the aspiration level $\tau$ to the TREC dataset to verify the generalizability of the observations made on SST-5. 

\subsubsection{TREC Dataset}
\begin{figure}[h]
    \centering
    \begin{minipage}{0.48\textwidth}
        \centering
        \includegraphics[width=\linewidth]{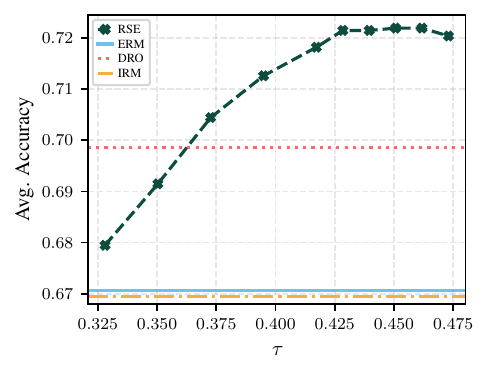}
        \centerline{\small (a) Severe shift ($\alpha = 1$)}
    \end{minipage}
    \hfill
    \begin{minipage}{0.48\textwidth}
        \centering
        \includegraphics[width=\linewidth]{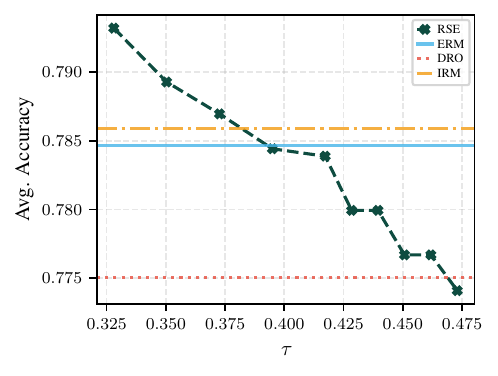}
        \centerline{\small (b) No shift ($\alpha = 0$)}
    \end{minipage}
    \caption{Sensitivity of RSE to aspiration level $\tau$ on the \textbf{TREC} dataset. (a) Performance under severe shift. (b) Performance under no shift.}
    \label{fig:tau_analysis_trec}
\end{figure}

Figure~\ref{fig:tau_analysis_trec} illustrates the performance on TREC. Similar to the SST-5 results, we observe a trade-off governed by $\tau$. Under severe shift (Figure~\ref{fig:tau_analysis_trec}a), increasing $\tau$ generally improves robustness. Conversely, in the absence of distribution shift (Figure~\ref{fig:tau_analysis_trec}b), aggressive constraints (higher $\tau$) can lead to a slight degradation in in-distribution accuracy.

\subsection{Sensitivity to Model-Pool Composition}
\label{app:model_pool_sensitivity}

We examine whether the performance of RSE depends on the size or composition
of the constituent-model pool. Starting from the original eight-model
ensemble used in the main experiments, we construct five alternative pool
settings while retaining the same label-shift evaluation protocol.

\paragraph{Pool construction.}
For \textbf{leave-one-out}, each of the eight original predictors is removed
once, producing eight seven-model pools. For \textbf{random subsets}, we
consider pool sizes from 2 through 7 and sample up to eight distinct subsets
at each size. These settings test sensitivity to both pool size and the
particular combination of available predictors.

For the \textbf{correlated} setting, we identify the predictor with the
highest validation accuracy and append four near-duplicate copies.
Each copy is formed by adding independent Gaussian noise to the model's log-probabilities, followed by
renormalization. This preserves predictions close to those of the original
model while substantially increasing redundancy within the pool.

For the \textbf{uncalibrated} setting, we append four temperature-scaled
predictors using
\[
T\in\{0.15,0.25,4.0,8.0\}.
\]
Given a probability vector $\mathbf p$, temperature scaling is applied as
\[
\widetilde{\mathbf p}
=
\operatorname{softmax}
\left(
\frac{\log \mathbf p}{T}
\right).
\]
Since all temperatures are positive, this transformation preserves the
predicted class while altering the confidence of the probability vector.
Temperatures below one make predictions more concentrated, whereas
temperatures above one make them less concentrated.

For the \textbf{weak-model} setting, we use five additional real LLMs that
are available for both SST-5 and TREC but are not members of the original
eight-model pools. These are Falcon3-10B, Llama-3.1-8B, Phi-3-mini,
Qwen3-8B, and Qwen3-14B. We rank the five candidates by their mean
held-out validation accuracy across the experimental splits and append the
three lowest-performing models to the original pool, resulting in an
eleven-model ensemble. The target class used to construct the label shift
is determined from the original eight-model pool and is therefore unchanged
when the weak models are added.

All variants use the same underlying train, validation, and shifted-test
construction as the corresponding main experiments. The reported
scenario-level values average performance across the evaluated shift
severities and the relevant pool configurations.

\begin{table}[htbp]
\centering
\caption{\textbf{Sensitivity to model-pool composition on SST-5.}
Mean accuracy (\%) across the full shift-severity spectrum. We modify the
original eight-model pool through leave-one-out removal, random subsets,
or the addition of correlated, uncalibrated, or weak predictors.
The highest mean accuracy in each row is in \textbf{bold}.}
\label{tab:model_pool_sensitivity_sst5}
\small
\setlength{\tabcolsep}{4.2pt}
\begin{tabular}{lcccccc}
\toprule
\textbf{Scenario} &
\textbf{ERM} &
\textbf{IRM} &
\textbf{DRO} &
\textbf{KL-RS} &
\textbf{SRM} &
\textbf{RSE} \\
\midrule
Original (8 Models) & 36.9 & 36.6 & 38.6 & 37.1 & 37.3 & \textbf{40.3} \\
Leave-one-out       & 36.4 & 36.3 & 38.1 & 36.5 & 37.1 & \textbf{39.4} \\
Random Subset       & 35.0 & 35.0 & 36.0 & 35.1 & 35.4 & \textbf{37.2} \\
+ Correlated        & 36.7 & 36.6 & 38.5 & 37.0 & 37.1 & \textbf{40.2} \\
+ Uncalibrated      & 35.2 & 35.2 & 37.8 & 35.5 & 36.4 & \textbf{38.5} \\
+ Weak Models       & 36.9 & 36.5 & 38.7 & 37.1 & 38.4 & \textbf{40.2} \\
\bottomrule
\end{tabular}
\end{table}
As shown in Table~\ref{tab:model_pool_sensitivity_sst5}, the SST-5 results
are consistent with the TREC results in
Table~\ref{tab:model_pool_sensitivity_trec}. RSE attains the highest mean
accuracy in every evaluated pool configuration, including substantial
reductions in pool size and the addition of redundant, miscalibrated, or
weak predictors. These results indicate that the gains of RSE are not tied
to a particular choice of the original eight constituent models.

\subsection{Input Shift: Selective Copying}
\label{app:selective_copying}

Real-world sequence-processing and retrieval-augmented systems often suffer from the ``Lost in the Middle'' phenomenon, where pre-trained models attend poorly to information positioned in the center of extended input contexts \citep{liu2024lost}. To formalize this vulnerability against input shifts, we constructed a synthetic Selective Copying extraction task where models must identify a single target class hidden within a sequence of 500 noise tokens. We calibrated the ensemble on 500 input sequences where the target information was located at the \emph{end}. We then evaluated the framework on a shifted test set of 500 sequences where the critical context was buried in the \emph{exact middle}.

\textbf{Model Selection for Selective Copying.} Across all adversarial and natural distribution shift experiments, we utilize a fixed suite of 8 models (including Gemma-3-12B-Instruct). However, evaluating input shift via Selective Copying requires constituent models that actually fail to positional sequence bias. Gemma-3-12B-Instruct demonstrates an unusually high native resilience to extraction failures within 500-token noise sequences. Including it masks the aggregation dynamics we aim to isolate, as the task would merely measure the performance of a single natively robust model. To construct a rigorous testbed where constituent models natively suffer from middle-context extraction failures, we substituted Gemma-3-12B-Instruct with Llama-3.1-8B-Instruct.

\textbf{Results.} RSE outperformed ERM by \textbf{+6.6\%} and DRO by \textbf{+5.0\%}, effectively mitigating position-induced extraction failures. Non-optimization baselines collapse because they cannot selectively downweight the majority of individual models that exploit the positional bias present in the calibration set.

\textbf{Statistical Significance.} Because the evaluation relies on a single fixed, shifted test set, we utilize non-parametric bootstrap resampling ($B=1000$ iterations) to estimate variance and statistical significance. Table~\ref{tab:sc_significance} details the bootstrapped accuracy across all baselines.

\subsection{Runtime and Memory Scaling}
\label{app:runtime_memory_scaling}

We further evaluate the computational scaling of RSE as the number of
constituent models $M$ and the number of classes $K$ increase. These
experiments are controlled computational stress tests based on synthetic
probability tensors. Their purpose is to isolate the aggregation cost of
RSE rather than to evaluate predictive performance on an additional
classification benchmark.

\paragraph{Experimental setup.}
We generate prediction tensors of shape $M\times N\times K$ with
approximately balanced class labels. For each model, logits are initialized
with independent Gaussian noise and the logit corresponding to the true
class is increased by a model-specific signal sampled uniformly from
$[0.8,2.2]$. We additionally introduce a small model-specific class bias
before applying a numerically stable softmax. This produces diverse,
non-degenerate probability vectors while allowing $M$ and $K$ to be varied
independently.

We consider two scaling experiments. For model-count scaling, we fix
$N=10{,}000$ and $K=10$ and vary
\[
M\in\{2,4,8,16,32,64\}.
\]
For class-count scaling, we fix $N=10{,}000$ and $M=16$ and report
\[
K\in\{5,10,20,50,100\}.
\]
Each configuration is repeated three times with independent random seeds.

We separately measure the one-time construction of the class-conditional
moment statistics
$\{\mathbf A_j,\mathbf b_j\}_{j=1}^{K}$ and the subsequent RSE weight
optimization using these cached quantities. Synthetic tensor generation is
excluded from all timing measurements, as is base-model inference, which is
a shared offline cost in the real-data experiments. The benchmark uses the
same project implementation of the RSE moment computation and proximal
solver as the main experiments. We use a Lagrange--B\"urmann truncation
order and solver configuration consistent with the RSE implementation,
with aspiration interpolation fraction $0.1$, maximum iteration count
$2000$, convergence tolerance $10^{-4}$, and proximal stepsize $0.2$.

The memory comparison reports two quantities. ``Stats'' is the exact memory
occupied by the cached matrices and vectors
$\{\mathbf A_j,\mathbf b_j\}_{j=1}^{K}$, which scales as
$O(M^2K)$. ``Cache'' is the memory required to retain the complete
$M\times N\times K$ prediction tensor together with the labels. Thus, the
reported memory values measure the objects that must be retained for the
corresponding computation rather than process-level memory of the Python
runtime.

\begin{table}[htbp]
\centering
\caption{\textbf{RSE runtime and memory scaling with the number of models.}
The experiment fixes $N=10{,}000$ and $K=10$ and varies $M$. Runtime is
reported as mean $\pm$ standard deviation over three runs. ``Moment''
denotes the one-time construction of
$\{\mathbf A_j,\mathbf b_j\}_{j=1}^{K}$ and ``Opt.'' denotes the subsequent
RSE optimization. Times exclude synthetic tensor generation and base-model
inference.}
\label{tab:rse_m_scaling}
\small
\setlength{\tabcolsep}{7pt}
\begin{tabular}{ccccc}
\toprule
$M$ & Moment (s) & Opt. (s) & Stats (KB) & Cache (MB) \\
\midrule
2  & $0.00\pm0.00$ & $0.28\pm0.23$ & 0.5   & 0.8  \\
4  & $0.00\pm0.00$ & $0.37\pm0.08$ & 1.6   & 1.6  \\
8  & $0.01\pm0.00$ & $0.38\pm0.05$ & 5.6   & 3.1  \\
16 & $0.04\pm0.00$ & $0.53\pm0.05$ & 21.3  & 6.2  \\
32 & $0.16\pm0.02$ & $0.57\pm0.05$ & 82.5  & 12.3 \\
64 & $0.60\pm0.02$ & $0.66\pm0.07$ & 324.6 & 24.5 \\
\bottomrule
\end{tabular}
\end{table}

Table~\ref{tab:rse_m_scaling} shows that the optimization stage grows only
moderately as the model pool becomes substantially larger. At $M=64$,
the one-time moment construction takes $0.60$\,s and the subsequent
optimization takes $0.66$\,s. The cached RSE statistics require
$324.6$\,KB, compared with $24.5$\,MB for retaining the complete prediction
cache.

\begin{table}[htbp]
\centering
\caption{\textbf{RSE runtime and memory scaling with the number of classes.}
The experiment fixes $N=10{,}000$ and $M=16$ and varies $K$. Runtime is
reported as mean $\pm$ standard deviation over three runs. ``Moment''
denotes the one-time construction of
$\{\mathbf A_j,\mathbf b_j\}_{j=1}^{K}$ and ``Opt.'' denotes the subsequent
RSE optimization. Times exclude synthetic tensor generation and base-model
inference.}
\label{tab:rse_k_scaling}
\small
\setlength{\tabcolsep}{7pt}
\begin{tabular}{ccccc}
\toprule
$K$ & Moment (s) & Opt. (s) & Stats (KB) & Cache (MB) \\
\midrule
5   & $0.02\pm0.00$ & $0.36\pm0.07$ & 10.2  & 3.1  \\
10  & $0.04\pm0.00$ & $0.46\pm0.15$ & 21.3  & 6.2  \\
20  & $0.06\pm0.00$ & $0.67\pm0.06$ & 43.0  & 12.3 \\
50  & $0.14\pm0.00$ & $1.17\pm0.08$ & 106.5 & 30.6 \\
100 & $0.27\pm0.01$ & $1.77\pm0.19$ & 213.0 & 61.1 \\
\bottomrule
\end{tabular}
\end{table}

Table~\ref{tab:rse_k_scaling} shows the corresponding behavior as the
number of classes increases. At $K=100$, moment construction requires
$0.27$\,s and optimization requires $1.77$\,s, while the cached sufficient
statistics occupy only $213.0$\,KB compared with a $61.1$\,MB prediction
cache.

Overall, these experiments support the computational scaling claim in the
main paper. Once the one-time moments have been constructed, RSE operates
only on the cached sufficient statistics, whose storage is $O(M^2K)$ and
does not depend on the number of calibration samples $N$. The increasing
moment-construction cost with $M$ is expected because constructing
$\mathbf A_j\in\mathbb R^{M\times M}$ requires the pairwise model moments.

\begin{table}[htbp]
    \centering
    \caption{\textbf{Significance testing on the Selective Copying task under input shift.} We report mean accuracy (\%) $\pm$ standard deviation obtained via $B=1000$ bootstrap resamples of the shifted test set. The highest accuracy is marked in \textbf{bold}. Statistical significance against our proposed method (RSE) is determined via a one-sided bootstrap hypothesis test: $^{*}$ denotes $p < 0.05$ and $^{**}$ denotes $p < 0.01$. RSE significantly outperforms both individual models and all ensemble baselines.}
    \label{tab:sc_significance}
    \small
    \begin{tabular}{l cc}
        \toprule
        \textbf{Method} & \textbf{Overall Accuracy (\%)} & \textbf{Significance} \\
        \midrule
        \textbf{RSE (Ours)} & \textbf{86.7 $\pm$ 1.5} & -- \\
        \midrule
        \multicolumn{3}{l}{\textit{Optimization Baselines}} \\
        DRO                 & 81.7 $\pm$ 1.7 & $^{**}$ \\
        KL-RS               & 80.1 $\pm$ 1.7 & $^{**}$ \\
        ERM                 & 80.1 $\pm$ 1.7 & $^{**}$ \\
        SRM                 & 79.3 $\pm$ 1.8 & $^{**}$ \\
        IRM                 & 78.1 $\pm$ 1.8 & $^{**}$ \\
        \midrule
        \multicolumn{3}{l}{\textit{Non-Optimization Baselines \& Models}} \\
        Best Individual     & 79.3 $\pm$ 1.8 & $^{**}$ \\
        Acc. Weighted       & 55.1 $\pm$ 2.2 & $^{**}$ \\
        Uniform             & 48.6 $\pm$ 2.3 & $^{**}$ \\
        Majority Vote       & 40.1 $\pm$ 2.2 & $^{**}$ \\
        \bottomrule
    \end{tabular}
\end{table}

\section{Expanding Theoretical Scope Beyond Label Shift}
\label{app:expanding_scope}

The main formulation indexes distribution shift by the class label because
labels provide a natural and universally available finite shift index.
However, the RSE construction is not intrinsically restricted to labels.
The same formulation applies whenever the anticipated distribution shift
can be represented as a change in the proportions of a fixed finite set of
cells.

\paragraph{Finite-cell reindexing.}
Let
\(
S \in \mathcal{S}:=\{1,\dots,E\}
\)
denote a finite cell index. Depending on the application, these cells may
represent demographic groups, domains, environments, time periods, or
joint label--group cells. We assume that the cells form a fixed partition
and that the shift changes their mixture proportions while preserving the
cell-conditional distributions. In particular,
\[
P_{\mathrm{target}}(X,Y\mid S=s)
=
P_{\mathrm{ref}}(X,Y\mid S=s),
\qquad s=1,\dots,E,
\]
while the marginal distribution of $S$ may change.

For each cell $s$, define the cell-conditional risk
\begin{equation}
\label{eq:group_risk}
r_G(\mathbf{w},s)
:=
\mathbb{E}_{X,Y\mid S=s}
\left[
L\big(Y,\widehat{\mathbf p}_{\mathbf w}(X)\big)
\right].
\end{equation}
Collecting these quantities gives
\[
\mathbf r_G(\mathbf w)
=
\big(
r_G(\mathbf w,1),\dots,r_G(\mathbf w,E)
\big)^\top.
\]
Let
\[
\mathbf q=(q_1,\dots,q_E)^\top\in\Delta_E,
\qquad
q_s=P(S=s),
\]
and let $\mathbf q_{\mathrm{ref}}^G\in\Delta_E$ denote the reference
cell distribution. The corresponding mixture risk is
\begin{equation}
\label{eq:group_mixture_risk}
R_G(\mathbf w,\mathbf q)
=
\sum_{s=1}^E q_s r_G(\mathbf w,s)
=
\mathbf q^\top \mathbf r_G(\mathbf w).
\end{equation}

The robust satisficing fragility is therefore defined exactly as in
Definition~\ref{def:fragility_main},
\begin{equation}
\label{eq:group_fragility_def}
\kappa_{\tau,G}(\mathbf w)
:=
\inf
\left\{
\kappa\ge0:
R_G(\mathbf w,\mathbf q)
\le
\tau
+
\kappa
D_{\mathrm{KL}}
\big(
\mathbf q\|
\mathbf q_{\mathrm{ref}}^G
\big),
\quad
\forall \mathbf q\in\Delta_E
\right\}.
\end{equation}
In the well-posed interior regime, this is equivalently
\begin{equation}
\label{eq:group_fragility_ratio}
\kappa_{\tau,G}(\mathbf w)
=
\sup_{
\mathbf q\in\Delta_E,\,
\mathbf q\neq\mathbf q_{\mathrm{ref}}^G
}
\frac{
\mathbf q^\top\mathbf r_G(\mathbf w)-\tau
}{
D_{\mathrm{KL}}
\big(
\mathbf q\|
\mathbf q_{\mathrm{ref}}^G
\big)
}.
\end{equation}

As in Remark~\ref{rem:shift_certificate_main}, this immediately gives the
shift-dependent certificate
\begin{equation}
\label{eq:group_shift_certificate}
\sup_{
\mathbf q\in\Delta_E:
D_{\mathrm{KL}}
(
\mathbf q\|
\mathbf q_{\mathrm{ref}}^G
)
\le\varepsilon
}
R_G(\mathbf w,\mathbf q)
\le
\tau+\kappa_{\tau,G}(\mathbf w)\varepsilon.
\end{equation}

Thus, whenever a distribution shift can be represented as a change in the
mixture proportions of fixed cells with stable cell-conditional
distributions, RSE directly provides the same type of robustness
certificate as in the label-shift formulation.

\paragraph{Quadratic Brier-risk structure.}
The Brier-score construction also carries over without modification.
Define
\[
\mathbf P(x)
:=
\begin{bmatrix}
\mathbf p_1(x) &
\cdots &
\mathbf p_M(x)
\end{bmatrix}
\in\mathbb R^{K\times M}.
\]
Then
\[
\widehat{\mathbf p}_{\mathbf w}(x)
=
\mathbf P(x)\mathbf w.
\]
Using the Brier loss gives
\begin{align}
r_G(\mathbf w,s)
&=
\mathbb E_{X,Y\mid S=s}
\left[
\left\|
\mathbf e_Y-\mathbf P(X)\mathbf w
\right\|_2^2
\right]
\nonumber\\
&=
\mathbf w^\top \mathbf A_s\mathbf w
+
\mathbf b_s^\top\mathbf w
+
1,
\label{eq:group_quad_risk}
\end{align}
where
\begin{equation}
\label{eq:group_moments}
\mathbf A_s
=
\mathbb E
\left[
\mathbf P(X)^\top\mathbf P(X)
\mid
S=s
\right],
\qquad
\mathbf b_s
=
-2
\mathbb E
\left[
\mathbf P(X)^\top\mathbf e_Y
\mid
S=s
\right].
\end{equation}
For any $\mathbf v\in\mathbb R^M$,
\[
\mathbf v^\top\mathbf A_s\mathbf v
=
\mathbb E
\left[
\|\mathbf P(X)\mathbf v\|_2^2
\mid
S=s
\right]
\ge0,
\]
so $\mathbf A_s\succeq0$. Hence the same convex quadratic structure used
by RSE is preserved after reindexing from classes to cells.

\paragraph{KL adversary and scalar fragility root.}
For fixed $\mathbf w$ and $\kappa>0$, the KL-penalized inner problem becomes
\begin{equation}
\label{eq:group_phi}
\Phi_G(\kappa;\mathbf w)
=
\sup_{\mathbf q\in\Delta_E}
\left\{
\mathbf q^\top\mathbf r_G(\mathbf w)
-
\kappa
D_{\mathrm{KL}}
\big(
\mathbf q\|
\mathbf q_{\mathrm{ref}}^G
\big)
\right\}.
\end{equation}
Assuming
\[
(q_{\mathrm{ref}}^G)_s>0,
\qquad
s=1,\dots,E,
\]
the KKT derivation of Lemma~\ref{lem:kl-inner} is unchanged and yields
\begin{equation}
\label{eq:group_qstar}
(q_G^\star)_s
=
\frac{
(q_{\mathrm{ref}}^G)_s
\exp\big(r_G(\mathbf w,s)/\kappa\big)
}{
\sum_{\ell=1}^E
(q_{\mathrm{ref}}^G)_\ell
\exp\big(r_G(\mathbf w,\ell)/\kappa\big)
}.
\end{equation}
The corresponding optimal value is
\begin{equation}
\label{eq:group_phi_closed}
\Phi_G(\kappa;\mathbf w)
=
\kappa
\log
\left(
\sum_{s=1}^E
(q_{\mathrm{ref}}^G)_s
\exp\big(r_G(\mathbf w,s)/\kappa\big)
\right).
\end{equation}

Define the cell-indexed well-posed region
\begin{equation}
\label{eq:group_wellposed}
\mathcal W_{\tau,G}
=
\left\{
\mathbf w:
R_G(\mathbf w,\mathbf q_{\mathrm{ref}}^G)
<
\tau
<
\max_s r_G(\mathbf w,s)
\right\}.
\end{equation}
For $\mathbf w\in\mathcal W_{\tau,G}$, the same existence and uniqueness
argument as in Appendix~\ref{app:solve_kappa} gives a unique positive root
\begin{equation}
\label{eq:group_root}
\tau
=
\kappa_{\tau,G}(\mathbf w)
\log
\left(
\sum_{s=1}^E
(q_{\mathrm{ref}}^G)_s
\exp
\left(
\frac{
r_G(\mathbf w,s)
}{
\kappa_{\tau,G}(\mathbf w)
}
\right)
\right).
\end{equation}

\paragraph{Lagrange--B\"urmann expansion.}
The expansion in Proposition~\ref{prop:lb-kappa-main} also changes only by
reindexing. Set
\[
\alpha_s
:=
r_G(\mathbf w,s),
\qquad
\tau_{0,G}
:=
R_G(\mathbf w,\mathbf q_{\mathrm{ref}}^G),
\]
and define $f_G:\mathbb R\to\mathbb R$ by
$f_G(0):=\tau_{0,G}$ and
\[
f_G(u)
:=
\frac{1}{u}
\log\left(
\sum_{s=1}^E
(q_{\mathrm{ref}}^G)_s e^{\alpha_s u}
\right)
\qquad\text{for }u\neq0.
\]
Its expansion around $u=0$ is
\[
f_G(u)
=
\tau_{0,G}
+
a_{2,G}u
+
a_{3,G}u^2
+
\cdots,
\]
where
\begin{equation}
\label{eq:group_a2}
a_{2,G}
=
\frac12
\sum_{s=1}^E
(q_{\mathrm{ref}}^G)_s
(\alpha_s-\tau_{0,G})^2.
\end{equation}
When the cell risks are non-degenerate so that $a_{2,G}>0$, the same
Lagrange--B\"urmann inverse expansion and the same geometric truncation
analysis in Proposition~\ref{prop:lb-kappa-main} and
Corollary~\ref{cor:lb-trunc-main} apply after replacing the class index
$j$ by the cell index $s$ and $K$ by $E$.

\paragraph{Proximal RSE guarantees.}
The proximal weight update is also unchanged. With
\[
\mathbf A_{\mathrm{adv},t}
=
\sum_{s=1}^E
(q_{G,t}^\star)_s\mathbf A_s,
\qquad
\mathbf b_{\mathrm{adv},t}
=
\sum_{s=1}^E
(q_{G,t}^\star)_s\mathbf b_s,
\]
the adversarially tilted risk remains a convex quadratic because
$\mathbf A_s\succeq0$ for every cell.

The convergence analysis uses the identity
\begin{equation}
\label{eq:group_phi_grad}
\nabla_{\mathbf w}\Phi_G(\kappa;\mathbf w)
=
\sum_{s=1}^E
(q_G^\star)_s
\nabla_{\mathbf w}r_G(\mathbf w,s),
\end{equation}
together with the smooth scalar KL root and the positive-semidefinite
quadratic risk. Therefore, Propositions~\ref{prop:rse_descent_main} and
\ref{prop:approx_root_stationarity_main} carry over under their
corresponding assumptions after replacing
$\mathcal W_\tau$ by $\mathcal W_{\tau,G}$ and the class-indexed moments by
their cell-indexed counterparts. The per-iteration complexity becomes
\[
O(M^2E+M^3),
\]
while the optimization loop remains independent of the number of
calibration samples.

\paragraph{Learned cells and covariate shifts.}
The cell index need not correspond to observed metadata. When robustness
to a covariate shift is desired, a fixed partition can instead be defined
from input features, metadata, or learned representations. For example,
an unsupervised clustering model may be fitted on the calibration data and
its fixed assignment rule used to define the cells. If the resulting
cell-conditional distributions remain stable at deployment, changes in
the input distribution are represented through changes in the cell
mixture coefficients and are therefore covered by the finite-cell
formulation above.

This view is related to latent label shift
\citep{roberts2022unsupervised}, which studies settings where latent-class
proportions vary while the corresponding class-conditional distributions
remain fixed. It is also related to GEORGE
\citep{sohoni2020no}, which constructs proxy subclasses by clustering
learned representations and then uses these proxy groups in a robust
optimization objective. These works support the use of learned finite
partitions when the relevant shift structure is not directly observed.

\subsection{Empirical Robustness Beyond the Explicit Extension}

The natural-shift experiments in the main paper use the base label-indexed
RSE objective rather than requiring explicit group or domain metadata in
the optimization. Their performance under subpopulation and temporal shift
therefore provides empirical evidence of cross-shift robustness beyond the
strict label-shift model.

One possible explanation is that broader distribution shifts can induce
systematic changes in class-wise risks, so reducing sensitivity to the
label-risk profile can also act as a useful regularizer under related
shifts. This interpretation is consistent with prior work showing that
label-shift adaptation over expanded or joint label spaces can mitigate
some shifts induced by spurious correlations
\citep{sun2023beyond}. These empirical results should be distinguished
from the formal finite-cell guarantee above, which applies when the shift
is represented as changing proportions of fixed cells with stable
cell-conditional distributions.

\section{Derivation of the Quadratic Risk Form}
\label{app:psd_derivation}

In this section we show that for each class
$y_j \in \mathcal{Y}$, the matrix $\mathbf{A}_j \in \mathbb{R}^{M\times M}$
defined in Section~\ref{subsec:quad-risk} is positive semidefinite.

\subsection{Explicit definition of the matrix $\mathbf{A}_j$}

Fix a class index $j \in \{1,\dots,K\}$.
Let $X$ be a random input drawn from the class-conditional distribution
$P(X \mid Y = y_j)$.

For each model $m \in \{1,\dots,M\}$, the model output is a probability vector
\[
\mathbf{p}_m(X)
=
\begin{bmatrix}
p_{m,1}(X)\\
\vdots\\
p_{m,K}(X)
\end{bmatrix}
\in \mathbb{R}^K,
\qquad
p_{m,k}(X) \ge 0,\quad \sum_{k=1}^K p_{m,k}(X) = 1.
\]

The matrix $\mathbf{A}_j$ is defined entrywise as
\[
(\mathbf{A}_j)_{ab}
=
\mathbb{E}_{X\mid Y=y_j}
\big[
\mathbf{p}_a(X)^\top \mathbf{p}_b(X)
\big],
\qquad a,b \in \{1,\dots,M\}.
\]

We assume $\mathbb{E}_{X\mid Y=y_j}\!\left[\|\mathbf{p}_m(X)\|_2^2\right] < \infty$ for all $m$,
which holds automatically here since $\mathbf{p}_m(X)$ has nonnegative entries and
$\sum_{k=1}^K p_{m,k}(X)=1$, hence $\|\mathbf{p}_m(X)\|_2^2=\sum_{k=1}^K p_{m,k}(X)^2\le \sum_{k=1}^K p_{m,k}(X)=1$.

Expanding the dot product coordinate-wise yields
\[
\mathbf{p}_a(X)^\top \mathbf{p}_b(X)
=
\sum_{k=1}^K p_{a,k}(X)\,p_{b,k}(X),
\]
and therefore
\[
(\mathbf{A}_j)_{ab}
=
\sum_{k=1}^K
\mathbb{E}_{X\mid Y=y_j}
\big[
p_{a,k}(X)\,p_{b,k}(X)
\big].
\]

Thus $\mathbf{A}_j$ is the $M\times M$ matrix
\[
\mathbf{A}_j
=
\begin{bmatrix}
\mathbb{E}_{X\mid Y=y_j}[\mathbf{p}_1(X)^\top\mathbf{p}_1(X)] &
\cdots &
\mathbb{E}_{X\mid Y=y_j}[\mathbf{p}_1(X)^\top\mathbf{p}_M(X)] \\
\vdots & \ddots & \vdots \\
\mathbb{E}_{X\mid Y=y_j}[\mathbf{p}_M(X)^\top\mathbf{p}_1(X)] &
\cdots &
\mathbb{E}_{X\mid Y=y_j}[\mathbf{p}_M(X)^\top\mathbf{p}_M(X)]
\end{bmatrix}.
\]

Moreover, $\mathbf{A}_j$ is symmetric because for any $a,b$,
\[
(\mathbf{A}_j)_{ab}
=\mathbb{E}_{X\mid Y=y_j}[\mathbf{p}_a(X)^\top\mathbf{p}_b(X)]
=\mathbb{E}_{X\mid Y=y_j}[\mathbf{p}_b(X)^\top\mathbf{p}_a(X)]
=(\mathbf{A}_j)_{ba},
\]
where we used $\mathbf{p}_a(X)^\top\mathbf{p}_b(X)=\mathbf{p}_b(X)^\top\mathbf{p}_a(X)$ for real vectors.

\subsection{Positive semidefiniteness via the quadratic form}

Recall that a symmetric matrix $\mathbf{H}\in\mathbb{R}^{M\times M}$ is
PSD if
$\mathbf{v}^\top \mathbf{H}\mathbf{v}\ge 0$ for all $\mathbf{v}\in\mathbb{R}^M$.

Let $\mathbf{v} = (v_1,\dots,v_M)^\top \in \mathbb{R}^M$ be an arbitrary vector in $\mathbb{R}^M$.
\begin{align*}
\mathbf{v}^\top \mathbf{A}_j \mathbf{v}
&=
\sum_{a=1}^M \sum_{b=1}^M
v_a (\mathbf{A}_j)_{ab} v_b \\
&=
\sum_{a=1}^M \sum_{b=1}^M
v_a v_b
\,
\mathbb{E}_{X\mid Y=y_j}
\big[
\mathbf{p}_a(X)^\top \mathbf{p}_b(X)
\big].
\end{align*}

Since the sums over $a$ and $b$ are finite, we may interchange the finite summations and the expectation. Therefore,
\begin{align*}
\mathbf{v}^\top \mathbf{A}_j \mathbf{v}
&=
\mathbb{E}_{X\mid Y=y_j}
\Big[
\sum_{a=1}^M \sum_{b=1}^M
v_a v_b
\,
\mathbf{p}_a(X)^\top \mathbf{p}_b(X)
\Big].
\end{align*}

Define, for each realization $X$, the vector
\[
\mathbf{u}(X; \mathbf{v})
:=
\sum_{m=1}^M v_m \mathbf{p}_m(X)
\in \mathbb{R}^K.
\]

Expanding $\mathbf{u}(X;\mathbf{v})^\top \mathbf{u}(X;\mathbf{v})$ gives
\[
\mathbf{u}(X;\mathbf{v})^\top \mathbf{u}(X;\mathbf{v})
=
\sum_{a=1}^M \sum_{b=1}^M
v_a v_b
\,
\mathbf{p}_a(X)^\top \mathbf{p}_b(X).
\]

Substituting this identity yields
\[
\mathbf{v}^\top \mathbf{A}_j \mathbf{v}
=
\mathbb{E}_{X\mid Y=y_j}
\big[
\mathbf{u}(X;\mathbf{v})^\top \mathbf{u}(X;\mathbf{v})
\big]
=
\mathbb{E}_{X\mid Y=y_j}
\big[
\|\mathbf{u}(X;\mathbf{v})\|_2^2
\big].
\]

Writing the squared norm coordinate-wise,
\[
\|\mathbf{u}(X;\mathbf{v})\|_2^2
=
\sum_{k=1}^K
\Big(
\sum_{m=1}^M v_m p_{m,k}(X)
\Big)^2
\ge 0
\quad
\text{for every realization of } X
\text{ and every } \mathbf{v}\in\mathbb{R}^M.
\]

Since the integrand is nonnegative for every $X$, its expectation is also
nonnegative. Therefore
\[
\mathbf{v}^\top \mathbf{A}_j \mathbf{v} \ge 0
\quad \text{for all } \mathbf{v} \in \mathbb{R}^M,
\]
which proves that $\mathbf{A}_j$ is positive semidefinite.

\subsection{Finite-sample estimator and preservation of positive semidefiniteness}

Fix a class index $j \in \{1,\dots,K\}$.
Let
\[
\mathcal{D}_j := \{x^{(i)} : y^{(i)} = y_j\}
\]
denote the subset of the training data consisting of all samples whose label
equals $y_j$, and let $|\mathcal{D}_j|$ denote its cardinality.

For each input $x \in \mathcal{D}_j$ and each model
$m \in \{1,\dots,M\}$, recall that the model produces a probability vector
\[
\mathbf{p}_m(x)
=
\begin{bmatrix}
p_{m,1}(x)\\
\vdots\\
p_{m,K}(x)
\end{bmatrix}
\in \mathbb{R}^K,
\qquad
p_{m,k}(x) \ge 0,
\quad
\sum_{k=1}^K p_{m,k}(x) = 1.
\]

Using these vectors, we define the matrix-valued function
\[
\mathbf{P}(x)
:=
\begin{bmatrix}
\mathbf{p}_1(x) & \mathbf{p}_2(x) & \cdots & \mathbf{p}_M(x)
\end{bmatrix}
\in \mathbb{R}^{K \times M}.
\]
That is, the $m$-th column of $\mathbf{P}(x)$ is the probability vector
$\mathbf{p}_m(x)$ produced by model $m$ at input $x$.

The empirical (finite-sample) estimator of $\mathbf{A}_j$ is then defined as
\[
\widehat{\mathbf{A}}_j
=
\frac{1}{|\mathcal{D}_j|}
\sum_{x^{(i)} \in \mathcal{D}_j}
\mathbf{P}(x^{(i)})^\top \mathbf{P}(x^{(i)}).
\]

First, fix an arbitrary input $x \in \mathcal{D}_j$.
Consider the matrix $\mathbf{P}(x)^\top \mathbf{P}(x) \in \mathbb{R}^{M\times M}$.
To verify that this matrix is positive semidefinite, we must show that
\[
\mathbf{v}^\top \mathbf{P}(x)^\top \mathbf{P}(x)\mathbf{v} \ge 0
\quad
\text{for all vectors } \mathbf{v} \in \mathbb{R}^M.
\]

Let $\mathbf{v} = (v_1,\dots,v_M)^\top \in \mathbb{R}^M$ be arbitrary.
We compute the quadratic form explicitly:
\begin{align*}
\mathbf{v}^\top \mathbf{P}(x)^\top \mathbf{P}(x)\mathbf{v}
&=
(\mathbf{P}(x)\mathbf{v})^\top (\mathbf{P}(x)\mathbf{v}).
\end{align*}

Next, we write $\mathbf{P}(x)\mathbf{v}$ explicitly.
\[
\mathbf{P}(x)\mathbf{v}
=
\sum_{m=1}^M v_m \mathbf{p}_m(x)
=
\begin{bmatrix}
\sum_{m=1}^M v_m p_{m,1}(x)\\
\vdots\\
\sum_{m=1}^M v_m p_{m,K}(x)
\end{bmatrix}
\in \mathbb{R}^K.
\]

Substituting this expression back into the quadratic form yields
\begin{align*}
(\mathbf{P}(x)\mathbf{v})^\top (\mathbf{P}(x)\mathbf{v})
&=
\sum_{k=1}^K
\Big(
\sum_{m=1}^M v_m p_{m,k}(x)
\Big)^2.
\end{align*}

Each term in the sum is a square of a real number and is therefore
nonnegative. Hence
\[
(\mathbf{P}(x)\mathbf{v})^\top (\mathbf{P}(x)\mathbf{v}) \ge 0
\quad
\text{for all } \mathbf{v} \in \mathbb{R}^M,
\]
which implies
\[
\mathbf{v}^\top \mathbf{P}(x)^\top \mathbf{P}(x)\mathbf{v} \ge 0
\quad
\text{for all } \mathbf{v} \in \mathbb{R}^M.
\]
By definition, this shows that the matrix
$\mathbf{P}(x)^\top \mathbf{P}(x)$ is positive semidefinite.

We now consider the empirical estimator $\widehat{\mathbf{A}}_j$.
For any $\mathbf{v} \in \mathbb{R}^M$, we have
\begin{align*}
\mathbf{v}^\top \widehat{\mathbf{A}}_j \mathbf{v}
&=
\mathbf{v}^\top
\left(
\frac{1}{|\mathcal{D}_j|}
\sum_{x^{(i)} \in \mathcal{D}_j}
\mathbf{P}(x^{(i)})^\top \mathbf{P}(x^{(i)})
\right)
\mathbf{v} \\
&=
\frac{1}{|\mathcal{D}_j|}
\sum_{x^{(i)} \in \mathcal{D}_j}
\mathbf{v}^\top
\mathbf{P}(x^{(i)})^\top \mathbf{P}(x^{(i)})
\mathbf{v}.
\end{align*}

As shown above, each term
$\mathbf{v}^\top \mathbf{P}(x^{(i)})^\top \mathbf{P}(x^{(i)}) \mathbf{v}$
is nonnegative. Since $|\mathcal{D}_j|^{-1} > 0$, the weighted sum of these
nonnegative terms is also nonnegative. Therefore,
\[
\mathbf{v}^\top \widehat{\mathbf{A}}_j \mathbf{v} \ge 0
\quad
\text{for all } \mathbf{v} \in \mathbb{R}^M.
\]

With this, we conclude that the matrix $\widehat{\mathbf{A}}_j$
is positive semidefinite.

\section{Derivation of \texorpdfstring{$\Phi(\kappa;\mathbf{w})$}{Phi(kappa;w)}}
\label{app:phi_derivation}

In this section, we derive the inner objective
\(\Phi(\kappa;\mathbf{w})\) starting from the robust satisficing definition in
Definition~\ref{def:fragility_main}.

We fix a reference label distribution
\(\mathbf{q}_{\mathrm{ref}} \in \Delta_K\)
and an aspiration level \(\tau > 0\).
Throughout this appendix, we assume that
\[
(q_{\mathrm{ref}})_j > 0
\quad \text{for all } j \in \{1,\dots,K\},
\]
which ensures that the Kullback--Leibler divergence
\(D_{\mathrm{KL}}(\mathbf{q}\,\|\,\mathbf{q}_{\mathrm{ref}})\)
is finite for all \(\mathbf{q}\in\Delta_K\).

Recall Definition~\ref{def:fragility_main}.
For a fixed ensemble weight vector \(\mathbf{w}\in\mathcal{W}_\tau\),
the fragility \(\kappa_\tau(\mathbf{w})\) is defined as the smallest
nonnegative scalar \(\kappa\) such that 
\begin{equation}
\label{eq:rs-ineq-app}
R(\mathbf{w},\mathbf{q})
\le
\tau + \kappa\,\Delta(\mathbf{q},\mathbf{q}_{\mathrm{ref}})
\qquad
\text{for all } \mathbf{q}\in\Delta_K.
\end{equation}

To analyze feasibility for a \emph{candidate} value \(\kappa \ge 0\),
we rewrite~\eqref{eq:rs-ineq-app} equivalently as
\begin{equation}
\label{eq:rs-ineq-rearranged-app}
R(\mathbf{w},\mathbf{q})
-
\kappa\,\Delta(\mathbf{q},\mathbf{q}_{\mathrm{ref}})
\le
\tau
\qquad
\text{for all } \mathbf{q}\in\Delta_K.
\end{equation}
The inequality~\eqref{eq:rs-ineq-rearranged-app} holds if and only if it holds
for the worst-case distribution. Equivalently, for a given \(\kappa\ge 0\),
\begin{equation}
\label{eq:sup-condition-app}
\sup_{\mathbf{q}\in\Delta_K}
\Big\{
R(\mathbf{w},\mathbf{q})
-
\kappa\,\Delta(\mathbf{q},\mathbf{q}_{\mathrm{ref}})
\Big\}
\le
\tau.
\end{equation}

This motivates defining a scalar-valued objective \(\Phi(\kappa;\mathbf{w})\) that
summarizes the worst-case left-hand side. For any \(\kappa \ge 0\), define
\begin{equation}
\label{eq:Phi-def-app}
\Phi(\kappa;\mathbf{w})
:=
\sup_{\mathbf{q}\in\Delta_K}
\Big\{
R(\mathbf{w},\mathbf{q})
-
\kappa\,\Delta(\mathbf{q},\mathbf{q}_{\mathrm{ref}})
\Big\}.
\end{equation}
With this definition, a candidate \(\kappa\) satisfies the robust satisficing
condition if and only if
\begin{equation}
\label{eq:Phi-feasible-app}
\Phi(\kappa;\mathbf{w}) \le \tau.
\end{equation}

The set of feasible $\kappa$ values,
$\{\kappa\ge0:\Phi(\kappa;\mathbf w)\le\tau\}$,
is closed, being the intersection over all
$\mathbf q\in\Delta_K$ of the closed sets defined by
the continuous inequalities in
\eqref{eq:rs-ineq-rearranged-app}.
Since $\Phi$ is nonincreasing in $\kappa$, every value above
a feasible value is also feasible. Thus, this set is
$[\kappa_\tau(\mathbf w),\infty)$.

Moreover, the fragility \(\kappa_\tau(\mathbf{w})\) is precisely the smallest
\(\kappa\ge 0\) for which~\eqref{eq:Phi-feasible-app} holds.

We now make explicit the dependence of the risk \(R(\mathbf{w},\mathbf{q})\)
on the label distribution \(\mathbf{q}\).
For each class $y_j$, recall the class-conditional risk (cf.\ \eqref{eq:class_conditional_risk_main})
\[
r(\mathbf{w},y_j)
:=
\mathbb{E}_{X\sim P(X\mid Y=y_j)}
\Big[
L\big(y_j,\widehat{\mathbf p}_{\mathbf w}(X)\big)
\Big].
\]

For a label distribution $\mathbf q\in\Delta_K$, the corresponding mixture risk is
\begin{equation}
\label{eq:risk-linear-app}
R(\mathbf{w},\mathbf{q})
=
\sum_{j=1}^K q_j\,r(\mathbf{w},y_j)
=
\mathbf{q}^\top \mathbf{r}(\mathbf{w}),
\end{equation}
where \(\mathbf{r}(\mathbf{w})\in\mathbb{R}^K\) denotes the vector of
class-conditional risks defined in~\eqref{eq:r(w)}.

We now instantiate the \(\Delta\) with the Kullback--Leibler
divergence:
\begin{equation}
\label{eq:kl-def-app}
\Delta(\mathbf{q},\mathbf{q}_{\mathrm{ref}})
=
D_{\mathrm{KL}}(\mathbf{q}\,\|\,\mathbf{q}_{\mathrm{ref}})
=
\sum_{j=1}^K
q_j \log\frac{q_j}{(q_{\mathrm{ref}})_j}.
\end{equation}
Substituting~\eqref{eq:risk-linear-app} and~\eqref{eq:kl-def-app}
into~\eqref{eq:Phi-def-app} yields the KL-based inner problem
\begin{equation}
\label{eq:Phi-kl-app}
\Phi(\kappa;\mathbf{w})
=
\sup_{\mathbf{q}\in\Delta_K}
\Big\{
\mathbf{q}^\top \mathbf{r}(\mathbf{w})
-
\kappa\,D_{\mathrm{KL}}(\mathbf{q}\,\|\,\mathbf{q}_{\mathrm{ref}})
\Big\}.
\end{equation}

Fix $\mathbf{w}\in\mathcal W_\tau$. For any $\kappa_1 \ge \kappa_2 \ge 0$ and any
$\mathbf{q}\in\Delta_K$, since
$D_{\mathrm{KL}}(\mathbf{q}\,\|\,\mathbf{q}_{\mathrm{ref}})\ge 0$, we have
\[
\mathbf{q}^\top \mathbf{r}(\mathbf{w})
-
\kappa_1 D_{\mathrm{KL}}(\mathbf{q}\,\|\,\mathbf{q}_{\mathrm{ref}})
\le
\mathbf{q}^\top \mathbf{r}(\mathbf{w})
-
\kappa_2 D_{\mathrm{KL}}(\mathbf{q}\,\|\,\mathbf{q}_{\mathrm{ref}})
\]
Taking the supremum over $\mathbf{q}\in\Delta_K$ on both sides and using the
definition in~\eqref{eq:Phi-kl-app} yields
\[
\Phi(\kappa_1;\mathbf{w}) \le \Phi(\kappa_2;\mathbf{w}),
\]
which shows that the function $\kappa\mapsto \Phi(\kappa;\mathbf{w})$ is nonincreasing on
$[0,\infty)$.

We next derive explicit upper and lower bounds on $\Phi(\kappa;\mathbf{w})$.
At $\kappa=0$, the KL term vanishes, and
\[
\Phi(0;\mathbf{w})
=
\sup_{\mathbf{q}\in\Delta_K} \mathbf{q}^\top \mathbf{r}(\mathbf{w})
=
\max_{j\in\{1,\dots,K\}} r(\mathbf{w},y_j),
\]
where the last equality follows because a linear function over the simplex
$\Delta_K$ attains its maximum at an extreme point.

For a lower bound, note that $\mathbf{q}_{\mathrm{ref}}\in\Delta_K$ is feasible
in~\eqref{eq:Phi-kl-app}. Substituting $\mathbf{q}=\mathbf{q}_{\mathrm{ref}}$
and using
$D_{\mathrm{KL}}(\mathbf{q}_{\mathrm{ref}}\,\|\,\mathbf{q}_{\mathrm{ref}})=0$
gives, for all $\kappa\ge 0$,
\[
\Phi(\kappa;\mathbf{w})
\ge
\mathbf{q}_{\mathrm{ref}}^\top \mathbf{r}(\mathbf{w})
=
R(\mathbf{w},\mathbf{q}_{\mathrm{ref}}).
\]

Combining the monotonicity of $\Phi(\kappa;\mathbf{w})$ with the value at $\kappa=0$ also
yields the uniform upper bound
\[
\Phi(\kappa;\mathbf{w})
\le
\Phi(0;\mathbf{w})
=
\max_{j\in\{1,\dots,K\}} r(\mathbf{w},y_j),
\qquad \forall \kappa\ge 0.
\]

Since we have fixed
$\mathbf w\in\mathcal W_\tau$, the definition of
$\mathcal W_\tau$ gives
\[
R(\mathbf{w},\mathbf{q}_{\mathrm{ref}})
<
\tau
<
\max_{j} r(\mathbf{w},y_j).
\]
If $\tau \ge \max_j r(\mathbf{w},y_j)$, then $\Phi(0;\mathbf{w})\le \tau$ and the
smallest feasible value is $\kappa=0$. Otherwise, if
$\tau\le R(\mathbf w,\mathbf q_{\mathrm{ref}})$,
no finite $\kappa\ge0$ is feasible. Otherwise, the robust
satisficing constraint becomes active at some $\kappa>0$, which we denote by
$\kappa_\tau(\mathbf{w})$, satisfying
\begin{equation}
\label{eq:Phi-root-app}
\Phi\!\big(\kappa_\tau(\mathbf{w});\mathbf{w}\big)
=
\tau.
\end{equation}

Moreover, the equation $\Phi(\kappa;\mathbf w)=\tau$ admits a unique finite root
$\kappa\in(0,\infty)$ precisely in the interior regime
$R(\mathbf w,\mathbf q_{\mathrm{ref}})<\tau<\max_j r(\mathbf w,y_j)$ (Appendix~\ref{app:solve_kappa}). The explicit solution of the maximization problem in~\eqref{eq:Phi-kl-app}, as
well as the closed-form expression for \(\Phi(\kappa;\mathbf{w})\), are derived in
Appendix~\ref{app:kl_derivation}.

\section{KL Inner Problem and Exponential Tilt}
\label{app:kl_derivation}

Throughout this section, we assume
$\kappa>0$.
Strict positivity follows directly from the robust
satisficing constraint. If $\kappa=0$ were feasible, then
\[
R(\mathbf w,\mathbf q)\le\tau
\qquad\text{for every }\mathbf q\in\Delta_K.
\]
Taking the supremum over $\mathbf q$ would imply
\[
\max_j r(\mathbf w,y_j)
=\sup_{\mathbf q\in\Delta_K}R(\mathbf w,\mathbf q)
\le\tau,
\]
contradicting $\mathbf w\in\mathcal W_\tau$. Therefore $\kappa_\tau(\mathbf w)>0$.
For fixed $\mathbf{w}$ we abbreviate $r_j = r(\mathbf{w},y_j)$ and consider
the penalized inner problem, which is a standard KL-regularized variational
optimization; the corresponding exponential-tilting optimizer appears, for example, as Proposition~A.5 in \citet{yan2024kl} for the uniform reference case.
We include the derivation here to match our notation and allow a general
$\mathbf{q}_{\mathrm{ref}}$.

\[
\Phi(\kappa;\mathbf{w})
=
\sup_{\mathbf{q}\in\Delta_K}
\bigg\{
    \sum_{j=1}^K q_j r_j
    -
    \kappa
    \sum_{j=1}^K
    q_j \log \frac{q_j}{(q_{\mathrm{ref}})_j}
\bigg\}.
\]

Fix $\mathbf w$ and $\kappa>0$, and write
$p_j:=(q_{\mathrm{ref}})_j>0$ and
$r_j:=r(\mathbf w,y_j)$.
The quantities $\mathbf w$, $\kappa$, $p_j$, and $r_j$
remain fixed; only $\mathbf q$ is varied.
The objective is
\[
F(\mathbf q)
=
\sum_{j=1}^K q_jr_j
-\kappa\sum_{j=1}^K q_j\log\frac{q_j}{p_j},
\qquad \mathbf q\in\Delta_K,
\]
where $0\log(0/p_j):=0$.
This function is continuous on the compact simplex,
so a maximizer exists. Rewriting it as
\[
F(\mathbf q)
=
\sum_{j=1}^K q_j(r_j+\kappa\log p_j)
-\kappa\sum_{j=1}^K q_j\log q_j
\]
shows that it is strictly concave: the first sum
is linear and the second is a strictly concave
entropy term because $\kappa>0$.
Hence the maximizer is unique.

We show that no class can have zero probability
at this maximizer.
Consider any $\mathbf q\in\Delta_K$ with $q_i=0$.
Since $\sum_j q_j=1$, some other coordinate satisfies
$q_\ell>0$.
For $0<\varepsilon<q_\ell/2$, define
\[
q_i^{(\varepsilon)}=\varepsilon,
\qquad
q_\ell^{(\varepsilon)}=q_\ell-\varepsilon,
\qquad
q_j^{(\varepsilon)}=q_j
\quad\text{for }j\notin\{i,\ell\}.
\]
This transfers probability $\varepsilon$ from class
$\ell$ to class $i$, preserving nonnegativity and
the sum of probabilities. Thus
$\mathbf q^{(\varepsilon)}\in\Delta_K$.

We now subtract the two objective values term by term.
For every $j\notin\{i,\ell\}$, the terms in
$F(\mathbf q^{(\varepsilon)})$ and $F(\mathbf q)$
are identical and cancel.
For the two changed coordinates, substituting the
old and new probabilities gives
\[
\begin{aligned}
F(\mathbf q^{(\varepsilon)})-F(\mathbf q)
={}&
\bigl[\varepsilon r_i-0\bigr]
+\bigl[(q_\ell-\varepsilon)r_\ell-q_\ell r_\ell\bigr]\\
&-\kappa\left[
\varepsilon\log\frac{\varepsilon}{p_i}-0
\right]\\
&-\kappa\left[
(q_\ell-\varepsilon)
\log\frac{q_\ell-\varepsilon}{p_\ell}
-q_\ell\log\frac{q_\ell}{p_\ell}
\right].
\end{aligned}
\]
Combining the first two brackets, we obtain three groups:
\[
\begin{aligned}
F(\mathbf q^{(\varepsilon)})-F(\mathbf q)
={}&
\underbrace{\varepsilon(r_i-r_\ell)}_{T_1}\\
&+\underbrace{
\left(-\kappa\varepsilon
\log\frac{\varepsilon}{p_i}\right)}_{T_2}\\
&+\underbrace{
\left(
-\kappa\left[
(q_\ell-\varepsilon)
\log\frac{q_\ell-\varepsilon}{p_\ell}
-q_\ell\log\frac{q_\ell}{p_\ell}
\right]\right)}_{T_3}.
\end{aligned}
\]
We examine these groups separately.

The first group is already explicit:
\[
T_1=\varepsilon(r_i-r_\ell).
\]
Its coefficient $r_i-r_\ell$ is fixed and finite,
although it may have either sign.

For the second group, the identity
$\log(\varepsilon/p_i)=\log\varepsilon-\log p_i$
gives
\[
T_2
=
\kappa\varepsilon\log(1/\varepsilon)
+\kappa\varepsilon\log p_i.
\]

For the third group, define
\[
h(x):=x\log(x/p_\ell),\qquad x>0.
\]
Here $x$ represents the probability assigned to class
$\ell$, and $p_\ell$ is fixed.
Then
\[
T_3=-\kappa\bigl[h(q_\ell-\varepsilon)-h(q_\ell)\bigr].
\]
The derivatives of $h$ are
\[
h'(x)=\log(x/p_\ell)+1,
\qquad
h''(x)=\frac{1}{x}.
\]
Taylor's theorem expresses a function's change as
\[
h(a+d)-h(a)
=
d\,h'(a)+\frac{d^2}{2}h''(\xi),
\]
where $\xi$ lies between $a$ and $a+d$.
The last term is an exact remainder, so this is
an equality rather than an approximation.

Applying this theorem with $a=q_\ell$ and
$d=-\varepsilon$ gives, for some
$\xi_\varepsilon\in(q_\ell-\varepsilon,q_\ell)$,
\[
h(q_\ell-\varepsilon)-h(q_\ell)
=
-\varepsilon\left(\log\frac{q_\ell}{p_\ell}+1\right)
+\frac{\varepsilon^2}{2\xi_\varepsilon}.
\]
Multiplying by $-\kappa$, we therefore obtain
\[
T_3
=
\kappa\varepsilon\left(\log\frac{q_\ell}{p_\ell}+1\right)
-\frac{\kappa\varepsilon^2}{2\xi_\varepsilon}.
\]
Since $\varepsilon<q_\ell/2$, the intermediate point
satisfies
\[
\xi_\varepsilon>q_\ell-\varepsilon>q_\ell/2.
\]
Consequently,
\[
0<
\frac{\kappa\varepsilon^2}{2\xi_\varepsilon}
< \frac{\kappa\varepsilon^2}{q_\ell},
\]
which bounds the magnitude of the negative remainder
in $T_3$.

Adding the three groups back together yields
\[
\begin{aligned}
F(\mathbf q^{(\varepsilon)})-F(\mathbf q)
={}&
\underbrace{\varepsilon(r_i-r_\ell)}_{T_1}\\
&+\underbrace{
\kappa\varepsilon\log(1/\varepsilon)
+\kappa\varepsilon\log p_i}_{T_2}\\
&+\underbrace{
\kappa\varepsilon\left(\log\frac{q_\ell}{p_\ell}+1\right)
-\frac{\kappa\varepsilon^2}{2\xi_\varepsilon}}_{T_3}.
\end{aligned}
\]
Dividing by $\varepsilon>0$ and using the remainder
bound gives
\[
\begin{aligned}
\frac{F(\mathbf q^{(\varepsilon)})-F(\mathbf q)}
{\varepsilon}
\ge{}&
\kappa\log(1/\varepsilon)
+(r_i-r_\ell)
+\kappa\log p_i\\
&+\kappa\left(\log\frac{q_\ell}{p_\ell}+1\right)
-\frac{\kappa\varepsilon}{q_\ell}.
\end{aligned}
\]
As $\varepsilon\downarrow0$, the term
$\kappa\log(1/\varepsilon)$ tends to $+\infty$.
The terms involving $r_i,r_\ell,p_i,p_\ell,q_\ell$
are fixed and finite, and
$\kappa\varepsilon/q_\ell$ tends to zero.
Thus the right-hand side is positive for all
sufficiently small $\varepsilon>0$, proving
\[
F(\mathbf q^{(\varepsilon)})>F(\mathbf q).
\]

Every point with a zero coordinate can therefore
be improved and cannot be a maximizer.
Consequently, the unique maximizer satisfies
$q_j^\star>0$ for every $j$, so it lies in the
relative interior of the simplex.

Introducing a Lagrange multiplier $\lambda$ for $\sum_j q_j=1$, the
Lagrangian is
\[
\mathcal{L}(\mathbf{q},\lambda)
=
\sum_{j=1}^K q_j r_j
-
\kappa
\sum_{j=1}^K
q_j \log \frac{q_j}{(q_{\mathrm{ref}})_j}
+
\lambda
\Big(
    \sum_{j=1}^K q_j - 1
\Big).
\]
Using $\frac{\partial}{\partial q_j}(q_j\log q_j)=\log q_j+1$, the
stationarity condition is
\[
r_j
-
\kappa
\Big(
    \log \tfrac{q_j}{(q_{\mathrm{ref}})_j} + 1
\Big)
+
\lambda
=
0.
\]
Rearranging and exponentiating yields $q_j \propto
(q_{\mathrm{ref}})_j\exp(r_j/\kappa)$. Enforcing $\sum_j q_j=1$ determines
the unique normalizing constant, giving the exponentially tilted
distribution
\[
(q^\star)_j
=
\frac{(q_{\mathrm{ref}})_j \exp(r_j/\kappa)}
{\sum_{k=1}^K (q_{\mathrm{ref}})_k \exp(r_k/\kappa)},
\]
which matches \eqref{eq:qstar-softmax-method}. Substituting
$\mathbf{q}^\star$ into the objective gives
\[
\Phi(\kappa;\mathbf{w})
=
\kappa \log
\Big(
    \sum_{j=1}^K
    (q_{\mathrm{ref}})_j
    \exp(r_j/\kappa)
\Big),
\]
which is \eqref{eq:Phi-def}.

\section{Solving for \texorpdfstring{$\kappa_\tau(\mathbf{w})$}{kappa}}
\label{app:solve_kappa}

Recall from Lemma~\ref{lem:kl-fragility-scalar} that for fixed $\mathbf w$ the KL fragility
is characterized by the root equation
\eqref{eq:tau-kappa-main}. There are three cases:
(i) if $\tau\ge \max_j r(\mathbf w,y_j)$ then $\kappa_\tau(\mathbf w)=0$;
(ii) otherwise, if $R(\mathbf w,\mathbf q_{\mathrm{ref}})\ge\tau$ then $\kappa_\tau(\mathbf w)=+\infty$;
(iii) if
\[
R(\mathbf w,\mathbf q_{\mathrm{ref}})<\tau<\max_j r(\mathbf w,y_j),
\]
then there is a unique $\kappa_\tau(\mathbf w)\in(0,\infty)$ satisfying \eqref{eq:tau-kappa-main}.
We focus on case (iii) below.

We now (i) establish existence and uniqueness, and
(ii) derive an analytic series representation of $\kappa_\tau(\mathbf{w})$
via Lagrange--B\"urmann inversion together with a non-asymptotic truncation
error bound for the resulting finite-sum approximation.

\subsection{Existence and uniqueness}

For brevity, fix $\mathbf{w}$.
Let $\delta_a$ denote the unit point mass at $a$,
and let $X$ be the random variable representing the class
risk when the class is sampled according to
$\mathbf{q}_{\mathrm{ref}}$.
Set
\[
\alpha_j := r(\mathbf{w},y_j),
\qquad
X \sim \sum_{j=1}^K (q_{\mathrm{ref}})_j \,\delta_{\alpha_j}.
\]
Introduce
\begin{equation}
\label{eq:S-def}
S := \frac{1}{\kappa} > 0,
\qquad
G(S) := \sum_{j=1}^K (q_{\mathrm{ref}})_j e^{\alpha_j S}.
\end{equation}
Then~\eqref{eq:tau-kappa-main} becomes
\begin{equation}
\label{eq:tau-fS}
\tau
=
f(S)
:=
\frac{1}{S}\,\log G(S).
\end{equation}
Since $G(0)=\sum_j (q_{\mathrm{ref}})_j=1$ and $G$ is analytic near $S=0$,
$\log G(S)$ admits a Taylor expansion of the form
\[
\log G(S)
=
\sum_{m=1}^\infty a_m S^m,
\qquad
a_m = \frac{1}{m!}\,\mathrm{cum}_m(X),
\]
where $\mathrm{cum}_m(X)$ are the cumulants of $X$.
Dividing by $S$ gives
\begin{equation}
\label{eq:f-series}
f(S)
=
a_1 + a_2 S + a_3 S^2 + a_4 S^3 + \cdots,
\end{equation}
with
\[
a_1 = \mathbb{E}[X]
=
R(\mathbf{w},\mathbf{q}_{\mathrm{ref}}),
\qquad
a_2 = \tfrac{1}{2}\mathrm{Var}(X).
\]

We assume $\mathrm{Var}(X)>0$ (i.e., the class-wise risks are non-degenerate under $\mathbf{q}_{\mathrm{ref}}$);
otherwise $R(\mathbf{w},\mathbf{q}_{\mathrm{ref}})=\max_j r(\mathbf{w},y_j)$ and the interval
$\big(R(\mathbf{w},\mathbf{q}_{\mathrm{ref}}),\max_j r(\mathbf{w},y_j)\big)$ collapses, making $\kappa_\tau(\mathbf{w})$
trivial.

Standard properties of log-moment generating functions imply:

\begin{itemize}
\item As $S\to 0^+$ (equivalently, $\kappa\to\infty$),
\[
f(S) \longrightarrow f(0) = R(\mathbf{w},\mathbf{q}_{\mathrm{ref}}).
\]
\item As $S\to\infty$ (equivalently, $\kappa\to 0^+$), the largest
class-wise risk dominates:
\[
f(S)
=
\frac{1}{S}\log G(S)
\longrightarrow \max_j \alpha_j
=
\max_j r(\mathbf{w},y_j).
\]
\end{itemize}

Moreover, $f$ is strictly increasing on $(0,\infty)$.
Let $g(S):=\log G(S)$. Since $g$ is convex and $g(0)=0$, the map $S\mapsto g(S)/S$
is nondecreasing on $(0,\infty)$, hence
\[
f'(S)
=
\frac{S g'(S) - g(S)}{S^2}
\ge 0
\quad\text{for all }S>0,
\]
and under $\mathrm{Var}(X)>0$ (non-degenerate risks), $g$ is strictly convex, so $f'(S)>0$ for all $S>0$. Therefore $f$ maps $(0,\infty)$ continuously and strictly
monotonically onto the interval
$\big(R(\mathbf{w},\mathbf{q}_{\mathrm{ref}}),\,\max_j r(\mathbf{w},y_j)\big)$.
For any
\[
\tau \in
\big(R(\mathbf{w},\mathbf{q}_{\mathrm{ref}}),\,\max_j r(\mathbf{w},y_j)\big)
\]
there is a unique $S>0$ solving~\eqref{eq:tau-fS},
and thus a unique $\kappa_\tau(\mathbf{w}) = 1/S$.

\subsection{Analytic inversion via Lagrange--B\"urmann}

We derive an analytic series for the inverse mapping
$\tau \mapsto S(\tau)$ in a neighborhood of the reference point
\[
\tau_0 := f(0)
=
R(\mathbf{w},\mathbf{q}_{\mathrm{ref}}).
\]
Expanding~\eqref{eq:f-series} gives
\[
f(S) = \tau_0 + a_2 S + a_3 S^2 + a_4 S^3 + \cdots,
\qquad
a_2>0.
\]
Since $f$ is analytic and $f'(0)=a_2\neq 0$, it admits a local analytic
inverse near $\tau_0$, which we write as
\begin{equation}
\label{eq:S-series}
S(\tau)
=
\sum_{n=1}^{\infty} c_n \,(\tau-\tau_0)^n.
\end{equation}
Introduce $H$ near zero by $H(0):=1/a_2$ and, for $S\ne0$,
\begin{equation}
\label{eq:H-def}
H(S)
:=
\frac{S}{f(S)-\tau_0}.
\end{equation}
By construction,
\[
f(S)-\tau_0
=
a_2 S + a_3 S^2 + \cdots,
\]
so $H(S)$ is analytic at $S=0$ with 
$$H(S) = \frac{1}{a_2} \cdot \frac{1}{1 + \frac{a_3}{a_2} S + O(S^2)}$$

Applying the geometric series expansion $\frac{1}{1+x} = 1 - x + x^2 - x^3 + \cdots$ with $x = \frac{a_3}{a_2} S + O(S^2)$:

$$H(S) = \frac{1}{a_2}\left(1 - \left(\frac{a_3}{a_2}S + O(S^2)\right) + O(S^2)\right)$$

$$H(S) = \frac{1}{a_2}\left(1 - \frac{a_3}{a_2}S + O(S^2)\right), \qquad H(0) = \frac{1}{a_2}.$$

The Lagrange--B\"urmann inversion theorem yields, for $n\ge 1$,
\begin{equation}
\label{eq:LB-general}
c_n
=
\frac{1}{n!}
\left[
\frac{d^{\,n-1}}{dS^{\,n-1}}
\Big(H(S)^n\Big)
\right]_{S=0}.
\end{equation}
This provides an analytical expression for each coefficient $c_n$ in terms
of the derivatives of $H$ at the origin, hence in terms of the cumulants of $X$.
For illustration, the first coefficients are
\begin{equation}
\label{eq:c123}
c_1 = \frac{1}{a_2},
\qquad
c_2 = -\,\frac{a_3}{a_2^{3}},
\qquad
c_3 = \frac{2a_3^{2} - a_2 a_4}{a_2^{5}}.
\end{equation}

\subsection{Laurent series for \texorpdfstring{$\kappa_\tau(\mathbf{w})$}{kappa}}

Using the shifted variable $z:=\tau-\tau_0$, we write
the inverse series from~\eqref{eq:S-series} as
\[
S(z):=\sum_{n=1}^{\infty}c_nz^n.
\]

The convergence radius of this power series is defined
by the Cauchy--Hadamard formula:
\begin{equation}
\label{eq:Rz_def}
R_z(\mathbf w)
:=
\left(\limsup_{n\to\infty}|c_n|^{1/n}\right)^{-1}
\in(0,\infty],
\end{equation}
with the convention $1/0=\infty$.
Thus, the power series for $S(z)$ converges for
$0<z<R_z(\mathbf w)$.

Since $c_1>0$, the function $S(z)$ is analytic near zero
with $S(0)=0$ and $S'(0)=c_1\neq0$.
Consequently, $\kappa_\tau(\mathbf w)=1/S(z)$ admits
a Laurent expansion with a simple pole at $z=0$.
For sufficiently small $z>0$,
\begin{equation}
\label{eq:kappa-laurent}
\kappa_\tau(\mathbf{w})
=
\frac{1}{S(z)}
=
\sum_{m=-1}^{\infty} d_m z^m,
\qquad
z = \tau-\tau_0,
\end{equation}
where the coefficients $\{d_m\}_{m\ge -1}$ are determined uniquely by
the identity $S(z)\,\kappa_\tau(\mathbf{w}) = 1$.
Multiplying~\eqref{eq:S-series} and~\eqref{eq:kappa-laurent} and matching
coefficients gives the recursion
\[
c_1 d_{-1} = 1
\;\Rightarrow\;
d_{-1} = \frac{1}{c_1},
\]
and for any integer $\ell\ge 0$,
\begin{equation}
\label{eq:d-recursion}
d_\ell
=
-\,\frac{1}{c_1}
\sum_{n=2}^{\ell+2} c_n d_{\ell+1-n}.
\end{equation}
Equations~\eqref{eq:LB-general}, \eqref{eq:S-series},
\eqref{eq:kappa-laurent}, and~\eqref{eq:d-recursion} provide a complete analytic
description of $\kappa_\tau(\mathbf{w})$ as a Laurent series in $z=\tau-\tau_0$.

In particular, the leading term is
\[
\kappa_\tau(\mathbf{w})
\sim
\frac{d_{-1}}{z}
=
\frac{1}{c_1(\tau-\tau_0)}
=
\frac{a_2}{\tau-\tau_0}
=
\frac{\tfrac{1}{2}\,\mathrm{Var}_{\mathbf{q}_{\mathrm{ref}}}
\big(r(\mathbf{w},y_j)\big)}
{\tau - R(\mathbf{w},\mathbf{q}_{\mathrm{ref}})}
\quad
\text{as }
\tau\downarrow\tau_0.
\]
\subsection{Truncation error bounds for the Lagrange--B\"urmann approximation}
\label{app:solve_kappa_trunc}

In implementation we approximate $S(z)$ (hence $\kappa_\tau(\mathbf w)=1/S(z)$)
by truncating the Lagrange--B\"urmann series. The bounds below are non-asymptotic
and decay geometrically with the truncation order.

\paragraph{Truncations.}
For an integer $N\ge 1$, define the order-$N$ truncation of~\eqref{eq:S-series} as
\begin{equation}
\label{eq:S_trunc_def}
S_N(z)
:=
\sum_{n=1}^{N} c_n z^n,
\qquad
R_N^{S}(z)
:=
S(z)-S_N(z).
\end{equation}
For the Laurent series~\eqref{eq:kappa-laurent}, define the order-$N$ truncation
\begin{equation}
\label{eq:kappa_trunc_def}
\kappa_N(z)
:=
\sum_{m=-1}^{N} d_m z^m,
\qquad
R_N^{\kappa}(z)
:=
\kappa_\tau(\mathbf w)-\kappa_N(z).
\end{equation}
Note that computing $\{d_m\}_{m=-1}^N$ via~\eqref{eq:d-recursion} requires only
$\{c_n\}_{n=1}^{N+2}$.

\subsubsection{A geometric tail bound for $S_N$}

\begin{assumption}[Analyticity disk and boundedness of $H$]
\label{ass:H_disk}
Fix $\mathbf w\in\mathcal W_\tau$.
There exist constants $\varrho>0$ and $B_H\in(0,\infty)$
such that $H$ defined in~\eqref{eq:H-def} is analytic
on an open neighborhood of the closed disk
\[
\overline{D_{\varrho}}
:=\{S\in\mathbb C:|S|\le\varrho\}
\]
and satisfies
\begin{equation}
\label{eq:H_bound}
\sup_{|S|\le\varrho}|H(S)|\le B_H.
\end{equation}
\end{assumption}

\begin{remark}
For each fixed $\mathbf w\in\mathcal W_\tau$, let
\[
V(\mathbf w):=
\sum_{j=1}^K(q_{\mathrm{ref}})_j
(\alpha_j-\tau_0)^2
=2a_2>0.
\]
Since $f$ is analytic near zero and
$f'(0)=a_2>0$, $H$ extends analytically to zero
with $H(0)=2/V(\mathbf w)$.
By continuity, Assumption~\ref{ass:H_disk} holds
with $B_H=4/V(\mathbf w)$ for sufficiently small
$\varrho>0$.
\end{remark}

\begin{theorem}[Geometric truncation bound for $S_N$]
\label{thm:S_trunc}
For fixed $\mathbf w\in\mathcal W_\tau$, choose
$\varrho$ and $B_H$ satisfying Assumption~\ref{ass:H_disk},
and define
\begin{equation}
\label{eq:U_def}
U \;:=\; \frac{\varrho}{B_H}.
\end{equation}
Then for every $z\in\mathbb C$ with $|z|<U$, the series~\eqref{eq:S-series}
converges absolutely and, for every integer $N\ge 1$,
\begin{equation}
\label{eq:S_tail_bound}
|S(z)-S_N(z)|
\;\le\;
\frac{\varrho\,\eta(z)^{N+1}}{1-\eta(z)},
\qquad
\eta(z)\;:=\;\frac{|z|}{U}
\;=\;\frac{|z|\,B_H}{\varrho}.
\end{equation}
\end{theorem}

\begin{proof}

For $|z|<U$, on $|S|=\varrho$ we have
$|zH(S)|\le |z|B_H<\varrho=|S|$.
Rouch\'e's theorem therefore gives exactly one simple root of
$S-zH(S)=0$ inside the disk. This root depends analytically on $z$
and extends the local inverse throughout $|z|<U$.

By~\eqref{eq:LB-general} and Cauchy's estimate on $|S|\le\varrho$,
\[
\left|
\left.\frac{d^{\,n-1}}{dS^{\,n-1}}
\Big(H(S)^n\Big)\right|_{S=0}
\right|
\le
\frac{(n-1)!}{\varrho^{n-1}}
\sup_{|S|\le\varrho}|H(S)|^n
\le
\frac{(n-1)!}{\varrho^{n-1}}\,B_H^n.
\]
Dividing by $n!$ yields
\begin{equation}
\label{eq:c_bound}
|c_n|
\le
\frac{1}{n}\,\frac{B_H^n}{\varrho^{n-1}}
\le
\frac{B_H^n}{\varrho^{n-1}}.
\end{equation}
Therefore, for $|z|<\varrho/B_H$,
\[
|R_N^{S}(z)|
=
\left|\sum_{n=N+1}^{\infty} c_n z^n\right|
\le
\sum_{n=N+1}^{\infty} |c_n|\,|z|^n
\le
\sum_{n=N+1}^{\infty} \frac{B_H^n}{\varrho^{n-1}}\,|z|^n
=
\varrho\sum_{n=N+1}^{\infty} \eta(z)^n
=
\frac{\varrho\,\eta(z)^{N+1}}{1-\eta(z)},
\]
which is~\eqref{eq:S_tail_bound}.
\end{proof}
The coefficient bound~\eqref{eq:c_bound} gives
\[
|c_n|
\le
\frac{B_H^n}{n\varrho^{n-1}}
=
\frac{\varrho}{n}
\left(\frac{B_H}{\varrho}\right)^n.
\]
Taking $n$th roots yields
\[
|c_n|^{1/n}
\le
\left(\frac{\varrho}{n}\right)^{1/n}
\frac{B_H}{\varrho}.
\]
Since $\varrho>0$ is fixed,
\[
\left(\frac{\varrho}{n}\right)^{1/n}
=
\exp\left(\frac{\log\varrho-\log n}{n}\right)
\longrightarrow 1.
\]
Therefore,
\[
\limsup_{n\to\infty}|c_n|^{1/n}
\le
\frac{B_H}{\varrho}.
\]
By the Cauchy--Hadamard formula~\eqref{eq:Rz_def},
\[
R_z(\mathbf w)
=
\left(\limsup_{n\to\infty}|c_n|^{1/n}\right)^{-1}
\ge
\frac{\varrho}{B_H}
=
U.
\]
Consequently, $U$ is a guaranteed, possibly conservative
convergence radius for the inverse series; it need not
equal the actual radius $R_z(\mathbf w)$.
\paragraph{Choosing $N$ for a target tolerance on $S$.}
Given $\varepsilon_S>0$ and $\eta(z)\in(0,1)$, it suffices to choose $N$ such that
\begin{equation}
\label{eq:S_chooseN}
\frac{\varrho\,\eta(z)^{N+1}}{1-\eta(z)} \le \varepsilon_S
\quad\Longleftrightarrow\quad
N+1 \ge
\frac{\log\!\big(\varepsilon_S(1-\eta(z))/\varrho\big)}{\log(\eta(z))},
\end{equation}
where $\log(\eta(z))<0$.

\subsubsection{A geometric tail bound for the Laurent truncation of $\kappa$}

Define the analytic function
\begin{equation}
\label{eq:kappa_tilde_def}
\widetilde{\kappa}(z)
:=
z\,\kappa_\tau(\mathbf w)
=
\frac{z}{S(z)}.
\end{equation}
Since $S(z)=c_1 z + O(z^2)$ with $c_1>0$, $\widetilde{\kappa}$ is analytic at $z=0$.

\begin{corollary}[Geometric truncation bound for $\kappa_N$]
\label{cor:kappa_trunc}
For fixed $\mathbf w\in\mathcal W_\tau$, choose
$\varrho$ and $B_H$ satisfying Assumption~\ref{ass:H_disk}. Fix any radius $r\in(0,U)$ for the complex $z$-disk
$\{z\in\mathbb{C}:|z|\le r\}$ (recall $z=\tau-\tau_0$) such that

\begin{equation}
\label{eq:r_condition}
\frac{B_H\,\eta_r}{1-\eta_r} \le \frac{c_1}{2},
\qquad
\eta_r := \frac{r}{U}=\frac{r B_H}{\varrho}.
\end{equation}
Then $\widetilde{\kappa}$ is analytic on $\{|z|\le r\}$ and satisfies
\begin{equation}
\label{eq:kappa_tilde_bound}
\sup_{|z|\le r}|\widetilde{\kappa}(z)| \le \frac{2}{c_1}.
\end{equation}
Moreover, for every $z$ with $0<|z|< r$ and every integer $N\ge -1$,
the Laurent truncation~\eqref{eq:kappa_trunc_def} satisfies
\begin{equation}
\label{eq:kappa_tail_bound}
\big|\kappa_\tau(\mathbf w)-\kappa_N(z)\big|
\le
\frac{2}{c_1}\cdot
\frac{\left(\frac{|z|}{r}\right)^{N+2}}{1-\frac{|z|}{r}}
\cdot
\frac{1}{|z|}.
\end{equation}
\end{corollary}

\begin{proof}
Write $Q(z):=S(z)/z = \sum_{n=1}^\infty c_n z^{n-1}$, so that $Q(0)=c_1$ and
$\widetilde{\kappa}(z)=1/Q(z)$.
Using the coefficient bound~\eqref{eq:c_bound}, for $|z|\le r$ we obtain
\[
|Q(z)-c_1|
\le
\sum_{n=2}^{\infty} |c_n|\,|z|^{n-1}
\le
\sum_{n=2}^{\infty} \frac{B_H^n}{\varrho^{n-1}}r^{\,n-1}
=
B_H\sum_{k=1}^{\infty}\eta_r^{k}
=
\frac{B_H\,\eta_r}{1-\eta_r}.
\]
Under~\eqref{eq:r_condition}, this implies $|Q(z)|\ge c_1/2$ on $|z|\le r$,
hence~\eqref{eq:kappa_tilde_bound}.
Since $\widetilde{\kappa}$ is analytic on $|z|\le r$ and bounded by $2/c_1$ there,
its Taylor remainder after degree $(N+1)$ satisfies the standard geometric estimate
\[
\left|\widetilde{\kappa}(z)-\sum_{m=0}^{N+1} d_{m-1}z^m\right|
\le
\frac{2}{c_1}\cdot
\frac{\left(\frac{|z|}{r}\right)^{N+2}}{1-\frac{|z|}{r}},
\qquad 0<|z|< r,
\]
and dividing by $|z|$ yields~\eqref{eq:kappa_tail_bound} because
$\kappa_N(z)=\sum_{m=-1}^N d_m z^m = \frac{1}{z}\sum_{m=0}^{N+1} d_{m-1} z^m$.
\end{proof}

\paragraph{Choosing $N$ for a target tolerance on $\kappa$.}
Fix $z\neq 0$ with $|z|< r$. Given $\varepsilon_\kappa>0$, it suffices to choose $N$
so that the right-hand side of~\eqref{eq:kappa_tail_bound} is at most $\varepsilon_\kappa$.

\paragraph{Uniformity over iterates.}
Suppose there exist common constants
$(\varrho,B_H)$ such that the analyticity and
boundedness conditions in Assumption~\ref{ass:H_disk}
hold for every iterate, and
$c_1(\mathbf w_t)\ge c_{1,\min}>0$.
Choose a common $r\in(0,U)$ satisfying~\eqref{eq:r_condition}
for every iterate, and assume
$0<|z_t|\le\bar\eta r$ for some $\bar\eta\in(0,1)$, where
$z_t=\tau-R(\mathbf w_t,\mathbf q_{\mathrm{ref}})$.
Writing $\kappa_{N,t}$ for the Laurent truncation at $\mathbf w_t$,
Corollary~\ref{cor:kappa_trunc} gives
\[
\sup_t
\left|\kappa_\tau(\mathbf w_t)-\kappa_{N,t}(z_t)\right|
\le
\frac{2}{c_{1,\min}r}
\frac{\bar\eta^{N+1}}{1-\bar\eta}.
\]

\section{Closed-Form Solution of the Simplex-Constrained QP}
\label{app:qp-solution}

In each iteration of Algorithm~\ref{alg:rse}, the ensemble weights are updated by
solving the convex quadratic program
\begin{equation}
\label{eq:qp-appendix}
\min_{\mathbf{w}\in\mathbb{R}^M}
\ \frac{1}{2}\mathbf{w}^\top \mathbf{H}\mathbf{w} + \mathbf{g}^\top \mathbf{w}
\quad\text{s.t.}\quad
\mathbf{w} \ge 0,\;\; \mathbf{1}^\top \mathbf{w} = 1,
\end{equation}
In our application at iteration $t$, we set $\mathbf{H}=\mathbf{H}_t$ and $\mathbf{g}=\mathbf{g}_t$ as defined below.

Let $\mathbf{w}^\star$ denote the unique minimizer of \eqref{eq:qp-appendix}.
Throughout this appendix, we assume that $H$ is symmetric
positive definite. This holds for the proximal RSE
subproblem in Eq.~(59), since
$H_t=2A_{\mathrm{adv},t}+\rho I\succ0$
when $A_{\mathrm{adv},t}\succeq0$ and $\rho>0$.
We define the active and inactive index sets at the optimum as
\[
S := \{i\in\{1,\dots,M\}: w_i^\star>0\},
\qquad
S^c := \{1,\dots,M\}\setminus S.
\]
For any index sets $I,J \subseteq \{1,\dots,M\}$ we use the standard block notation:
\[
\mathbf{w}_I := (w_i)_{i\in I},\quad
\mathbf{g}_I := (g_i)_{i\in I},\quad
\mathbf{1}_I := (1)_{i\in I},\quad
\mathbf{H}_{IJ} := (H_{ij})_{i\in I, j\in J}.
\]

In our RSE update \eqref{eq:adv-qp-method}, the QP has the form
\[
\min_{\mathbf{w}\in\Delta_M}
\ \mathbf{w}^\top \mathbf{A}_{\mathrm{adv},t}\mathbf{w}
+\mathbf{b}_{\mathrm{adv},t}^\top \mathbf{w}
+\frac{\rho}{2}\|\mathbf{w}-\mathbf{w}_t\|_2^2.
\]

Expanding the proximal term and collecting quadratic/linear parts gives
\eqref{eq:qp-appendix} (at iteration $t$) with $\mathbf{H}=\mathbf{H}_t$ and $\mathbf{g}=\mathbf{g}_t$, where
\begin{equation}
\label{eq:Hg_from_main}
\mathbf{H}_t = 2\mathbf{A}_{\mathrm{adv},t} + \rho \mathbf{I},
\qquad
\mathbf{g}_t = \mathbf{b}_{\mathrm{adv},t} - \rho \mathbf{w}_t,
\end{equation}
(up to an additive constant independent of $\mathbf{w}$).

The feasible region is the probability simplex $\Delta_M$. In our RSE update \eqref{eq:Hg_from_main}, $\rho>0$ implies
\[
\mathbf{H}_t = 2\mathbf{A}_{\mathrm{adv},t} + \rho \mathbf{I} \succ 0,
\]
so problem~\eqref{eq:qp-appendix} is strictly convex and therefore has a unique global minimizer, denoted $\mathbf{w}^\star$.
Since $\mathbf{H}_t\succ 0$, every principal submatrix is positive definite; hence $(\mathbf{H}_t)_{SS}\succ 0$ and is invertible.

Geometrically, this step projects the unconstrained quadratic minimizer onto the
simplex. Indeed, since \(\mathbf H\succ0\), the unconstrained minimizer is
\(-\mathbf H^{-1}\mathbf g\), and completing the square gives
\[
\frac12\mathbf w^\top\mathbf H\mathbf w+\mathbf g^\top\mathbf w
=
\frac12
\left\|
\mathbf w+\mathbf H^{-1}\mathbf g
\right\|_{\mathbf H}^2
-
\frac12\mathbf g^\top\mathbf H^{-1}\mathbf g,
\]
where \(\|\mathbf u\|_{\mathbf H}^2:=\mathbf u^\top\mathbf H\mathbf u\). Hence
\[
\mathbf w^\star
=
\arg\min_{\mathbf w\in\Delta_M}
\left\|
\mathbf w-(-\mathbf H^{-1}\mathbf g)
\right\|_{\mathbf H}^2,
\]
i.e., \(\mathbf w^\star\) is the projection of
\(-\mathbf H^{-1}\mathbf g\) onto \(\Delta_M\) in the \(\mathbf H\)-norm.

\subsection{KKT Conditions and Active Set Framework}
\label{app:qp-kkt}

The Lagrangian of~\eqref{eq:qp-appendix} is
\[
\mathcal{L}(\mathbf{w},\lambda,\boldsymbol{\mu})
=
\frac12\mathbf{w}^\top\mathbf{H}\mathbf{w}
+
\mathbf{g}^\top\mathbf{w}
+
\lambda(\mathbf{1}^\top\mathbf{w}-1)
-
\boldsymbol{\mu}^\top\mathbf{w},
\]
where
\begin{itemize}
    \item $\lambda\in\mathbb{R}$ is the multiplier for the equality constraint
    $\mathbf{1}^\top\mathbf{w}=1$;
    \item $\boldsymbol{\mu}\in\mathbb{R}^M_{\ge0}$ are the multipliers for
    the nonnegativity constraints $w_i \ge 0$.
\end{itemize}

The full KKT conditions are
\begin{align}
\mathbf{H}\mathbf{w}+\mathbf{g}+\lambda\mathbf{1}-\boldsymbol{\mu} &= 0,
\label{eq:kkt-grad}\\
\mathbf{w} \ge 0,\quad \boldsymbol{\mu}\ge 0,\qquad
\mu_i w_i &= 0\;\;\forall i,
\label{eq:kkt-comp}\\
\mathbf{1}^\top \mathbf{w} &= 1.
\label{eq:kkt-sum}
\end{align}

Complementary slackness~\eqref{eq:kkt-comp} implies
\[
i\in S \Rightarrow \mu_i = 0, \qquad
i\in S^c \Rightarrow w_i = 0,\;\; \mu_i \ge 0.
\]

Splitting the gradient condition~\eqref{eq:kkt-grad} yields:
\begin{align}
\mathbf{H}_{SS}\mathbf{w}_S + \mathbf{g}_S + \lambda \mathbf{1}_S &= 0,
\label{eq:kkt-act}\\
\mathbf{H}_{S^c S}\mathbf{w}_S + \mathbf{g}_{S^c} + \lambda \mathbf{1}_{S^c}
&= \boldsymbol{\mu}_{S^c} \ge 0.
\label{eq:kkt-inact}
\end{align}
The equality constraint reduces to
\begin{equation}
\mathbf{1}_S^\top \mathbf{w}_S = 1.
\label{eq:kkt-sumS}
\end{equation}

Equations~\eqref{eq:kkt-act} and~\eqref{eq:kkt-sumS} produce the linear system
\begin{equation}
\label{eq:closed-kkt-system}
\begin{bmatrix}
\mathbf{H}_{SS} & \mathbf{1}_S \\
\mathbf{1}_S^\top & 0
\end{bmatrix}
\begin{bmatrix}
\mathbf{w}_S \\[2pt] \lambda
\end{bmatrix}
=
\begin{bmatrix}
-\mathbf{g}_S \\[2pt] 1
\end{bmatrix}.
\tag{A1}
\end{equation}
The inactive coordinates are fixed at $\mathbf{w}_{S^c}=0$, and feasibility
requires $\boldsymbol{\mu}_{S^c} \ge 0$ via~\eqref{eq:kkt-inact}.

Thus, solving the QP reduces to identifying the correct active set $S$ and then
solving \eqref{eq:closed-kkt-system}.

\subsection{Closed-Form Analytical Solution for a Fixed Active Set}
\label{app:qp-closed-form}

System~\eqref{eq:closed-kkt-system} admits the explicit closed-form solution:
\begin{align}
\lambda^\star
&=
\frac{
\mathbf{1}_S^\top \mathbf{H}_{SS}^{-1}(-\mathbf{g}_S) - 1
}{
\mathbf{1}_S^\top \mathbf{H}_{SS}^{-1}\mathbf{1}_S
},
\tag{A2}
\label{eq:closed-lambda}
\\[5pt]
\mathbf{w}_S^\star
&=
-\mathbf{H}_{SS}^{-1}\left(\mathbf{g}_S + \lambda^\star \mathbf{1}_S\right),
\tag{A3}
\label{eq:closed-w}
\end{align}
with
\[
\mathbf{w}_{S^c}^\star = \mathbf{0}.
\]

These equations follow directly from block elimination (Schur complement)
applied to~\eqref{eq:closed-kkt-system}.  
When a candidate active set $S$ is correct, $\mathbf{w}_S^\star$ computed from
\eqref{eq:closed-w} satisfies
\[
\mathbf{w}_S^\star > 0,\qquad
\boldsymbol{\mu}_{S^c} = \mathbf{H}_{S^cS}\mathbf{w}_S^\star + \mathbf{g}_{S^c}
+ \lambda^\star \mathbf{1}_{S^c} \ge 0.
\]

If any component of $\mathbf{w}_S^\star$ is nonpositive, the candidate
active set $S$ does not satisfy the KKT conditions and must be revised.

In our implementation we solve \eqref{eq:qp-appendix} using a standard convex QP solver; \eqref{eq:closed-lambda}-\eqref{eq:closed-w} provide the KKT characterization of the unique minimizer.

\section{Convergence of the Proximal RSE Alternating Procedure}
\label{app:rse_convergence}

This appendix analyzes the proximal offline RSE iteration in Algorithm~\ref{alg:rse}.
We first differentiate the KL fragility root $\kappa_\tau(\mathbf{w})$ defined by $\Phi(\kappa;\mathbf{w})=\tau$,
and relate $\nabla_{\mathbf{w}}\kappa_\tau(\mathbf{w})$ to the gradient of the adversarially tilted quadratic risk.
We then show that the proximal QP step yields a sufficient decrease in $\kappa_\tau(\mathbf{w}_t)$ and that every limit point
satisfies the simplex stationarity conditions. A stopping certificate follows from the QP KKT conditions.

\subsection{Standing definitions and well-posed regime}
\label{app:rse_convergence:setup}

Recall the class-conditional risks $r(\mathbf{w},y_j)$ and the risk vector
$\mathbf{r}(\mathbf{w})=(r(\mathbf{w},y_1),\dots,r(\mathbf{w},y_K))^\top$
defined in \eqref{eq:class_conditional_risk_main}--\eqref{eq:r(w)}.
Under the Brier-score quadratic model (Section~\ref{subsec:quad-risk}), each
$r(\mathbf{w},y_j)$ is the convex quadratic \eqref{eq:quad-risk-matrix}:
\[
r(\mathbf{w},y_j) = \mathbf{w}^\top \mathbf{A}_j \mathbf{w} + \mathbf{b}_j^\top \mathbf{w} + 1,
\qquad \mathbf{A}_j\succeq 0.
\]

For $\kappa>0$, define the KL-penalized inner value (as in \eqref{eq:penalized-kl})
\begin{equation}
\label{eq:Phi_kappa_w_appconv}
\Phi(\kappa;\mathbf{w})
:=
\sup_{\mathbf{q}\in\Delta_K}
\Big\{
\mathbf{q}^\top \mathbf{r}(\mathbf{w})
-\kappa\,D_{\mathrm{KL}}(\mathbf{q}\,\|\,\mathbf{q}_{\mathrm{ref}})
\Big\}.
\end{equation}
By Lemma~\ref{lem:kl-inner}, $\Phi$ has the closed form \eqref{eq:Phi-def}, and the
unique maximizer is the exponential tilt $\mathbf{q}^\star(\kappa,\mathbf{w})$ from
\eqref{eq:qstar-softmax-method}.

We recall the well-posed regime $\mathcal W_\tau$ and the extended definition of $\kappa_\tau(\mathbf w)$ from
Section~\ref{sec:problem_formulation}. Throughout this appendix we work on $\mathcal W_\tau$ (or a compact subset $\mathcal K\subset\mathcal W_\tau$).

\begin{assumption}[Iterates remain in a compact well-posed region]
\label{ass:compact_wellposed}
Assume $(q_{\mathrm{ref}})_j>0$ for all $j$.
Moreover, there exists a compact set $\mathcal K\subset \mathcal W_\tau$
such that the iterates satisfy $\mathbf w_t\in \mathcal K$ for all $t$.
\end{assumption}

\begin{remark}
The standing full-support assumption ensures that every
class is included in the KL adversary's support.
The compactness condition in Assumption~I.1 keeps the
iterates away from the boundary of the well-posed
region $\mathcal W_\tau$.
\end{remark}

\begin{lemma}[Uniform boundedness of the class-conditional risks]
\label{lem:risk_bounded}
Under the Brier score, for every $\mathbf w\in\Delta_M$ and every class $j$,
\[
0 \le r(\mathbf w,y_j)\le 2.
\]
Consequently, for every $\mathbf q\in\Delta_K$, $0\le R(\mathbf w,\mathbf q)\le 2$.
\end{lemma}

\begin{proof}
For a probability vector $\widehat{\mathbf p}\in\Delta_K$ and true class $y_j$,
the Brier loss is
$L(y_j,\widehat{\mathbf p})=\sum_{k=1}^K(\widehat p_k-\mathbf 1\{k=j\})^2$.
It is always nonnegative. Its maximum over $\Delta_K$ is achieved at a vertex
$\widehat{\mathbf p}=\mathbf e_\ell$ with $\ell\neq j$, giving
$L= (0-1)^2+(1-0)^2=2$. Taking expectations yields the bounds for $r$ and $R$.
\end{proof}

\begin{lemma}[Uniform gap and bounded fragility on compact subsets of $\mathcal W_\tau$]
\label{lem:uniform_gap_bounds}
Under Assumption~\ref{ass:compact_wellposed}, there exist constants
$0<\underline\kappa\le \overline\kappa<\infty$ and $\delta>0$ such that for all
$\mathbf w\in\mathcal K$,
\[
\kappa_\tau(\mathbf w)\in[\underline\kappa,\overline\kappa],
\qquad
\mathbb E_{\mathbf q^\star(\kappa_\tau(\mathbf w),\mathbf w)}[r(\mathbf w,Y)]-\tau \ge \delta.
\]
Equivalently, $\partial_\kappa \Phi(\kappa_\tau(\mathbf w);\mathbf w)$ is uniformly bounded away from $0$ on $\mathcal K$.
\end{lemma}

\begin{proof}
On $\mathcal W_\tau$, $\kappa_\tau(\mathbf w)$ is the unique root of
$\Phi(\kappa;\mathbf w)=\tau$ with $\kappa\in(0,\infty)$, and
$\kappa\mapsto \Phi(\kappa;\mathbf w)$ is continuous and strictly decreasing.
Standard root-continuity arguments imply $\mathbf w\mapsto \kappa_\tau(\mathbf w)$
is continuous on $\mathcal W_\tau$. Hence it is continuous on the compact set
$\mathcal K$, so it attains a finite minimum and maximum:
$\underline\kappa:=\min_{\mathcal K}\kappa_\tau(\mathbf w)>0$ and
$\overline\kappa:=\max_{\mathcal K}\kappa_\tau(\mathbf w)<\infty$.

Define
$g(\mathbf w):=\mathbb E_{\mathbf q^\star(\kappa_\tau(\mathbf w),\mathbf w)}[r(\mathbf w,Y)]-\tau$.
Fix any $\mathbf w\in\mathcal W_\tau$ and let
$\kappa=\kappa_\tau(\mathbf w)$ and $\mathbf q^\star=\mathbf q^\star(\kappa,\mathbf w)$.
Since $\mathbf q^\star$ attains the supremum in \eqref{eq:Phi_kappa_w_appconv}, at the root
$\Phi(\kappa;\mathbf w)=\tau$ we have
\[
\tau
=
\Phi(\kappa;\mathbf w)
=
\mathbb E_{\mathbf q^\star}[r(\mathbf w,Y)]
-\kappa\,D_{\mathrm{KL}}(\mathbf q^\star\|\mathbf q_{\mathrm{ref}}),
\]
hence
\[
g(\mathbf w)=\mathbb E_{\mathbf q^\star}[r(\mathbf w,Y)]-\tau
=
\kappa\,D_{\mathrm{KL}}(\mathbf q^\star\|\mathbf q_{\mathrm{ref}}).
\]
Moreover, on $\mathcal W_\tau$ we have $R(\mathbf w,\mathbf q_{\mathrm{ref}})<\tau$, so the maximizer
at the root cannot be $\mathbf q^\star=\mathbf q_{\mathrm{ref}}$; therefore
$D_{\mathrm{KL}}(\mathbf q^\star\|\mathbf q_{\mathrm{ref}})>0$, and since $\kappa>0$ on $\mathcal W_\tau$,
we conclude $g(\mathbf w)>0$ for all $\mathbf w\in\mathcal W_\tau$.
By continuity of $g$ and compactness of $\mathcal K$, it follows that
\[
\delta:=\min_{\mathbf w\in\mathcal K} g(\mathbf w) > 0.
\]

\end{proof}

\begin{lemma}[Lower bound on the scaling $c(\mathbf w)$]
\label{lem:cmin}
Under Assumption~\ref{ass:compact_wellposed}, define
\[
c(\mathbf w):=\frac{\kappa_\tau(\mathbf w)}{\mathbb E_{\mathbf q^\star}[r(\mathbf w,Y)]-\tau},
\quad
\mathbf q^\star=\mathbf q^\star(\kappa_\tau(\mathbf w),\mathbf w).
\]
Then $c(\mathbf w)$ is continuous on $\mathcal K$ and admits a uniform lower bound
\[
c(\mathbf w)\ge c_{\min}:=\min_{\mathbf u\in\mathcal K} c(\mathbf u) >0.
\]
Moreover, using Lemma~\ref{lem:risk_bounded} one has the explicit bound
\[
c(\mathbf w)\ \ge\ \frac{\underline\kappa}{2-\tau}
\qquad\text{for all }\mathbf w\in\mathcal K,
\]
where $\underline\kappa$ is from Lemma~\ref{lem:uniform_gap_bounds}.
\end{lemma}

\begin{proof}
By Lemma~\ref{lem:uniform_gap_bounds}, the denominator is uniformly $\ge\delta>0$
on $\mathcal K$ and $\kappa_\tau(\mathbf w)$ is continuous and positive on $\mathcal K$.
Thus $c(\mathbf w)$ is continuous and positive on compact $\mathcal K$, so it attains
a positive minimum.

For the explicit bound, Lemma~\ref{lem:risk_bounded} gives
$\mathbb E_{\mathbf q^\star}[r(\mathbf w,Y)]\le 2$, hence the denominator
$\le 2-\tau$. Using $\kappa_\tau(\mathbf w)\ge\underline\kappa$ yields
$c(\mathbf w)\ge \underline\kappa/(2-\tau)$.
\end{proof}

\begin{remark}[Interpretation of \(c_{\min}\)]
The constant \(c_{\min}\) is a local conditioning constant for the convergence
proof, not a hyperparameter used by the algorithm. At the exact root,
\[
\tau
=
\mathbb E_{\mathbf q^\star}[r(\mathbf w,Y)]
-
\kappa_\tau(\mathbf w)
D_{\mathrm{KL}}(\mathbf q^\star\|\mathbf q_{\mathrm{ref}}),
\]
and therefore
\[
c(\mathbf w)
=
\frac{\kappa_\tau(\mathbf w)}
{\mathbb E_{\mathbf q^\star}[r(\mathbf w,Y)]-\tau}
=
\frac{1}
{D_{\mathrm{KL}}(\mathbf q^\star\|\mathbf q_{\mathrm{ref}})}.
\]
Hence \(c_{\min}\) can be small when the adversarial tilt
\(\mathbf q^\star\) is very far from \(\mathbf q_{\mathrm{ref}}\), for example
when \(\mathbf q_{\mathrm{ref}}\) assigns very small mass to a class. However,
for a fixed full-support reference distribution with
\[
q_{\min}:=\min_j(q_{\mathrm{ref}})_j>0,
\]
we have
\[
D_{\mathrm{KL}}(\mathbf q^\star\|\mathbf q_{\mathrm{ref}})
\le
\log\frac{1}{q_{\min}},
\]
and therefore
\[
c(\mathbf w)
\ge
\frac{1}{\log(1/q_{\min})}.
\]
Thus \(c_{\min}>0\) under the stated full-support and compact well-posedness
assumptions. The magnitude of the lower bound only affects the conservativeness
of the sufficient condition on \(\rho\); it does not change the algorithm or the
validity of the convergence statement under the stated assumptions.
\end{remark}

\begin{lemma}[Lipschitz continuity of $\nabla \kappa_\tau$ on $\mathcal K$]
\label{lem:kappa_grad_lipschitz}
Under Assumption~\ref{ass:compact_wellposed}, the map $\mathbf w\mapsto \kappa_\tau(\mathbf w)$
is continuously differentiable on $\mathcal K$. Moreover, there exists $L<\infty$ such that
\[
\|\nabla \kappa_\tau(\mathbf w)-\nabla \kappa_\tau(\mathbf w')\|_2
\le L\|\mathbf w-\mathbf w'\|_2
\qquad \forall \mathbf w,\mathbf w'\in\mathcal K.
\]
\end{lemma}

\begin{proof}

The proof has three steps: (i) show that $\kappa_\tau$ is a smooth function of
$\mathbf w$ on an open set containing $\mathcal K$; (ii) use compactness to bound
its gradient and Hessian on a slightly enlarged set $\mathcal K_\epsilon$; and
(iii) derive the Lipschitz bound by treating nearby and distant pairs of points
separately, which avoids any convexity assumption on $\mathcal K$.

\paragraph{Step 1: smoothness of $\kappa_\tau$.}
Define
\[
F(\mathbf u,\kappa):=\Phi(\kappa;\mathbf u)-\tau.
\]
The class risks are quadratic polynomials in
$\mathbf u$, and
\[
\Phi(\kappa;\mathbf u)
=
\kappa\log\sum_j(q_{\mathrm{ref}})_j
\exp\bigl(r(\mathbf u,y_j)/\kappa\bigr).
\]
Thus, $F$ is $C^\infty$ on
$\mathbb R^M\times(0,\infty)$: the denominator
$\kappa$ is nonzero and the logarithm's argument
is positive.

Fix $\mathbf w_0\in\mathcal K$ and write
$\kappa_0:=\kappa_\tau(\mathbf w_0)>0$.
By definition,
\[
F(\mathbf w_0,\kappa_0)=0.
\]
Moreover, Lemma~\ref{lem:uniform_gap_bounds} gives
\[
\left|\partial_\kappa F(\mathbf w_0,\kappa_0)\right|
=
\frac{
\mathbb E_{\mathbf q^\star(\kappa_0,\mathbf w_0)}
[r(\mathbf w_0,Y)]-\tau
}{\kappa_0}
\ge\frac{\delta}{\overline\kappa}>0.
\]
The smooth implicit function theorem therefore
provides a $C^\infty$ function $g$ near $\mathbf w_0$
such that
\[
g(\mathbf w_0)=\kappa_0,
\qquad
\Phi(g(\mathbf u);\mathbf u)=\tau.
\]
By continuity, we may restrict this neighborhood
so that $g(\mathbf u)>0$ and
\[
\sum_j(q_{\mathrm{ref}})_j r(\mathbf u,y_j)
<\tau<\max_j r(\mathbf u,y_j)
\]
throughout it. Here, a neighborhood consists of
vectors satisfying $\|\mathbf u-\mathbf w_0\|_2<h$
for a sufficiently small $h>0$.

Under these inequalities, the positive-root
uniqueness argument of
Lemma~\ref{lem:kl-fragility-scalar} applies.
Hence $g$ agrees with $\kappa_\tau$ at feasible
weights. Repeating the construction around every
point of $\mathcal K$ gives local smooth functions
that agree wherever their neighborhoods overlap,
because the positive root is unique.
They therefore define one $C^\infty$ function on
an open set $\mathcal O\supset\mathcal K$.

Thus, $\kappa_\tau$ can be viewed as a smooth
function on $\mathcal O$, with its values on
$\mathcal K$ unchanged.
The set $\mathcal O$ may include points outside
the simplex, where the quadratic risk formulas
remain defined. This allows us to work with
gradients and Hessians in a neighborhood of
$\mathcal K$; the ensemble weights remain
constrained to $\Delta_M$.

\paragraph{Step 2: uniform bounds on a compact enlargement of $\mathcal K$.}
Since $\mathcal K$ is compact and $\mathcal O$ is an open set containing it,
there is $\epsilon>0$ such that the closed $\epsilon$-enlargement
\[
\mathcal K_\epsilon
:=
\{\mathbf u\in\mathbb R^M:
\operatorname{dist}(\mathbf u,\mathcal K)\le\epsilon\}
\]
still lies in $\mathcal O$. The set $\mathcal K_\epsilon$ is closed and bounded,
hence compact. Since $\nabla\kappa_\tau$ and $\nabla^2\kappa_\tau$ are continuous
on $\mathcal O$, they attain finite maxima on $\mathcal K_\epsilon$:
\[
H:=\sup_{\mathbf u\in\mathcal K_\epsilon}
\|\nabla^2\kappa_\tau(\mathbf u)\|_{\mathrm{op}}<\infty,
\qquad
G:=\sup_{\mathbf u\in\mathcal K_\epsilon}
\|\nabla\kappa_\tau(\mathbf u)\|_2<\infty.
\]
Here $H$ bounds how fast the gradient can change, and $G$ bounds the size of
the gradient itself.

\paragraph{Step 3: the Lipschitz bound.}
The set $\mathcal K$ need not be convex (indeed, $\mathcal W_\tau$ is not convex
in general), so the straight segment between two points of $\mathcal K$ may
leave $\mathcal K$. We therefore distinguish two cases.

\emph{Case 1: nearby points, $\|\mathbf w-\mathbf w'\|_2\le\epsilon$.}
Every point $\mathbf w+s(\mathbf w'-\mathbf w)$, $s\in[0,1]$, on the segment is
at distance at most $\|\mathbf w-\mathbf w'\|_2\le\epsilon$ from
$\mathbf w\in\mathcal K$, so the whole segment lies in $\mathcal K_\epsilon$,
where $\kappa_\tau$ is smooth. By the fundamental theorem of calculus applied
to $\nabla\kappa_\tau$ along the segment,
\[
\nabla\kappa_\tau(\mathbf w')-\nabla\kappa_\tau(\mathbf w)
=\int_0^1\nabla^2\kappa_\tau\big(\mathbf w+s(\mathbf w'-\mathbf w)\big)
(\mathbf w'-\mathbf w)\,ds,
\]
and bounding the Hessian by $H$ gives
\[
\|\nabla\kappa_\tau(\mathbf w)
-\nabla\kappa_\tau(\mathbf w')\|_2
\le H\|\mathbf w-\mathbf w'\|_2.
\]

\emph{Case 2: distant points, $\|\mathbf w-\mathbf w'\|_2>\epsilon$.}
Here the segment may leave $\mathcal K_\epsilon$, so we do not use the Hessian.
Instead, by the triangle inequality and the gradient bound $G$,
\[
\|\nabla\kappa_\tau(\mathbf w)
-\nabla\kappa_\tau(\mathbf w')\|_2
\le\|\nabla\kappa_\tau(\mathbf w)\|_2+\|\nabla\kappa_\tau(\mathbf w')\|_2
\le 2G.
\]
Since $\|\mathbf w-\mathbf w'\|_2/\epsilon>1$ in this case,
\[
2G\le\frac{2G}{\epsilon}\|\mathbf w-\mathbf w'\|_2 .
\]
That is, a large gradient difference is acceptable here because the two points
are also far apart.

\paragraph{Conclusion.}
Combining the two cases, the Lipschitz bound holds for all
$\mathbf w,\mathbf w'\in\mathcal K$ with
\[
L=\max\Big\{H,\ \frac{2G}{\epsilon}\Big\}<\infty .
\]
Continuous differentiability of $\kappa_\tau$ on $\mathcal K$ follows from
Step~1. No convexity of $\mathcal K$ is required: convexity is used only
locally, on segments of length at most $\epsilon$, which always lie in
$\mathcal K_\epsilon$.

\end{proof}

\subsection{Differentiability of $\Phi$ and $\kappa_\tau(\mathbf{w})$}
\label{app:rse_convergence:smoothness}

Define the normalizer
\[
Z(\kappa,\mathbf{w})
:=
\sum_{j=1}^K (q_{\mathrm{ref}})_j \exp\!\big(r(\mathbf{w},y_j)/\kappa\big),
\]
so that $\Phi(\kappa;\mathbf{w})=\kappa\log Z(\kappa,\mathbf{w})$ by \eqref{eq:Phi-def}.
Also define the KL-adversary (Lemma~\ref{lem:kl-inner})
\begin{equation}
\label{eq:qstar_def_appconv}
(q^\star(\kappa,\mathbf{w}))_j
=
\frac{(q_{\mathrm{ref}})_j \exp\!\big(r(\mathbf{w},y_j)/\kappa\big)}{Z(\kappa,\mathbf{w})}.
\end{equation}

\begin{lemma}[Gradients of $\Phi$]
\label{lem:Phi_grad_appconv}
For any $\kappa>0$ and $\mathbf{w}\in\Delta_M$, the function $\Phi(\kappa;\mathbf{w})$ is continuously
differentiable in $(\kappa,\mathbf{w})$ and satisfies
\begin{align}
\nabla_{\mathbf{w}}\Phi(\kappa;\mathbf{w})
&=
\sum_{j=1}^K (q^\star(\kappa,\mathbf{w}))_j \,\nabla_{\mathbf{w}} r(\mathbf{w},y_j),
\label{eq:gradw_Phi_appconv}
\\
\frac{\partial}{\partial \kappa}\Phi(\kappa;\mathbf{w})
&=
\log Z(\kappa,\mathbf{w})
-\frac{1}{\kappa}\sum_{j=1}^K (q^\star(\kappa,\mathbf{w}))_j\, r(\mathbf{w},y_j)
=
\frac{\Phi(\kappa;\mathbf{w})-\mathbb{E}_{\mathbf{q}^\star}[r(\mathbf{w},Y)]}{\kappa}.
\label{eq:gradk_Phi_appconv}
\end{align}
\end{lemma}

\begin{proof}
Using $\Phi(\kappa;\mathbf{w})=\kappa\log Z(\kappa,\mathbf{w})$ and the chain rule,
\[
\nabla_{\mathbf{w}}\Phi(\kappa;\mathbf{w})
=
\kappa\frac{1}{Z(\kappa,\mathbf{w})}\nabla_{\mathbf{w}} Z(\kappa,\mathbf{w}).
\]
Since
\[
\nabla_{\mathbf{w}} Z(\kappa,\mathbf{w})
=
\sum_{j=1}^K (q_{\mathrm{ref}})_j \exp(r(\mathbf{w},y_j)/\kappa)\cdot \frac{1}{\kappa}\nabla_{\mathbf{w}} r(\mathbf{w},y_j),
\]
substitution yields \eqref{eq:gradw_Phi_appconv} after recognizing \eqref{eq:qstar_def_appconv}.

For the $\kappa$-derivative, write
\[
\frac{\partial}{\partial\kappa}\Phi(\kappa;\mathbf{w})
=
\log Z(\kappa,\mathbf{w})
+
\kappa\frac{1}{Z(\kappa,\mathbf{w})}\frac{\partial}{\partial\kappa}Z(\kappa,\mathbf{w}).
\]
Because $\frac{\partial}{\partial\kappa}\exp(r/\kappa)=\exp(r/\kappa)\cdot(-r/\kappa^2)$, we have
\[
\frac{\partial}{\partial\kappa}Z(\kappa,\mathbf{w})
=
-\frac{1}{\kappa^2}\sum_{j=1}^K (q_{\mathrm{ref}})_j \exp(r(\mathbf{w},y_j)/\kappa)\,r(\mathbf{w},y_j),
\]
and substituting gives the first equality in \eqref{eq:gradk_Phi_appconv}. The last equality follows from
$\Phi(\kappa;\mathbf{w})=\kappa\log Z(\kappa,\mathbf{w})$ and the definition of $\mathbf{q}^\star$.
\end{proof}

\begin{lemma}[Implicit gradient of $\kappa_\tau(\mathbf{w})$]
\label{lem:kappa_grad_appconv}
Under Assumption~\ref{ass:compact_wellposed}, the function $\mathbf{w}\mapsto \kappa_\tau(\mathbf{w})$
is continuously differentiable along the iterates $\{\mathbf{w}_t\}$, and for any $\mathbf{w}\in\mathcal{W}_\tau$:
\begin{equation}
\label{eq:kappa_grad_formula_appconv}
\nabla_{\mathbf{w}}\kappa_\tau(\mathbf{w})
=
-\frac{\nabla_{\mathbf{w}}\Phi(\kappa_\tau(\mathbf{w});\mathbf{w})}{\partial_\kappa \Phi(\kappa_\tau(\mathbf{w});\mathbf{w})}
=
\frac{\kappa_\tau(\mathbf{w})}{\mathbb{E}_{\mathbf{q}^\star}[r(\mathbf{w},Y)]-\tau}
\sum_{j=1}^K (q^\star)_j \,\nabla_{\mathbf{w}} r(\mathbf{w},y_j),
\end{equation}
where $\mathbf{q}^\star=\mathbf{q}^\star(\kappa_\tau(\mathbf{w}),\mathbf{w})$.
Moreover,
\begin{equation}
\label{eq:Phi_kappa_negative_appconv}
\partial_\kappa \Phi(\kappa_\tau(\mathbf{w});\mathbf{w})<0
\quad\text{and thus}\quad
\mathbb{E}_{\mathbf{q}^\star}[r(\mathbf{w},Y)]>\tau
\qquad\text{for all }\mathbf{w}\in\mathcal{W}_\tau.
\end{equation}
\end{lemma}

\begin{proof}
Fix $\mathbf{w}\in\mathcal{W}_\tau$ and define $F(\kappa,\mathbf{w}):=\Phi(\kappa;\mathbf{w})-\tau$.
By Lemma~\ref{lem:Phi_grad_appconv}, $F$ is $C^1$. Since $\kappa_\tau(\mathbf{w})$ satisfies
$F(\kappa_\tau(\mathbf{w}),\mathbf{w})=0$, the implicit function theorem implies differentiability provided
$\partial_\kappa \Phi(\kappa_\tau(\mathbf{w});\mathbf{w})\neq 0$.

At $\kappa=\kappa_\tau(\mathbf{w})$, we have
$\Phi(\kappa;\mathbf{w})=\tau$. Let
\(\mathbf q^\star=\mathbf q^\star(\kappa,\mathbf w)\). Then
\[
\tau
=
\Phi(\kappa;\mathbf{w})
=
\mathbf{q}^{\star\top}\mathbf{r}(\mathbf{w})
-\kappa\,D_{\mathrm{KL}}(\mathbf{q}^\star\|\mathbf{q}_{\mathrm{ref}}).
\]
We claim that
\[
D_{\mathrm{KL}}(\mathbf q^\star\|\mathbf q_{\mathrm{ref}})>0.
\]
Indeed, if \(D_{\mathrm{KL}}(\mathbf q^\star\|\mathbf q_{\mathrm{ref}})=0\), then
\(\mathbf q^\star=\mathbf q_{\mathrm{ref}}\). Hence
\[
\tau
=
\Phi(\kappa;\mathbf w)
=
R(\mathbf w,\mathbf q_{\mathrm{ref}}),
\]
contradicting \(\mathbf w\in\mathcal W_\tau\), which requires
\[
R(\mathbf w,\mathbf q_{\mathrm{ref}})<\tau.
\]
Therefore \(D_{\mathrm{KL}}(\mathbf q^\star\|\mathbf q_{\mathrm{ref}})>0\). Since
\(\kappa=\kappa_\tau(\mathbf w)>0\), we obtain
\[
\tau
=
\mathbf{q}^{\star\top}\mathbf{r}(\mathbf{w})
-\kappa\,D_{\mathrm{KL}}(\mathbf{q}^\star\|\mathbf{q}_{\mathrm{ref}})
<
\mathbf{q}^{\star\top}\mathbf{r}(\mathbf{w})
=
\mathbb{E}_{\mathbf{q}^\star}[r(\mathbf{w},Y)].
\]

Therefore
$\mathbb{E}_{\mathbf{q}^\star}[r(\mathbf{w},Y)]>\tau$, and substituting into \eqref{eq:gradk_Phi_appconv} gives
$\partial_\kappa\Phi(\kappa_\tau(\mathbf{w});\mathbf{w})=(\tau-\mathbb{E}_{\mathbf{q}^\star}[r])/\kappa_\tau(\mathbf{w})<0$,
establishing \eqref{eq:Phi_kappa_negative_appconv}. 

Finally, differentiate the identity
$
F(\kappa_\tau(\mathbf{w}),\mathbf{w})=\Phi(\kappa_\tau(\mathbf{w});\mathbf{w})-\tau=0
$
with respect to $\mathbf{w}$. By the chain rule,
\[
\mathbf{0}
=
\nabla_{\mathbf{w}} F(\kappa_\tau(\mathbf{w}),\mathbf{w})
=
\nabla_{\mathbf{w}}\Phi(\kappa_\tau(\mathbf{w});\mathbf{w})
+
\partial_\kappa \Phi(\kappa_\tau(\mathbf{w});\mathbf{w})\;\nabla_{\mathbf{w}}\kappa_\tau(\mathbf{w}).
\]
Since $\partial_\kappa \Phi(\kappa_\tau(\mathbf{w});\mathbf{w})\neq 0$ (indeed it is negative by
\eqref{eq:Phi_kappa_negative_appconv}), we can solve for $\nabla_{\mathbf{w}}\kappa_\tau(\mathbf{w})$:
\[
\nabla_{\mathbf{w}}\kappa_\tau(\mathbf{w})
=
-\frac{\nabla_{\mathbf{w}}\Phi(\kappa_\tau(\mathbf{w});\mathbf{w})}
{\partial_\kappa \Phi(\kappa_\tau(\mathbf{w});\mathbf{w})}.
\]
Using Lemma~\ref{lem:Phi_grad_appconv} gives
$\nabla_{\mathbf{w}}\Phi(\kappa_\tau(\mathbf{w});\mathbf{w})
=
\sum_{j=1}^K (q^\star)_j \nabla_{\mathbf{w}} r(\mathbf{w},y_j)$,
and at the root $\Phi(\kappa_\tau(\mathbf{w});\mathbf{w})=\tau$ we also have from \eqref{eq:gradk_Phi_appconv}
\[
\partial_\kappa \Phi(\kappa_\tau(\mathbf{w});\mathbf{w})
=
\frac{\tau-\mathbb{E}_{\mathbf{q}^\star}[r(\mathbf{w},Y)]}{\kappa_\tau(\mathbf{w})}.
\]
Substituting these two identities into the previous display yields \eqref{eq:kappa_grad_formula_appconv}.

\end{proof}

\subsection{Proximal QP update, descent, and a stopping certificate}
\label{app:rse_convergence:descent}

At iteration $t$, Algorithm~\ref{alg:rse} forms the adversary
$\mathbf{q}_t^\star=\mathbf{q}^\star(\kappa_t,\mathbf{w}_t)$ where
$\kappa_t=\kappa_\tau(\mathbf{w}_t)$, and defines the tilted (adversarial) quadratic risk
\begin{equation}
\label{eq:tilted_risk_def_appconv}
\widetilde{R}_t(\mathbf{w})
:=
\sum_{j=1}^K (q_t^\star)_j\, r(\mathbf{w},y_j)
=
\mathbf{w}^\top \mathbf{A}_{\mathrm{adv},t}\mathbf{w}
+
\mathbf{b}_{\mathrm{adv},t}^\top \mathbf{w}
+
1,
\end{equation}
with $\mathbf{A}_{\mathrm{adv},t}=\sum_j (q_t^\star)_j \mathbf{A}_j$ and
$\mathbf{b}_{\mathrm{adv},t}=\sum_j (q_t^\star)_j \mathbf{b}_j$.
We use the proximal QP update
\begin{equation}
\label{eq:prox_qp_update_appconv}
\mathbf{w}_{t+1}
=
\arg\min_{\mathbf{w}\in\Delta_M}
\Big\{
\widetilde{R}_t(\mathbf{w}) + \frac{\rho}{2}\|\mathbf{w}-\mathbf{w}_t\|_2^2
\Big\},
\qquad \rho>0,
\end{equation}
and define the step $\mathbf{d}_t:=\mathbf{w}_{t+1}-\mathbf{w}_t$.

\begin{lemma}[Well-defined proximal update and VI/KKT condition]
\label{lem:prox_well_defined_vi_appconv}
For any $\rho>0$, the subproblem \eqref{eq:prox_qp_update_appconv} admits a \emph{unique} minimizer
$\mathbf{w}_{t+1}\in\Delta_M$. Moreover, $\mathbf{w}_{t+1}$ satisfies the variational inequality
\begin{equation}
\label{eq:prox_vi_appconv}
\big\langle \nabla \widetilde{R}_t(\mathbf{w}_{t+1}) + \rho(\mathbf{w}_{t+1}-\mathbf{w}_t),\; \mathbf{w}-\mathbf{w}_{t+1}\big\rangle \ge 0
\qquad \forall \mathbf{w}\in\Delta_M,
\end{equation}
equivalently,
\begin{equation}
\label{eq:prox_kkt_appconv}
\mathbf{0}\in \nabla \widetilde{R}_t(\mathbf{w}_{t+1}) + \rho(\mathbf{w}_{t+1}-\mathbf{w}_t) + N_{\Delta_M}(\mathbf{w}_{t+1}),
\end{equation}
where $N_{\Delta_M}(\cdot)$ is the normal cone of the simplex.
\end{lemma}

\begin{proof}
$\widetilde{R}_t$ is convex quadratic on $\Delta_M$ and the proximal term is $\rho$-strongly convex in $\mathbf{w}$,
so the objective is $\rho$-strongly convex. Minimization over the closed convex set $\Delta_M$ therefore has a unique
solution. The VI/KKT conditions \eqref{eq:prox_vi_appconv}--\eqref{eq:prox_kkt_appconv} are the standard first-order
optimality conditions for minimizing a differentiable convex function over a closed convex set.
\end{proof}

\begin{lemma}[Prox-QP implies a negative directional derivative]
\label{lem:prox_negative_dir_appconv}
Fix $\mathbf{w}_t\in\mathcal{W}_\tau$ and let $\mathbf{w}_{t+1}$ be the minimizer of \eqref{eq:prox_qp_update_appconv}.
Then
\begin{equation}
\label{eq:prox_neg_inner_tilted_appconv}
\big\langle \nabla \widetilde{R}_t(\mathbf{w}_t),\, \mathbf{d}_t \big\rangle \le -\rho \|\mathbf{d}_t\|_2^2.
\end{equation}
Consequently,
\begin{equation}
\label{eq:prox_neg_inner_kappa_appconv}
\big\langle \nabla \kappa_\tau(\mathbf{w}_t),\, \mathbf{d}_t \big\rangle \le -c_t\,\rho \|\mathbf{d}_t\|_2^2,
\qquad
c_t:=\frac{\kappa_t}{\mathbb{E}_{\mathbf{q}_t^\star}[r(\mathbf{w}_t,Y)]-\tau}>0,
\end{equation}
where $\kappa_t=\kappa_\tau(\mathbf{w}_t)$ and $\mathbf{q}_t^\star=\mathbf{q}^\star(\kappa_t,\mathbf{w}_t)$.
\end{lemma}

\begin{proof}
From \eqref{eq:prox_kkt_appconv} we have
$-\rho\mathbf{d}_t\in \nabla \widetilde{R}_t(\mathbf{w}_{t+1})+N_{\Delta_M}(\mathbf{w}_{t+1})$.
Equivalently, for all $\mathbf{w}\in\Delta_M$,
\[
\big\langle \nabla \widetilde{R}_t(\mathbf{w}_{t+1})+\rho\mathbf{d}_t,\ \mathbf{w}-\mathbf{w}_{t+1}\big\rangle \ge 0.
\]
Taking $\mathbf{w}=\mathbf{w}_t$ yields
$\langle \nabla \widetilde{R}_t(\mathbf{w}_{t+1}),\mathbf{d}_t\rangle \le -\rho\|\mathbf{d}_t\|_2^2$.
Since $\widetilde{R}_t$ is convex quadratic, its Hessian is PSD, hence
$\langle \nabla \widetilde{R}_t(\mathbf{w}_t),\mathbf{d}_t\rangle
\le \langle \nabla \widetilde{R}_t(\mathbf{w}_{t+1}),\mathbf{d}_t\rangle$,
which gives \eqref{eq:prox_neg_inner_tilted_appconv}. The second inequality
\eqref{eq:prox_neg_inner_kappa_appconv} follows from Lemma~\ref{lem:kappa_grad_appconv}, i.e.,
$\nabla \kappa_\tau(\mathbf{w}_t)=c_t \nabla \widetilde{R}_t(\mathbf{w}_t)$ with $c_t>0$.
\end{proof}

\begin{corollary}[Stopping certificate via KKT residual]
\label{cor:stopping_certificate_appconv}
Let $\mathbf{w}_{t+1}$ be the proximal update \eqref{eq:prox_qp_update_appconv}. Then
\begin{equation}
\label{eq:kkt_residual_bound_appconv}
\mathrm{dist}\!\left(\mathbf{0},\ \nabla \widetilde{R}_t(\mathbf{w}_{t+1}) + N_{\Delta_M}(\mathbf{w}_{t+1})\right)
\le \rho\,\|\mathbf{w}_{t+1}-\mathbf{w}_t\|_2.
\end{equation}
In particular, if the algorithm stops when $\|\mathbf{w}_{t+1}-\mathbf{w}_t\|_2\le \varepsilon_{\mathrm{stop}}$,
then the stationarity residual of the tilted-risk problem at
$\mathbf{w}_{t+1}$ is at most
$\rho\,\varepsilon_{\mathrm{stop}}$.
\end{corollary}

\begin{proof}
From \eqref{eq:prox_kkt_appconv},
$-\rho(\mathbf{w}_{t+1}-\mathbf{w}_t)\in \nabla \widetilde{R}_t(\mathbf{w}_{t+1}) + N_{\Delta_M}(\mathbf{w}_{t+1})$.
Thus $\mathrm{dist}(\mathbf{0},S)\le \|s\|$ for any $s\in S$ implies \eqref{eq:kkt_residual_bound_appconv}.
\end{proof}

\begin{lemma}[Uniform sufficient decrease for $\kappa_\tau$]
\label{lem:prox_suff_decrease_appconv}
Under Assumption~\ref{ass:compact_wellposed}, let $\mathcal{K}$ be a compact subset of the open domain $\mathcal{W}_\tau$. There exists an $\epsilon > 0$ such that the closed $\epsilon$-neighborhood $\mathcal{K}_\epsilon := \{\mathbf{w} \in \Delta_M : \mathrm{dist}(\mathbf{w}, \mathcal{K}) \le \epsilon\}$ is entirely contained within $\mathcal{W}_\tau$. Let $L<\infty$ be the Lipschitz constant of $\nabla \kappa_\tau$ on $\mathcal{K}_\epsilon$, let $c_{\min}>0$ be the uniform lower bound from Lemma~\ref{lem:cmin}, and let $M_{\mathrm{grad}} < \infty$ be a uniform upper bound on $\|\nabla \widetilde{R}_t(\mathbf{w})\|_2$ over the simplex. 

If $\rho > \max\left\{\frac{L}{2c_{\min}}, \frac{M_{\mathrm{grad}}}{\epsilon}\right\}$, then there exists a constant $c>0$ independent of $t$ such that
\begin{equation}
\label{eq:prox_sufficient_decrease_appconv}
\kappa_\tau(\mathbf{w}_{t+1}) \le \kappa_\tau(\mathbf{w}_t) - c\,\|\mathbf{w}_{t+1}-\mathbf{w}_t\|_2^2.
\end{equation}
\end{lemma}

\begin{proof}
To validly apply the standard descent lemma, the entire line segment connecting $\mathbf{w}_t$ and $\mathbf{w}_{t+1}$ must lie within a domain where $\kappa_\tau$ is $L$-smooth. While $\mathcal{W}_\tau$ is not globally convex, our definition of the compact $\epsilon$-neighborhood $\mathcal{K}_\epsilon \subset \mathcal{W}_\tau$ guarantees that $\nabla \kappa_\tau$ is $L$-Lipschitz continuous on $\mathcal{K}_\epsilon$ for some $L < \infty$.

We first prove that the proximal step bound traps the segment within this neighborhood. From the proximal optimality condition~\eqref{eq:prox_kkt_appconv}, taking the inner product with $\mathbf{d}_t = \mathbf{w}_{t+1} - \mathbf{w}_t$ and applying the Cauchy-Schwarz inequality yields:
\[
\rho\|\mathbf{d}_t\|_2^2 \le \langle -\nabla \widetilde{R}_t(\mathbf{w}_{t+1}), \mathbf{d}_t \rangle \le \|\nabla \widetilde{R}_t(\mathbf{w}_{t+1})\|_2 \|\mathbf{d}_t\|_2.
\]
Since the quadratic risk gradients are uniformly bounded on the simplex, we have $\|\nabla \widetilde{R}_t(\mathbf{w}_{t+1})\|_2 \le M_{\mathrm{grad}}$. Thus, the step size is strictly bounded by $\|\mathbf{d}_t\|_2 \le M_{\mathrm{grad}} / \rho$.

By enforcing the condition $\rho > M_{\mathrm{grad}} / \epsilon$, we mathematically guarantee that $\|\mathbf{w}_{t+1} - \mathbf{w}_t\|_2 < \epsilon$. Because $\mathbf{w}_t \in \mathcal{K}$, this ensures the entire straight-line segment $[\mathbf{w}_t, \mathbf{w}_{t+1}]$ remains strictly inside the safe neighborhood $\mathcal{K}_\epsilon$. 

Because the segment does not exit the valid domain, we can safely apply the standard descent lemma along it:
\[
\kappa_\tau(\mathbf{w}_{t+1}) \le \kappa_\tau(\mathbf{w}_t)
+ \langle \nabla \kappa_\tau(\mathbf{w}_t),\mathbf{d}_t\rangle + \frac{L}{2}\|\mathbf{d}_t\|_2^2.
\]
Using \eqref{eq:prox_neg_inner_kappa_appconv} and $c_t\ge c_{\min}$ gives
\[
\kappa_\tau(\mathbf{w}_{t+1}) \le \kappa_\tau(\mathbf{w}_t) - \Big(c_{\min}\rho - \frac{L}{2}\Big)\|\mathbf{d}_t\|_2^2.
\]
Since $\rho > \frac{L}{2c_{\min}}$, we set $c:=c_{\min}\rho-\frac{L}{2}>0$ to obtain \eqref{eq:prox_sufficient_decrease_appconv}.
\end{proof}

\subsection{Convergence to a stationary point}
\label{app:rse_convergence:stationary}

Let $N_{\Delta_M}(\mathbf{w})$ denote the normal cone of $\Delta_M$ at $\mathbf{w}$.

\begin{definition}[Stationary point of $\kappa_\tau$ on the simplex]
\label{def:stationary_kappa_appconv}
A point $\mathbf{w}^\star\in\mathcal{W}_\tau$ is \emph{stationary} for the robust satisficing objective if
\[
\mathbf{0}\in \nabla_{\mathbf{w}}\kappa_\tau(\mathbf{w}^\star) + N_{\Delta_M}(\mathbf{w}^\star).
\]
\end{definition}

\begin{definition}[$\eta$-stationarity on the simplex]
\label{def:eta_stationary_kappa_appconv}
For $\eta\ge0$, a point $\mathbf w\in\mathcal W_\tau$ is called
$\eta$-stationary for $\min_{\mathbf w\in\Delta_M}\kappa_\tau(\mathbf w)$ if
\[
\operatorname{dist}
\left(
\mathbf 0,\,
\nabla_{\mathbf w}\kappa_\tau(\mathbf w)+N_{\Delta_M}(\mathbf w)
\right)
\le \eta.
\]
The case $\eta=0$ recovers Definition~\ref{def:stationary_kappa_appconv}.
\end{definition}

\begin{theorem}[Convergence of proximal RSE to a stationary point]
\label{thm:rse_convergence_appconv}
Suppose Assumption~\ref{ass:compact_wellposed} holds and the update is the proximal QP step \eqref{eq:prox_qp_update_appconv}
with a fixed $\rho$ chosen so that Lemma~\ref{lem:prox_suff_decrease_appconv} holds. Then:
\begin{enumerate}
\item The sequence $\{\kappa_\tau(\mathbf{w}_t)\}_{t\ge 0}$ is nonincreasing and convergent.
\item The steps vanish: $\|\mathbf{w}_{t+1}-\mathbf{w}_t\|_2\to 0$.
\item Every limit point of $\{\mathbf{w}_t\}$ is a stationary point in the sense of
Definition~\ref{def:stationary_kappa_appconv}. In particular, if the stationary point is unique, then
$\mathbf{w}_t\to \mathbf{w}^\star$.
\end{enumerate}
\end{theorem}

\begin{proof}
(1) Lemma~\ref{lem:prox_suff_decrease_appconv} gives monotone decrease of $\kappa_\tau(\mathbf{w}_t)$, and
$\kappa_\tau(\mathbf{w}_t)\ge 0$, hence convergence.

(2) Summing \eqref{eq:prox_sufficient_decrease_appconv} from $t=0$ to $T-1$ yields
\[
c\sum_{t=0}^{T-1}\|\mathbf{w}_{t+1}-\mathbf{w}_t\|_2^2
\le
\kappa_\tau(\mathbf{w}_0)-\kappa_\tau(\mathbf{w}_T)
\le
\kappa_\tau(\mathbf{w}_0),
\]
so $\sum_{t\ge 0}\|\mathbf{w}_{t+1}-\mathbf{w}_t\|_2^2<\infty$ and therefore
$\|\mathbf{w}_{t+1}-\mathbf{w}_t\|_2\to 0$.

(3) Since $\Delta_M$ is compact, pick a convergent subsequence $\mathbf{w}_{t_n}\to \bar{\mathbf{w}}$.
By (2), also $\mathbf{w}_{t_n+1}\to \bar{\mathbf{w}}$.
Let $\kappa_{t_n}=\kappa_\tau(\mathbf{w}_{t_n})$ and $\mathbf{q}_{t_n}^\star=\mathbf{q}^\star(\kappa_{t_n},\mathbf{w}_{t_n})$.
By continuity of $\mathbf{w}\mapsto \kappa_\tau(\mathbf{w})$ on $\mathcal{W}_\tau$ and of \eqref{eq:qstar_def_appconv},
we have $\kappa_{t_n}\to \bar{\kappa}:=\kappa_\tau(\bar{\mathbf{w}})$ and
$\mathbf{q}_{t_n}^\star\to \bar{\mathbf{q}}^\star:=\mathbf{q}^\star(\bar{\kappa},\bar{\mathbf{w}})$.

The proximal optimality condition \eqref{eq:prox_kkt_appconv} at time $t_n$ is
\[
\mathbf{0}\in \nabla \widetilde{R}_{t_n}(\mathbf{w}_{t_n+1}) + \rho(\mathbf{w}_{t_n+1}-\mathbf{w}_{t_n})
+ N_{\Delta_M}(\mathbf{w}_{t_n+1}),
\]
with $\nabla \widetilde{R}_{t_n}(\mathbf{w})=\sum_{j=1}^K (q_{t_n}^\star)_j \nabla r(\mathbf{w},y_j)$.
Since $\mathbf{w}_{t_n+1}-\mathbf{w}_{t_n}\to 0$, passing to the limit and using continuity of $\nabla r(\cdot,y_j)$ yields
\[
\mathbf{0}\in \sum_{j=1}^K (\bar{q}^\star)_j \nabla r(\bar{\mathbf{w}},y_j) + N_{\Delta_M}(\bar{\mathbf{w}}).
\]
Finally, Lemma~\ref{lem:kappa_grad_appconv} shows that
$\nabla \kappa_\tau(\bar{\mathbf{w}})$ is a strictly positive scalar multiple of
$\sum_j (\bar{q}^\star)_j \nabla r(\bar{\mathbf{w}},y_j)$, hence
\[
\mathbf{0}\in \nabla \kappa_\tau(\bar{\mathbf{w}}) + N_{\Delta_M}(\bar{\mathbf{w}}),
\]
so $\bar{\mathbf{w}}$ is stationary (Definition~\ref{def:stationary_kappa_appconv}).
\end{proof}

\subsection{Proof of Proposition~\ref{prop:approx_root_stationarity_main}}
\label{app:rse_convergence:approx_root}

\begin{remark}[Exact vs.\ approximate $\kappa_\tau(\mathbf{w})$]
\label{rem:approx_kappa_appconv}
The convergence proof above treats
\(\kappa_t=\kappa_\tau(\mathbf w_t)\) as the exact root of
\(\Phi(\kappa;\mathbf w_t)=\tau\). In implementation, \(\widehat\kappa_t\) is obtained
by taking the reciprocal of the truncated
Lagrange--B\"urmann inverse series. Its scalar root residual is
\[
\varepsilon_t
:=
\Phi(\widehat\kappa_t;\mathbf w_t)-\tau.
\]
The next proposition shows that this residual directly controls the exact
stationarity residual of \(\kappa_\tau(\cdot)\) on the simplex.
\end{remark}

\begin{proposition}[Approximate-root stationarity certificate]
\label{prop:approx_root_stationarity_appconv}
Suppose Assumption~\ref{ass:compact_wellposed} holds. Let
\[
\kappa_t:=\kappa_\tau(\mathbf w_t),
\qquad
\widehat\kappa_t
\]
denote the exact and truncated roots, respectively, and define
\[
\varepsilon_t
:=
\Phi(\widehat\kappa_t;\mathbf w_t)-\tau.
\]
Assume $\widehat\kappa_t>0$ for every $t$.
Assume the truncation is accurate enough that the segment between
\(\widehat\kappa_t\) and \(\kappa_t\) lies in a neighborhood on which
\[
|\partial_\kappa\Phi(\kappa;\mathbf w_t)|\ge m_\kappa>0
\]
uniformly in \(t\). Let \(\widehat{\mathbf q}_t
=\mathbf q^\star(\widehat\kappa_t,\mathbf w_t)\), and suppose
\(\mathbf w_{t+1}\) is the proximal QP update formed using
\(\widehat{\mathbf q}_t\). Then there exist constants
\(C_\varepsilon,C_d<\infty\), independent of \(t\), such that
\[
\operatorname{dist}
\left(
\mathbf 0,\,
\nabla_{\mathbf w}\kappa_\tau(\mathbf w_{t+1})
+
N_{\Delta_M}(\mathbf w_{t+1})
\right)
\le
C_\varepsilon |\varepsilon_t|
+
C_d\|\mathbf w_{t+1}-\mathbf w_t\|_2.
\]
Consequently, if the algorithm stops with
\[
|\varepsilon_t|\le \varepsilon_{\mathrm{root}},
\qquad
\|\mathbf w_{t+1}-\mathbf w_t\|_2\le \varepsilon_{\mathrm{stop}},
\]
then \(\mathbf w_{t+1}\) is
\[
\left(
C_\varepsilon\varepsilon_{\mathrm{root}}
+
C_d\varepsilon_{\mathrm{stop}}
\right)\text{-stationary}
\]
in the sense of Definition~\ref{def:eta_stationary_kappa_appconv}.
\end{proposition}

\begin{proof}
We first control the scalar root error. Since
\[
\Phi(\kappa_t;\mathbf w_t)=\tau,
\]
the mean value theorem gives some
\(\widetilde\kappa_t\) between \(\widehat\kappa_t\) and \(\kappa_t\) such that
\[
\Phi(\widehat\kappa_t;\mathbf w_t)-\Phi(\kappa_t;\mathbf w_t)
=
\partial_\kappa\Phi(\widetilde\kappa_t;\mathbf w_t)
(\widehat\kappa_t-\kappa_t).
\]
Using the uniform slope bound,
\[
|\widehat\kappa_t-\kappa_t|
\le
\frac{|\Phi(\widehat\kappa_t;\mathbf w_t)-\tau|}{m_\kappa}
=
\frac{|\varepsilon_t|}{m_\kappa}.
\]

Next define, for \(\kappa>0\),
\[
G(\kappa,\mathbf u;\mathbf v)
:=
\sum_{j=1}^K
q_j^\star(\kappa,\mathbf u)\,
\nabla_{\mathbf v}r(\mathbf v,y_j).
\]
There exists a constant $L_G<\infty$, independent of $t$,
such that
\[
\begin{aligned}
&\left\|
G(\kappa_\tau(\mathbf w_{t+1}),\mathbf w_{t+1};
  \mathbf w_{t+1})
-
G(\widehat\kappa_t,\mathbf w_t;\mathbf w_{t+1})
\right\|_2\\
&\qquad\le
L_G\left(
|\kappa_\tau(\mathbf w_{t+1})-\widehat\kappa_t|
+\|\mathbf w_{t+1}-\mathbf w_t\|_2
\right).
\end{aligned}
\]

\begin{remark}
\label{rem:G_comparison_bound}
A finite comparison constant $L_G$ can be obtained
from the exponential-tilt formula, using boundedness
of the Brier risks and their gradients on the simplex
and the positive lower bound of $\kappa_\tau$ on
$\mathcal K$.
\end{remark}

By the same neighborhood and near/far-pair argument used
in Lemma~\ref{lem:kappa_grad_lipschitz}, applied to
$\kappa_\tau$ using its bounded gradient on a compact
neighborhood of $\mathcal K$, the map $\kappa_\tau$
is Lipschitz on $\mathcal K$ with some constant
$L_\kappa<\infty$, 

\[
|\kappa_\tau(\mathbf w_{t+1})-\widehat\kappa_t|
\le
L_\kappa\|\mathbf w_{t+1}-\mathbf w_t\|_2
+
\frac{|\varepsilon_t|}{m_\kappa}.
\]
Therefore
\[
\left\|
G(\kappa_\tau(\mathbf w_{t+1}),\mathbf w_{t+1};\mathbf w_{t+1})
-
G(\widehat\kappa_t,\mathbf w_t;\mathbf w_{t+1})
\right\|_2
\le
L_G
\left(
(L_\kappa+1)\|\mathbf w_{t+1}-\mathbf w_t\|_2
+
\frac{|\varepsilon_t|}{m_\kappa}
\right).
\]

The approximate-root QP optimality condition gives
\[
\mathbf 0
\in
G(\widehat\kappa_t,\mathbf w_t;\mathbf w_{t+1})
+
\rho(\mathbf w_{t+1}-\mathbf w_t)
+
N_{\Delta_M}(\mathbf w_{t+1}).
\]
Now define
\[
c(\mathbf w)
:=
\frac{
\kappa_\tau(\mathbf w)
}{
\mathbb E_{\mathbf q^\star(\kappa_\tau(\mathbf w),\mathbf w)}
[r(\mathbf w,Y)]
-\tau
}.
\]
By Lemma~\ref{lem:uniform_gap_bounds}, \(c(\mathbf w)\) is uniformly bounded
above on \(\mathcal K\); let
\[
c_{\max}:=\sup_{\mathbf w\in\mathcal K}c(\mathbf w)<\infty.
\]
The exact implicit-gradient formula gives
\[
\nabla_{\mathbf w}\kappa_\tau(\mathbf w_{t+1})
=
c(\mathbf w_{t+1})
G(\kappa_\tau(\mathbf w_{t+1}),\mathbf w_{t+1};\mathbf w_{t+1}).
\]
Since \(N_{\Delta_M}(\mathbf w_{t+1})\) is a cone, multiplying the QP
optimality condition by the positive scalar \(c(\mathbf w_{t+1})\) preserves
membership in the normal cone. Hence
\[
\operatorname{dist}
\left(
\mathbf 0,\,
\nabla_{\mathbf w}\kappa_\tau(\mathbf w_{t+1})
+
N_{\Delta_M}(\mathbf w_{t+1})
\right)
\]
is bounded by
\[
c(\mathbf w_{t+1})
\left\|
G(\kappa_\tau(\mathbf w_{t+1}),\mathbf w_{t+1};\mathbf w_{t+1})
-
G(\widehat\kappa_t,\mathbf w_t;\mathbf w_{t+1})
\right\|_2
+
c(\mathbf w_{t+1})\rho
\|\mathbf w_{t+1}-\mathbf w_t\|_2.
\]
Using the previous Lipschitz bound and \(c(\mathbf w_{t+1})\le c_{\max}\), we get
\[
\operatorname{dist}
\left(
\mathbf 0,\,
\nabla_{\mathbf w}\kappa_\tau(\mathbf w_{t+1})
+
N_{\Delta_M}(\mathbf w_{t+1})
\right)
\le
C_\varepsilon |\varepsilon_t|
+
C_d\|\mathbf w_{t+1}-\mathbf w_t\|_2,
\]
where one may take
\[
C_\varepsilon
=
\frac{c_{\max}L_G}{m_\kappa},
\qquad
C_d
=
c_{\max}\big(L_G(L_\kappa+1)+\rho\big).
\]
This gives the desired bound and thus completes the proof.
\end{proof}

\end{document}